%% file: main.tex
\documentclass[12pt]{article}

\usepackage[margin=1in]{geometry}

\usepackage[round]{natbib}

\usepackage[utf8]{inputenc} 
\usepackage[T1]{fontenc}    

\usepackage[
  bookmarks=true,
  bookmarksnumbered=true,
  bookmarksopen=true,
  pdfborder={0 0 0},
  breaklinks=true,
  colorlinks=true,
  linkcolor=black,
  citecolor=blue,
  filecolor=black,
  urlcolor=blue,
]{hyperref}

\usepackage{url}      
\usepackage{booktabs} 
\usepackage{amsfonts,dsfont,amssymb}    
\usepackage{nicefrac}     
\usepackage{microtype}    
\usepackage{xcolor}    
\usepackage{lmodern}

\title{EconEvals: Benchmarks and Litmus Tests for \\ Economic Decision-Making by LLM Agents\thanks{We gratefully acknowledge support in the form of API credits from Anthropic, Google, OpenAI, and AISST. Fish was supported by an NSF Graduate Research Fellowship and a Kempner Institute Graduate Fellowship. Gonczarowski's research was supported by the National Science Foundation (NSF-BSF grant No.\ 2343922), the Harvard FAS Dean’s Competitive Fund for Promising Scholarship, and the Harvard FAS Inequality in America Initiative. Shorrer's research was supported by a grant from the United States–Israel Binational Science Foundation (BSF grant 2022417). We thank Eric Bigelow, Pranav Misra, Core Francisco Park, Sunny Qin, and attendees of the Harvard Kempner Institute All Hands Meeting for insightful comments and discussions.}}

\hypersetup{pdfauthor={Sara Fish, Julia Shephard, Minkai Li, Ran Shorrer, Yannai A. Gonczarowski},
pdftitle={EconEvals: Benchmarks and Litmus Tests for Economic Decision-Making by LLM Agents}}

\author{
Sara Fish\\Harvard University\\
  \and
  Julia Shephard\thanks{Equal contribution.} \\
  Harvard University \\
  \and
  Minkai Li\footnotemark[\value{footnote}] \\
  Harvard University \\
  \and
  Ran Shorrer \\
  Penn State University \\
  \and
  Yannai A. Gonczarowski \\
  Harvard University \\
}

\date{February 18, 2026}

\providecommand{\tightlist}{
  \setlength{\itemsep}{0pt}\setlength{\parskip}{0pt}}
\usepackage{graphicx}
\usepackage{amsmath}
\usepackage{xcolor}
\usepackage{fancyvrb}
\usepackage{fvextra}
\usepackage[capitalize,noabbrev]{cleveref}

\usepackage{tabularx,booktabs}

\usepackage[normalem]{ulem}

\begin{document}

\maketitle

\begin{abstract}

We develop evaluation methods for measuring the economic decision-making capabilities and tendencies of LLMs. First, we develop benchmarks derived from key problems in economics---procurement, scheduling, and pricing---that test an LLM's ability to learn from the environment in context. Second, we develop the framework of \emph{litmus tests}, evaluations that quantify an LLM's choice behavior on a stylized decision-making task with multiple conflicting objectives. Each litmus test outputs a \emph{litmus score}, which quantifies an LLM's tradeoff response, a \emph{reliability score}, which measures the coherence of an LLM's choice behavior, and a \emph{competency score}, which measures an LLM's capability at the same task when the conflicting objectives are replaced by a single, well-specified objective. Evaluating a broad array of frontier LLMs, we (1) investigate changes in LLM capabilities and tendencies over time, (2) derive economically meaningful insights from the LLMs' choice behavior and chain-of-thought, (3) validate our litmus test framework by testing self-consistency, robustness, and generalizability. Overall, this work provides a foundation for evaluating LLM agents as they are further integrated into economic decision-making.

\end{abstract}

\clearpage 
\section{Introduction}

Large language model (LLM) capabilities are improving rapidly across a broad array of domains \citep{epoch2025aicapabilitiesoverpastyear}. Accordingly, organizations may increasingly incorporate LLM-based technology in their economic decision-making workflows.\footnote{In an April 2025 appearance on Bloomberg Technology, Visa CEO Ryan McInerney describes Visa's vision for ``[LLM] agents to buy on your behalf'' \citep{bloomberg_visa_2025}. \cite{menache_how_2025} describe how LLM-based technology enabled productivity gains in Microsoft's supply chain management, writing: \emph{``In the next few years LLM-based technology will support end-to-end decision-making scenarios.''} \cite{surane_jpmorgan_2026} reports that JPMorgan Chase plans to use AI to make shareholder voting decisions. More broadly, \cite{anthropic_economic_impacts_2025} analyze usage data of Claude, an LLM chatbot by Anthropic, and find that 5.9\% of conversations relate to business or finance.} This motivates evaluating the suitability of LLMs across different economic decision-making contexts. 

Standard evaluation approaches are not well-equipped to measure certain core aspects of economic decision-making. First, economic decision-making often requires \emph{learning from the environment in context}. A pricing agent, for example, may face unknown demand conditions that can only be discovered by experimenting with prices and observing the results. Second, economic decision-making often involves \emph{navigating open-ended tradeoffs}. For example, when a manager assigns tasks to workers, they may face a tradeoff between efficiency and equality. In such cases, there is not necessarily a single objectively correct answer. Rather, different goals may be simultaneously desirable but ultimately incompatible, leading to several possible answers that could each be viewed as ``correct.''

In this work, we develop methods for evaluating these two aspects of economic decision-making in LLMs. 

First, to evaluate an LLM's capacity for learning from the environment in context, we advocate for the use of benchmarks in which an LLM agent repeatedly acts in a stylized environment, whose specification the LLM agent can only learn from deliberate multi-turn exploration, and is scored based on its ability to optimize for a well-specified objective. We develop three benchmarks of this form simulating three core business tasks: \emph{procurement}, which requires bin-packing a budget with items of unknown (to be learned) quality; \emph{scheduling}, which requires producing a stable matching between workers and tasks with unknown (to be learned) preferences; and \emph{pricing}, which requires setting profit-maximizing prices by adjusting to an unknown (to be learned) demand environment. We employ each of the benchmarks at three different difficulty levels: \textsc{Basic}, \textsc{Medium}, and \textsc{Hard}. As the benchmarks consist of synthetic environments, they can be quickly scaled in size and complexity even beyond these three levels as LLM capabilities progress. 

Second, to evaluate how LLMs navigate open-ended tradeoffs, we develop the framework of \textit{litmus tests}: evaluations that quantify an LLM's choice behavior on a stylized decision-making task with multiple conflicting objectives. Each litmus test outputs a \emph{litmus score}, which measures how the LLM trades off between the conflicting objectives, as well as two \emph{meaningfulness} scores: a \emph{reliability score}, which measures the extent to which the litmus score represents a consistent tendency rather than noise due to LLM incoherence, and a \emph{competency score}, which measures the LLM's capability at optimizing a single, well-specified objective in the same---or very similar---setting (absent sufficient such capability, the LLM cannot be expected to meaningfully trade off the conflicting objectives). We develop three litmus tests to showcase the broad scope of this framework: \emph{Patience versus Impatience}, which measures intertemporal preferences from single-shot queries; \emph{Efficiency versus Equality}, which elicits efficiency--equality tradeoff preferences in a repeated-interaction task allocation setting; and \emph{Collusiveness versus Competitiveness}, which measures collusive tendencies in a multi-agent pricing setting.

\paragraph{Experimental Results.} To showcase the potential of this methodology, we evaluate a broad array of frontier LLMs released between May 2024 and November 2025 on our benchmarks and litmus tests.\footnote{The EconEvals code is publicly available at \url{https://github.com/sara-fish/econ-evals-paper}.} 

Our first category of results investigates trends in LLM capabilities and tendencies over time. Across all three benchmarks, we observe that more recently-released LLMs generally earn higher benchmark scores, indicating that (for the LLMs we test) the capabilities of LLMs in economic decision-making contexts are improving with time. For the litmus tests, we observe that more recently-released LLMs generally earn higher competency and reliability scores, indicating that the meaningfulness of LLM choice behavior under conflicting objectives is increasing with time. 

Our second category of results derives economic insights from the LLMs' benchmark performance. For each of the benchmarks, we analyze not just the LLMs' overall benchmark score, but also various features of their behavior over the course of each experimental run, to understand in greater depth how they reason about and learn from the benchmark environment (as well as other economic insights). For example, in the scheduling benchmark, in which the LLM agent must produce a stable matching between workers and tasks with unknown preferences over each other (that must be learned from exploration), we observe that better-performing LLMs tend to reason about more complex (i.e., longer) preference lists.

Our third category of results involves validating our litmus test framework. By analyzing the competency, reliability, and litmus scores of the LLMs we test, we provide preliminary answers to questions such as: To what extent are an LLM's choices within a given litmus test meaningful and self-consistent? To what extent is an LLM's choice behavior in a given litmus test robust to small changes in the prompt? And finally, to what extent do elicited litmus scores generalize across contexts? To address the first question, we analyze the elicited competency and reliability scores, and find that the vast majority of recently-released LLMs exhibit high competency and reliability scores. To address the second question, we re-run the Patience versus Impatience litmus test with 36 prompt variants, and find that changes to the prompt have little impact on the litmus score. Finally, to address the third question, we compare the results of our Efficiency versus Equality litmus test with the two evaluations of LLM political bias \citep{westwood_measuring_2025,openai_defining_2025}, and find that a tendency to prefer equality in our litmus test corresponds to a greater bias in favor of the U.S.~Democratic party. Overall, these findings point to the promise of litmus tests as a way to coherently quantify aspects of LLMs' ``character,'' in ways that generalize across domains.

\section{Related Work}\label{sec:related-work}

\paragraph{LLMs in Economics and the Social Sciences.} A rich literature seeks to use LLMs to simulate human subjects in lab experiments in the social sciences \citep[e.g.,][]{aher_using_2023,horton_large_2023,goli_frontiers_2024,manning_automated_2024}. By contrast, we study LLMs as economic agents in their own right. In this sense, our study is related to a large literature that studies how other AI algorithms interact with markets and the broader society \citep[see, e.g.,][]{calvano2020protecting,gillis2021fairness,banchio2022artificial,liang2022algorithmic,banchio2023adaptive,brunnermeier2023inverse,brynjolfsson_generative_2023,raymond2023market,rocher2023adversarial}.

We in particular draw inspiration from prior work that studies the behavior of LLMs in stylized economic environments. \cite{krishnamurthy_can_2024} study in-context exploration of LLMs in a multi-armed bandit environment. (Interestingly, they observe a tendency for underexploration in the early-generation LLMs they test, which we also find in \cref{sec:exploration}.) \cite{deng_llms_2024} study the strategic behavior of two LLM agents in a Rubinstein bargaining setting, comparing LLM behavior with theoretical predictions. Their work includes a preliminary qualitative analysis demonstrating that LLM negotiators ``have some kind of internal discount factors,'' a premise we validate more broadly in \cref{sec:reliability_scores}. \cite{fish_algorithmic_2024} study the collusiveness of two LLM agents in a Bertrand oligopoly environment and analyze the LLMs' chain-of-thought outputs to understand their strategies in greater depth---a methodology we adopt and extend in \cref{subsec:economic_analysis}. Unlike \cite{deng_llms_2024,fish_algorithmic_2024}, our benchmarks focus on single-agent contexts in which we can isolate the capabilities of a single LLM agent. Building on \cite{krishnamurthy_can_2024}, we design benchmark environments that target many additional kinds of economic decision-making capabilities.

\paragraph{Benchmarks.} Our benchmarks contribute to a growing literature that seeks to analyze LLM capabilities related to economics from a variety of perspectives. The related work can be categorized as follows: 
\begin{itemize}
  \item \emph{Static question answering benchmarks.} These include EconLogicQA, which measures verbal comprehension for economics-related news articles \citep{quan_econlogicqa_2024}; STEER, which consists of textbook-style decision-theoretic questions \citep{raman_steer_2024}; and STEER-ME, which consists of textbook-style LLM-(re)written microeconomics questions \citep{raman2025steerme}.
  \item \emph{Fact retrieval benchmarks.} These include FinanceBench, which tests fact retrieval for financial question answering \citep{islam_financebench_2023}; and EconWebArena, which tests economic fact retrieval from authoritative websites \citep{liu2025econwebarena}.
  \item \emph{End-to-end simulation benchmarks.} These include TheAgentCompany, an end-to-end simulation of a company in which LLM agents use business software \citep{xu2024theagentcompanybenchmarkingllmagents}; and VendingBench, an end-to-end simulation testing an LLM agent's ability to operate a vending machine \citep{backlund_vending-bench_2025}.
\end{itemize}

More broadly, while our benchmark design is motivated by economic decision-making, the LLM capabilities we test also relate more generally to multi-turn incomplete-information LLM agent benchmarks. \citet{ma_agentboard_2024} provide a taxonomy of such benchmarks into four categories: \emph{embodied} (physical instructions), \emph{web} (browser usage), \emph{tool} (measuring the ability to usefully call external functions), and \emph{game} (video game-style environments). For some representative papers in these categories, see e.g.~\citet{wang_voyager_2023,mialon_gaia_2023,zhou_webarena_2023,liu_agentbench_2023,xie_osworld_2024,ma_agentboard_2024,he_webvoyager_2024}.
While our LLM agents interact with the environment via tool usage, this is not the main capability tested, unlike in fact retrieval (or ``tool'') benchmarks. And, while our benchmarks can be viewed as simulations of procurement, scheduling, and pricing scenarios, compared to the end-to-end simulation (or ``game'') benchmarks, our benchmark environments are more stylized. This is a purposeful design choice: stylized environments allow for performance metrics that are directly grounded in an underlying theoretical model. For example, while both VendingBench and our pricing benchmark score the LLM based on its profit earned in a multi-turn simulation, only in our pricing benchmark is it possible to solve for, e.g., the optimal profit-maximizing actions in each period. This in turn allows us to understand, e.g., when a given difficulty level is saturated.\footnote{\cite{backlund_vending-bench_2025} write: \emph{``Since the [VendingBench] benchmark score does not have a defined upper limit, saturation is not a precisely defined point.''}}
Thus, our benchmarks do not fit into any of the above categories. Rather, the primary capability our benchmarks are testing is \emph{optimization}---and more specifically, optimization while learning about a stylized environment in context. 

Finally, our benchmarks contribute to the broader area of benchmark design for frontier LLMs. As LLM capabilities advance, designing benchmarks that give useful signals (e.g., are not saturated) becomes increasingly resource-intensive. For example, when curating the HLE benchmark, \cite{phan_humanitys_2025} spent \$500,000 incentivizing external contributors. Other modern oft-used benchmarks, such as SWE-Bench Verified \citep{chowdhury2024swebenchverified}, GPQA~Diamond \citep{rein_gpqa_2024}, FrontierMath \citep{glazer_frontiermath_2024}, GDPVal \citep{patwardhan_gdpval_2025}, and ARC-AGI \citep{chollet_arc_2025} likewise involve a labor-intensive question curation process. By contrast, as our benchmarks consist of synthetic environments, it is simple to generate additional tasks, even at difficulty levels beyond \textsc{Hard} \citep[see, e.g.,][who also use synthetic environments for this purpose]{valmeekam_planbench_2023}.

\paragraph{Litmus tests.} Our work on litmus tests builds on prior work that shares our aim of measuring aspects of LLMs' economic choice behavior beyond capability. 

Reporting competency and reliability scores alongside litmus scores is a key component of our framework. Regarding competency scores, our methodology builds on \cite{fish_algorithmic_2024,ross_llm_2024}, who require LLMs to pass a ``competence test'' (terminology from \citealt{ross_llm_2024}) as a prerequisite for interpreting LLM choice behavior as meaningful. For example, \cite{fish_algorithmic_2024} utilize a competency test that tests the LLM's ability to find a profit-maximizing price in a single-agent setting, and conduct multi-agent pricing experiments only on sufficiently competent LLMs.\footnote{In their experiments eliciting intertemporal preferences, \cite{ross_llm_2024} assess competency via monotonicity and individual rationality (however, we would argue this is closer to a reliability test than a competency test).} Regarding reliability scores, our methodology relates to \cite{chen_emergence_2023,mazeika2025utilityengineeringanalyzingcontrolling}, who assess the self-consistency of LLM choice behavior (\citealt{chen_emergence_2023} via the generalized axiom of revealed preference, and \citealt{mazeika2025utilityengineeringanalyzingcontrolling} via loss from a best-fit utility function). 
Our contribution is to elucidate the distinct function of each score and to advocate for reporting \emph{both} scores alongside litmus scores, as competency and reliability scores are both necessary, complementary prerequisites for a litmus score to have a meaningful interpretation.

Our \emph{Patience versus Impatience} litmus test follows a rich prior work that measures intertemporal preferences in LLMs. \cite{goli_frontiers_2024} study the effect of language (e.g.~German or Mandarin) on an LLM's choices, \cite{ross_llm_2024} compare with human choice behavior and study the effect of different prompting methods, \cite{deng_llms_2024} elicit discount factors in a Rubinstein bargaining setting, and \cite{mazeika2025utilityengineeringanalyzingcontrolling} suggest potential hyperbolic temporal discounting. Our contribution is (1) more methodologically sound calculation method for the main (litmus) score, (2) more methodologically sound meaningfulness score calculation, (3) measuring two distinct meaningfulness scores---competency score and reliability score---to allow for a more nuanced interpretation of the litmus scores, (4) conducting an extensive prompt robustness check, and (5) running experiments on a broad array of more modern and competent LLMs.

Our high-level aim of measuring the preferences, character or values of LLMs in stylized environments connects to the literature that studies the behavior of LLMs in game theoretic environments \citep[see, e.g.,][]{guo_economics_2024,duan_gtbench_2024,piatti_cooperate_2024,akata_playing_2025,huang_how_2025,agashe_llm-coordination_2025}, or multi-agent economic decision-making settings more generally \citep[see, e.g.,][]{chen2023money,fish_algorithmic_2024}. These works tend to adopt a normative perspective, which sometimes leads to conceptual mismatches regarding what constitutes ``good'' LLM performance: For example, in \cite{guo_economics_2024} it means ``rational'' play, and in \cite{piatti_cooperate_2024,agashe_llm-coordination_2025} it means cooperative play. These desiderata are at odds with each other, e.g., in the Prisoner's Dilemma. By contrast, our litmus test framework adopts a positive perspective---our aim is to measure LLM behavior, rather than normatively specify a single optimization target for quality scores. As a proof-of-concept for multi-agent litmus tests, we adapt \cite{fish_algorithmic_2024}, a multi-agent pricing game, into a litmus test (\emph{Collusiveness vs.~Competitiveness}). 

Finally, our litmus test framework relates more generally to the broader area of understanding how LLMs behave when faced with tradeoffs, beyond economic settings. Prior work studies tradeoffs such as reward maximization vs.~ethical behavior \citep{pan2023machiavelli}, safety vs.~utility \citep{bhatt_cyberseceval_2024}, moral judgements \citep{marraffini_greatest_2024}, honesty vs.~helpfulness \citep{liu_how_2024}, and informational vs.~social utility \citep{murthy_using_2025}. These evaluations measure LLM tradeoff responses using human labels---which can be resource-intensive---or LLM-as-a-judge labels---which can be unreliable. By contrast, our litmus tests utilize stylized environments, which means the metrics that measure tradeoff responses are directly grounded in an underlying theoretical model.

\section{Benchmark Design}\label{sec:benchmark-experimental-design}

We design benchmark environments that simulate three core economic tasks: procurement, scheduling, and pricing. In each setting, the LLM agent acts in the environment for 100 periods. Each period culminates with the LLM agent taking a single action (e.g., setting a price), after which the LLM agent receives feedback.\footnote{In this sense, our environments can be viewed as POMDPs \citep[see, e.g.,][for such framing]{ma_agentboard_2024}.} In all of our environments, there is a well-defined notion of an optimal action (in a given period), and a natural way to measure the relative quality of a non-optimal action (in that period). 

In \cref{subsec:benchmark-interaction-method}, we describe the ``API'' via which the LLM agent acts in the benchmark environment. In \cref{subsec:llm-agent-architecture}, we describe the architecture of the LLM agents we test. Finally, in \cref{subsec:economic_environment}, we provide the design details of the three benchmark environments.

\subsection{Benchmark Interaction Method}\label{subsec:benchmark-interaction-method}

\begin{figure}
    \centering
    \includegraphics[width=0.5\linewidth]{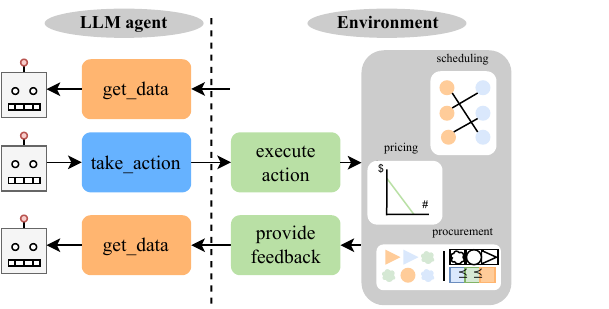}
    \caption{Illustration of how the LLM agent interacts with the benchmark environment. The LLM agent obtains information and takes actions via tool use (see \cref{subsec:benchmark-interaction-method,subsec:llm-agent-architecture}). The environment performs computations based on the tools used and returns information (see \cref{subsec:economic_environment}).}
    \label{fig:architecture}
\end{figure}

Rather than designing benchmark questions with which to query an LLM, we design benchmark \textit{environments} in which an \textit{LLM agent} must act (and is evaluated). LLM agent technology is nascent and there is currently no singular standard interaction protocol.\footnote{Examples include Anthropic's \href{https://www.anthropic.com/news/model-context-protocol}{Model Context Protocol} and Google's \href{https://developers.googleblog.com/en/a2a-a-new-era-of-agent-interoperability/}{Agent2Agent protocol}. See also \citet{chan2025infrastructureaiagents}.} To ensure versatility and future-proofness of our benchmarks, we only require a lightweight interaction protocol using \textbf{tool use} (also referred to as function calling). We select this interaction method because it has rich precedent in the literature on agentic workflows \citep[see, e.g.][]{schick_toolformer_2023} and is included in frontier LLMs as a built-in feature.

Each economic environment is associated with a list of tools. There are two types of tools: \textit{getter tools}, which return information about the environment, and \textit{action tools}, which execute an action (e.g., setting a price). \cref{tab:tools} lists the associated tools for each benchmark environment (further detail in \cref{app:prompts}). When the LLM agent calls a getter tool, the relevant quantity is computed according to parameters of the underlying (synthetic) economic environment and returned; when the LLM agent calls an action tool, the underlying economic environment computes the consequences of that action and advances to the next period. See \cref{fig:architecture} for an illustration.

\begin{table}[h]
\centering
\small 
\begin{tabularx}{\linewidth}{@{}
  l
  >{\raggedright\arraybackslash\hsize=.6\hsize}X
  >{\raggedright\arraybackslash\hsize=1.4\hsize}X
@{}}
\toprule
\textbf{Environment} & \textbf{Action tool} & \textbf{Getter tools}  \\
\midrule
Procurement &
  {submit\_purchase\_plan}  &
  {get\_previous\_purchase\_data},
  {get\_budget},
  {get\_equipment\_information},
  {get\_attempt\_number} \\[2pt]
\hline 
Scheduling &
  {submit\_assignment} &
  {get\_previous\_attempts\_data},
  {{get\_worker\_ids}},
  {get\_task\_ids},
  {get\_attempt\_number} \\[2pt] 
\hline 
Pricing  &
  {set\_prices} &
  {{get\_previous\_pricing\_data}},
  {{get\_product\_ids}},
  {{get\_attempt\_number}}  \\
\bottomrule

\end{tabularx}
\caption{Overview of tools associated with each economic environment}
\label{tab:tools}
\end{table}

Accordingly, any LLM agent capable of using the tools listed in \cref{tab:tools} can be evaluated using our benchmarks. While LLM agents interact with our environments using the above tools, we remark that LLM agents are not limited to using only these tools. For example, in this work we test LLM agents equipped with additional tools allowing for memory between periods (see \cref{subsec:llm-agent-architecture}). As the capacity for LLMs to use increasingly large sets of tools advances, one could imagine evaluating LLM agents that have access to additional tools, e.g., a (secure) Python interpreter.

\subsection{LLM Agent Architecture}\label{subsec:llm-agent-architecture}

For each LLM that we test, we construct an LLM agent by equipping the LLM with tools that allow it to act in the benchmark environment as well as formulate and keep track of its plans. Each period is conducted in a single chat session.\footnote{Our benchmarks thus require a relatively long context window, a condition satisfied by the LLMs we evaluate.} At the start of each period, the LLM agent is given the same initial instructions and a list of tools it can use to interact in the economic environment. The tools include the environment-specific tools described in \cref{tab:tools}, as well as two additional \textit{notes tools}, write\_notes and read\_notes, that allow the LLM agent to read and write notes to itself that persist between periods.
For further details on the functionality of the notes tools, see \cref{app:prompts}.

\subsection{Economic Environments}\label{subsec:economic_environment}

We design three benchmark environments to simulate a broad array of key economic tasks. The environments we construct come in two forms: \textit{stationary} and \textit{non-stationary}.

In the stationary environments (procurement and scheduling), the quality of an action does not depend on the period in which it is taken, and accordingly the LLM agent is scored based on the quality of its best (in procurement) or final (in scheduling) action. In particular, to earn a perfect score in a stationary environment, it suffices for the LLM agent to identify and take an optimal action once.

In the non-stationary environments (pricing), the quality of an action changes over time according to a predictable pattern that the LLM agent must learn. Accordingly, the LLM agent is scored based on its ability to consistently take high-quality actions after an initial exploration period. In particular, to earn a perfect score in a non-stationary environment, the LLM agent must take optimal actions many periods in a row, changing them appropriately as the environment changes.

\subsubsection{Procurement}\label{procurement:specification}

\paragraph{High-level overview.} The LLM agent is given a list of prices for bundles of products (e.g., ``\$2 for 2 units of product A and 3 units of product B''), and a budget. Every period, the LLM agent proposes a purchase plan, and receives as feedback the quality of that purchase plan (determined by a simple, but unknown to the LLM agent, mathematical formula). The LLM agent's goal is to identify the best purchase plan within the budget. 

\paragraph{Environment.} There are $n$ products $A := \{a_1, \dots, a_n \}$ with effectiveness scores $e_1, \dots, e_n \in \mathbb{N}$. The products are partitioned into $k$ categories $A := A_1 \sqcup \dots \sqcup A_k$ (where $\sqcup$ denotes disjoint union). 

Given quantities purchased of each product $(z_1, \dots, z_n) \in \mathbb{Z}_{\ge 0}^n$, the quantity of workers supported by these products is given by \[ f(z_1, \dots, z_n) := \textstyle\prod_{i=1}^k \left( \sum_{a_{j} \in A_i} e_j  z_j  \right)^{1/k}.\]
Thus, products within the same category are substitutes, and products across different categories are complements. 

Products can be purchased through \emph{deals}. There are three types of deals: \textit{simple} (a bundle of products is assigned a per-copy price), \textit{bulk only} (like simple, but requires purchasing at least some minimum number of copies), and \textit{two-part tariff} (like simple, but in addition to the per-copy price there is also an upfront cost for the deal that is independent of the number of copies purchased). For further details see \cref{app:procurement}.

\paragraph{Task.} The LLM agent is given a budget $B >0$ and a menu consisting of $m$ deals. It is asked to find the purchase plan of deals that maximizes the quantity of workers supported within the budget. 

\paragraph{Tools.} The LLM agent has access to the following tools: get\_previous\_purchase\_data, get\_equipment\_information, get\_budget, get\_attempt\_number, submit\_purchase\_plan. For further details see \cref{app:prompts_procurement}.

\paragraph{Feedback.} In each period, the LLM agent may propose a purchase plan. If the purchase plan exceeds the budget, the agent is informed that the plan is not feasible. Otherwise, the agent receives feedback on the quantity of workers supported by that purchase plan.

\paragraph{Key Unknowns.} The LLM agent is not given the effectiveness scores $e_1, \dots, e_n \in \mathbb{R}$, and must learn information about these weights indirectly from the feedback.

\paragraph{Instantiation.} We set $n\!=\!12$ and $k\!=\!3$ for \textsc{Basic}, $n\!=\!30$ and $k\!=\!5$ for \textsc{Medium}, and $n\!=\!100$ and $k\!=\!10$ for \textsc{Hard}. The effectiveness scores $e_1, \dots, e_n$ are sampled uniformly from $\{1,2,3\}$ for \textsc{Basic}, $\{1,2,\dots,5\}$ for \textsc{Medium}, and $\{1,2,\dots,20\}$ for \textsc{Hard}. For each difficulty level we set the menu size $m := n$ and we use equal category sizes $|A_1| = \dots = |A_k| = n/k$. For details of menu generation see \cref{app:procurement}.

\paragraph{Success Metric.} Each experimental run is scored based on the quantity of workers supported by the best purchase plan the LLM agent proposed, normalized by the quantity of workers supported by the optimal purchase plan within budget $B$: 
\[ \frac{f(\text{LLM's quantities purchased of each product})}{\textsf{OPT}}.\]

\subsubsection{Scheduling}\label{scheduling:specification}

\paragraph{High-level overview.} The LLM agent is given a list of workers and tasks. The workers have preferences over the tasks, and the tasks have ``preferences'' over the workers (e.g., determined by how suitable a worker is for that task), but the LLM agent is not explicitly told any of these preferences. Every period, the LLM agent proposes an assignment of workers to tasks, and receives as feedback one or more ``problems'' (blocking pairs) with that assignment. The LLM agent's goal is to identify an assignment with no, or as few as possible, ``problems.''

\paragraph{Environment.} There are $n$ workers $W := \{w_1, \dots, w_n \}$ and $n$ tasks $T := \{t_1, \dots, t_n \}$. Each worker $w_i$ has a complete strict preference order $\succ_{w_i}$ over tasks, and each task~$t_i$ has a complete strict preference order $\succ_{t_i}$ over workers. 

\paragraph{Task.} The LLM agent is asked to find a \textit{(perfect) matching} that is \textit{stable}. A \emph{matching} is a bijection $\mu : W \to T$. A worker-task pair $(w,t) \in W \times T$ is a \textit{blocking pair} for a matching $\mu$ if $t \succ_w \mu(w)$ and $w \succ_t \mu(t)$, that is, $w$ and $t$ each prefer the other over their match in the matching. A matching is \emph{stable} if it has no blocking pairs. The existence of a stable matching is guaranteed by \citet{gale_college_1962}. 

\paragraph{Tools.} The LLM agent has access to the following tools: get\_previous\_attempts\_data, get\_worker\_ids, get\_task\_ids, get\_attempt\_number, submit\_assignment. For details about the precise functionality of these tools see \cref{app:prompts_scheduling}.

\paragraph{Feedback.} In each period, the LLM agent may propose a matching. If the matching is stable, the experiment ends. Otherwise, the agent receives feedback in the form of $k$ randomly chosen blocking pairs (or all blocking pairs, if there are fewer than~$k$).\footnote{A stable matching can be computed in polynomial time based on this input, even if only one, adversarially chosen, blocking pair is returned \citep{bei_complexity_2013,emamjomeh-zadeh_complexity_2020}.}

\paragraph{Key Unknowns.} The LLM agent is not given the preferences of the tasks and workers $\succ_{w_i}$ and $\succ_{t_i}$, and must learn information about these preferences indirectly from the blocking-pair feedback. 

\paragraph{Instantiation.} We set $n\!=\!10$ and $k\!=\!1$ for \textsc{Basic}, $n\!=\!20$ and $k\!=\!2$ for \textsc{Medium}, and $n\!=\!50$ and $k\!=\!5$ for \textsc{Hard}. For each difficulty level, we randomly generate the preferences of the workers and tasks using the public scores model \citep{ashlagi_welfare_2023}. For details of preference generation see \cref{app:scheduling}. 

\paragraph{Success Metric.} Each experimental run is scored based on the quality of the final matching the LLM agent proposes,\footnote{In the final period, the following additional instruction is included in the LLM prompt: \textit{``**This is your final attempt.** This time, you should submit the highest quality assignment possible, that has the fewest problems.''} This ensures that the LLM agent is evaluated based on a matching for which it was instructed to minimize the number of blocking pairs (mitigating the risk that it uses the final period to explore).} according to the following formula: 
\[ 1 - \frac{\text{\# blocking pairs in agent's final matching}}{\mathbb{E}_{\text{unif.~random matching $\mu$}}[\text{\# blocking pairs in $\mu$}]}.\]
Note that the formula allows for negative scores if the LLM agent proposes a final matching that is worse than the uniform random baseline. 

\subsubsection{Pricing}\label{pricing:specification}

\paragraph{High-level overview.} The LLM agent is given a list of products. Every period, the LLM agent sets prices for those products, and receives as feedback the quantity sold and profit earned from each product (determined by a simple, but unknown to the LLM agent, mathematical formula). The LLM agent's goal is to set prices in a way that maximizes profits. Moreover, the market conditions change according to a predictable pattern, and to price optimally, the LLM agent must anticipate this pattern and price accordingly (e.g., learn to steadily increase or decrease prices).

\paragraph{Environment.} There are $n$ products $G := \{g_1, \dots, g_n \}$ partitioned into $k$ categories $G := G_1 \sqcup \dots \sqcup G_k$ (where $\sqcup$ denotes disjoint union). Given prices $p_1, \dots, p_n$, the quantity demanded $q_i$ for the $i$th product $g_i$ in the $j$th category~$G_j$ is given by a nested logit demand model \citep{berry_estimating_1994}:
\[ q_i := M \frac{\exp(\frac{a_i - p_i/\alpha_i}{1-\sigma})}{D_j} \cdot \frac{D_j^{1-\sigma}}{\exp(\frac{a_0}{1-\sigma}) + \sum_{j' \in [k]} D_{j'}^{(1-\sigma)}},\]
where $D_{j'} := \sum_{g_k \in G_{j'}} \exp(\frac{a_k - p_k/\alpha_k}{1-\sigma})$ for $j' \in [k]$. Here, $a_i$ is the quality of product~$g_i$ (higher is better), $a_0$ is the quality of an outside option (higher means outside option more attractive), $\alpha_i$ determines the price sensitivity, $D_j$ is the market share of category~$G_j$, $\sigma$ is the elasticity of substitution, and $M$ scales overall market share.

Given costs $c_1, \dots, c_n$ of the products, the profit from good $g_i$ is $\pi_i := (p_i/\alpha_i - c_i)q_i$. The total profit is $\pi := \sum_{i=1}^n \pi_i$. 

To make this pricing environment non-stationary, we vary the $\{\alpha_i\}_{i=1}^n$ parameters between periods, according to a predictable pattern that the LLM must learn. We consider two kinds of patterns: \textit{linear shifts}, in which each $\alpha_i$ is increased or decreased by a constant step size in each period (the step sizes differ between products $i \in [n]$), and \textit{periodic shifts}, in which each $\alpha_i$ varies according to a sinusoidal pattern (the frequency and phase are the same for all products $i \in [n]$, but the amplitudes may differ). 

\paragraph{Task.} The LLM agent is asked to set prices for the $n$ products in a way that maximizes total profit~$\pi$. 

\paragraph{Tools.} The LLM agent has access to the following tools: get\_previous\_pricing\_data, get\_product\_ids, get\_attempt\_number, set\_prices. For details about the precise functionality of these tools, see \cref{app:prompts_pricing}.

\paragraph{Feedback.} At the end of each period, the LLM agent sets prices for the $n$ products. In the following period, the LLM agent is given as feedback the quantity sold and profit earned for each product, as well as total profit. 

\paragraph{Key Unknowns.} The LLM agent is not given the parameters $\{a_i\}_{i=1}^n, \{\alpha_i \}_{i=1}^n, a_0, \sigma, M$ that characterize the demand response (nor how they evolve, where applicable), and must learn information about these parameters indirectly from the feedback. 

\paragraph{Instantiation.} To scale the difficulty, we scale the number of products. We set $n = 1$ for \textsc{Basic}, $n=4$ for \textsc{Medium}, and $n=10$ for \textsc{Hard}. Across all difficulty levels, we set $\sigma = 0.5$ and $M = 100$. We sample the costs $c_i \sim \textsf{Unif}([1,10])$ and qualities $a_i \sim \textsf{Unif}([2,3])$ independently. For each product $i \in [n]$, its category membership is determined by sampling from a (right-)truncated geometric distribution $\textsf{Geom}(0.2)$. To make the pricing environment non-stationary, we vary the $\{\alpha_i\}_{i=1}^n$ parameters with time according to a predictable pattern (either linear shifts or periodic shifts). For further details see \cref{app:pricing}. 

\paragraph{Success Metric.} Each experimental run is scored based on the total profit earned in the last 50 periods, normalized by the total profit that would have been earned from pricing optimally in those periods:
\[ \frac{\text{total profit $\pi$ from last 50 periods}}{\textsf{OPT}}.\]

\section{Benchmark Results}\label{sec:benchmark-results}

In this section, we evaluate an array of LLMs on the three EconEvals benchmarks (procurement, scheduling, and pricing). We pursue three interrelated goals: (1)~measuring how LLMs perform on the benchmarks, (2)~gaining economic insights from behavioral features underlying that performance, and (3)~validating key aspects of our benchmark design.

Importantly, we view our main contribution as methodological. Our benchmarks share several key design features that, we argue, enable richer analysis than traditional Q\&A-style evaluations. First, the difficulty of each benchmark task lies (partly) in that \textbf{the LLM agent must learn from the environment in context}---e.g.~in scheduling, the preferences of the workers and tasks are not given to the LLM agent, and can only be learned via deliberate exploration. Second, because each environment is synthetically generated according to an underlying economic model, our benchmarks enjoy two additional properties: \textbf{scalable difficulty} and \textbf{(economically) grounded evaluation}. Scalable difficulty allows for adjusting the difficulty of the benchmark to suit the LLMs being compared. In this work, we instantiate each environment at three difficulty levels---\textsc{Basic}, \textsc{Medium}, and \textsc{Hard}---but it is possible to generate instances at arbitrary difficulty levels. Grounded evaluation means that performance can be measured using objective metrics derived directly from the economic model. For example, in procurement, we can solve for the utility-maximizing purchase plan, and score any other purchase plan relative to this optimum. Other multi-turn benchmarks---such as those based on games or software tasks---typically rely on coarse evaluation: binary success/failure after a long episode, or proxy metrics with unclear relationships to the ultimate objective. Our economic environments, by contrast, yield fine-grained metrics at every step that have a direct, economically meaningful interpretation.

Accordingly, we propose that evaluation methods sharing these features enable deeper insight into LLM economic decision-making compared to traditional approaches. The results presented here serve as a proof-of-concept, demonstrating the kinds of analyses and insights that become possible with benchmarks like ours.

In \cref{sec:benchmark-data-collection}, we describe the data collection process. In \cref{sec:analysis-of-benchmark-scores}, we analyze the benchmark scores. In \cref{subsec:capability_over_time}, we find that more recently-released LLMs generally earn higher benchmark scores, indicating that (for the LLMs we test) the capabilities of LLMs in economic decision-making contexts are improving with time. In \cref{subsec:difficulty-scaling}, we validate our difficulty scaling approach. In \cref{subsec:inter-llm-comparisons}, we present further results comparing LLM performance across benchmarks. Finally in \cref{subsec:economic_analysis}, we demonstrate that economic insights can be uncovered by analyzing the behavior of the LLMs beyond their overall scores. We do so for each of the three benchmarks using two techniques: first, analyzing carefully-chosen trajectory-level metrics to quantify learning and exploration, and second, analyzing the LLM-written notes to understand other aspects of the LLMs' (stated) strategies.

\subsection{Benchmark Data Collection}\label{sec:benchmark-data-collection}

We measure the performance of an array of LLMs released between May 2024 and November 2025 on the EconEvals benchmarks. Specifically, we test Claude-3.5 Sonnet, Gemini 1.5 Pro, GPT-4o, GPT-4.1, o4-mini, Gemini 2.5 Pro, GPT-5, and Gemini 3 Pro Preview. For each LLM, we measure its performance on the three EconEvals benchmark environments (procurement, scheduling, and pricing) at each of the three difficulty levels (\textsc{Basic}, \textsc{Medium}, \textsc{Hard}). For each environment and difficulty level, we randomly generate 12 instances and run all LLM agents for 100 periods on the same instances.

For additional experimental details, including information on data collection timeframes and LLM parameter settings, see \cref{app:experimental_details}. For the raw benchmark results, including a table of all benchmark scores (\cref{tab:benchmark_results}) and visualizations of per-instance results (\cref{fig:benchmark_scores_swarmplot}), see \cref{app:benchmark_results}.

\subsection{Analysis of Benchmark Scores}\label{sec:analysis-of-benchmark-scores}

\subsubsection{Capability Improvement Over Time}\label{subsec:capability_over_time}

\begin{figure}
    \centering
    \includegraphics[width=\linewidth]{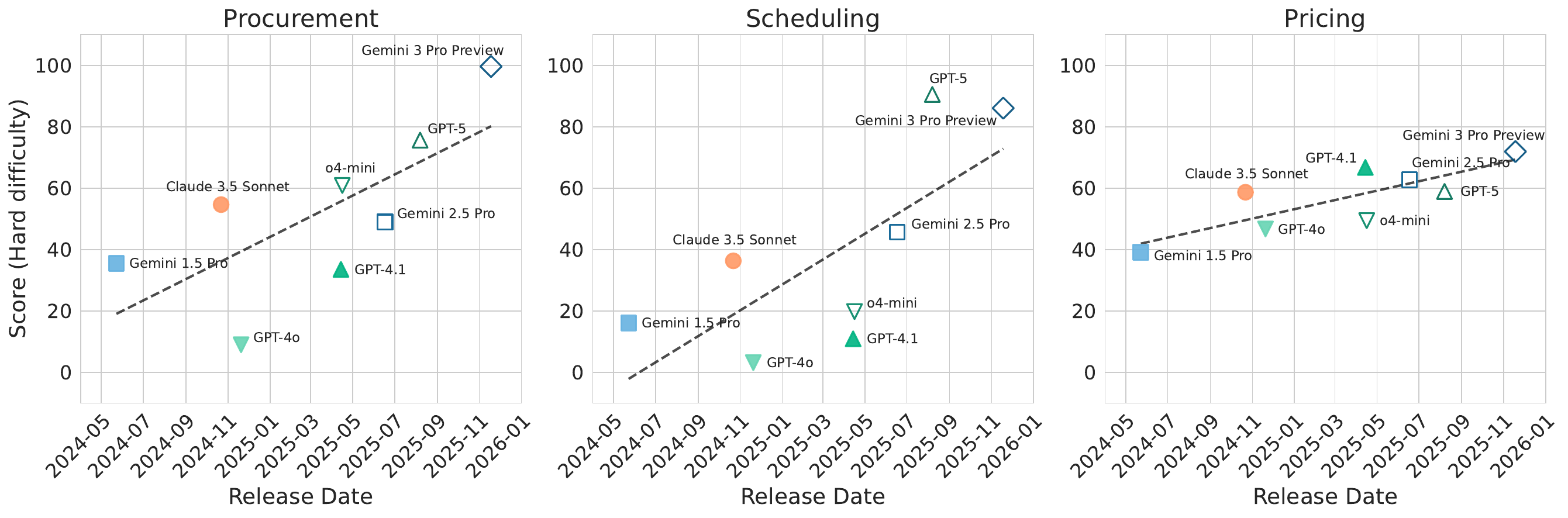}
    \caption{Benchmark scores of Claude 3.5 Sonnet, Gemini 1.5 Pro, GPT-4o, GPT-4.1, o4-mini, Gemini 2.5 Pro, GPT-5, and Gemini 3 Pro Preview on the three EconEvals benchmark environments---procurement, scheduling, and pricing---at the \textsc{Hard} difficulty level, by LLM release date. Each point represents the average score over 12 randomly generated instances. The highest possible score is 100. The dashed lines represent OLS linear regression fits. More recently-released LLMs generally earn higher benchmark scores, indicating that the capabilities of LLMs in economic decision-making contexts are improving with time.}
    \label{fig:capability_over_time}
\end{figure}

In this section, we examine the relationship between LLM benchmark performance and LLM release date. \cref{fig:capability_over_time} displays the \textsc{Hard} benchmark scores of each LLM on each environment, by LLM release date. \textbf{We observe that more recently-released LLMs generally earn higher benchmark scores, indicating that the capabilities of LLMs in economic decision-making contexts are improving with time.}\footnote{See also \cite{measuring-ai-ability-to-complete-long-tasks}, which observes a similar trend in software tasks.}

To further quantify this relationship, for each of the three benchmark environments, we run a OLS linear regression with robust standard errors clustered by LLM. For all three benchmark environments, we find a positive slope in the relationship between \textsc{Hard} benchmark score and release date ($p < 0.01$). For procurement, the slope corresponds to an average score increase of 41 percentage points per year; for scheduling, 50 percentage points per year; and for pricing, 18 percentage points per year. See \cref{app:capability_over_time} for further details, including detailed regression results and a similar analysis for \textsc{Basic} and \textsc{Medium}.

\subsubsection{Difficulty Scaling Validation}\label{subsec:difficulty-scaling}

\begin{figure}
    \centering
    \includegraphics[width=\linewidth]{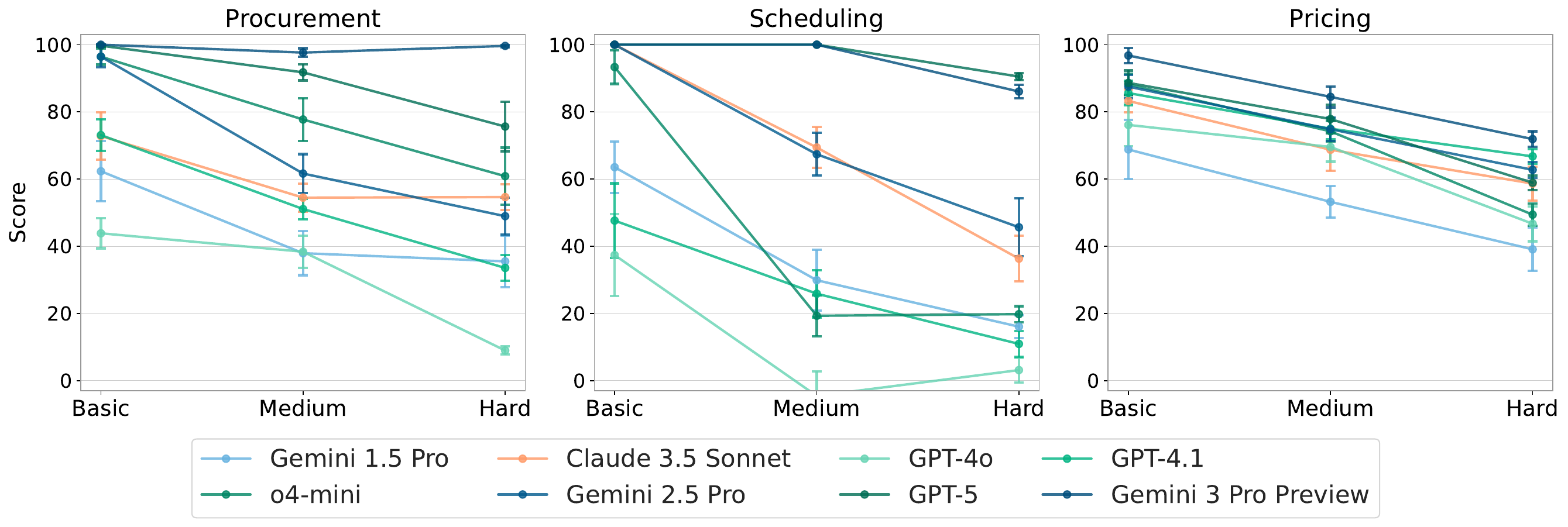}
    \caption{Benchmark scores of Claude 3.5 Sonnet, Gemini 1.5 Pro, GPT-4o, GPT-4.1, o4-mini, Gemini 2.5 Pro, and Gemini 3 Pro Preview on the three EconEvals benchmark environments---procurement, scheduling, and pricing---at the \textsc{Basic}, \textsc{Medium}, and \textsc{Hard} difficulty levels. Each point represents the average score over 12 randomly generated instances; error bars represent standard errors of the mean across instances. The highest possible score is 100. More difficult instances generally lead to lower benchmark scores, indicating that our difficulty scaling technique is effective.}
    \label{fig:difficulty-scaling}
\end{figure}

In this section, we examine the relationship between LLM benchmark performance and benchmark difficulty level. \cref{fig:difficulty-scaling} displays the benchmark scores of each LLM on each environment, by difficulty level. \textbf{We observe that our approach for scaling the difficulty of benchmarks---namely, increasing the instance size---is effective.} For 23 out of 24 environment-LLM pairs, scores on \textsc{Hard} instances are lower than scores on \textsc{Basic} instances ($p < 0.05$, two-sided Welch's $t$-test). (The one exception is Gemini 3 Pro Preview on procurement, where the \textsc{Basic} score is 100.0 and the \textsc{Hard} score is 99.6. That said, Gemini 3 Pro Preview fully solves 12/12 \textsc{Basic} instances, compared to 8/12 \textsc{Hard} instances, indicating some degree of difficulty increase.) 

\subsubsection{Inter-LLM Comparisons}\label{subsec:inter-llm-comparisons}

In this section, guided by the benchmark score data alone, we conduct a preliminary comparison of the capabilities of the LLMs that we test. In \cref{subsec:economic_analysis}, we conduct a more in-depth analysis. 

\paragraph{Difficulty-pooled comparisons.} For each of the three benchmark environments and each pair of LLMs, we compare their benchmark scores across the 36 total instances (12 per difficulty level) using a paired Wilcoxon signed-rank test.

\begin{figure}[t]
  \centering
\includegraphics[width=\textwidth]{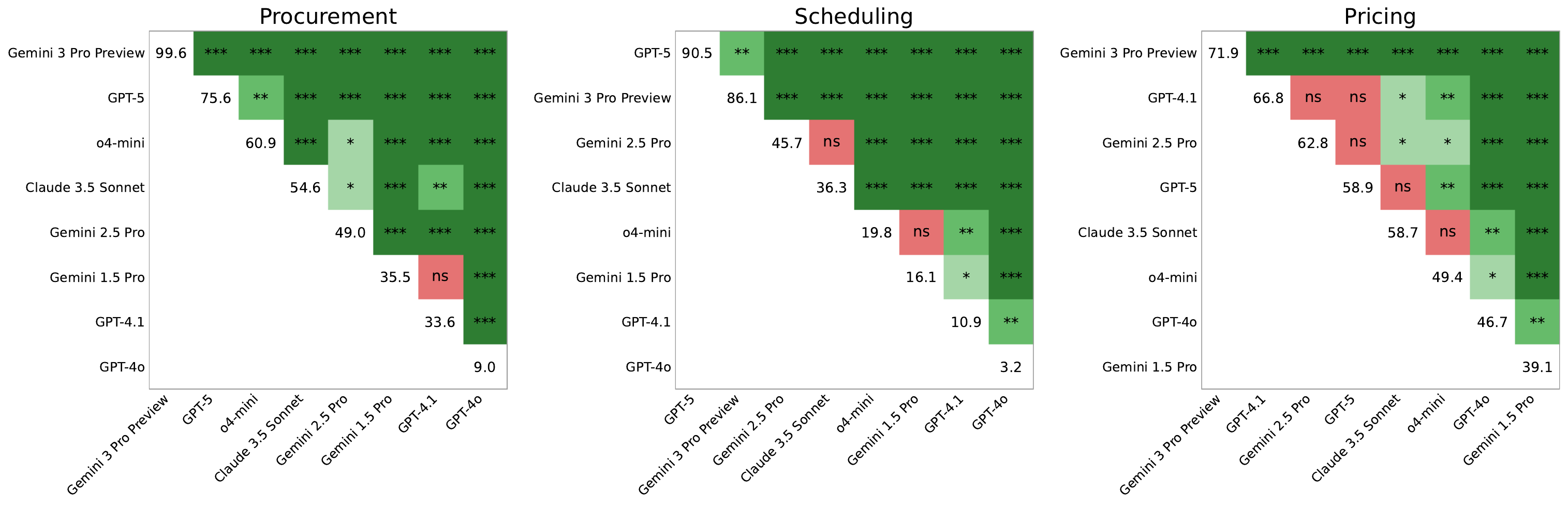}
\caption{Pairwise comparisons between LLM scores on the three benchmark environments. Each off-diagonal entry shows the result of a paired Wilcoxon signed-rank test comparing two LLMs across 36 matched instances (12 instances for each of the three difficulty levels), with significance levels: ***: $p < 0.01$, **: $p < 0.05$, *: $p < 0.1$. Diagonal entries display the mean score on \textsc{Hard} difficulty, which also determines the row/column ordering in each subplot.}
  \label{fig:inter-llm-comparisons}
\end{figure}

\cref{fig:inter-llm-comparisons} displays the results. In accordance with \cref{subsec:capability_over_time}, we observe that the two most recently-released LLMs, Gemini 3 Pro Preview and GPT-5, outperform older LLMs: Gemini 3 Pro Preview earns higher scores than all other LLMs at procurement and pricing ($p < 0.01$, Welch's two-sided $t$-test pooling all three difficulty levels, run separately for each LLM), and GPT-5 earns higher scores than all other LLMs at scheduling ($p < 0.05$, Welch's two-sided $t$-test pooling all three difficulty levels, run separately for each LLM).

Looking beyond the top-scoring LLMs, \textbf{we observe that different LLMs exhibit varying strengths across the benchmark environments}. For example, in procurement, the reasoning model o4-mini scores third-best, compared to fifth-best in scheduling and sixth-best in pricing. In \cref{subsec:procurement_economic_analysis}, we attribute this to the fact that procurement is more calculation-heavy than the other two benchmarks, and reasoning models excel at calculations compared to non-reasoning models. 

\paragraph{Comparisons disaggregated by difficulty.} Additionally, we conduct comparisons disaggregated by difficulty level: For each benchmark environment, difficulty level, and pair of LLMs, we compare their benchmark scores across the 12 total instances using a paired Wilcoxon signed-rank test. The results are displayed in \cref{fig:inter-llm-comparisons-disaggregated} in \cref{app:inter_llm_comparisons_scores}. \textbf{We observe that different difficulty levels are best-suited for comparing different pairs of LLMs.} For example, in scheduling, o4-mini decisively outperforms Gemini 1.5 Pro, GPT-4.1, and GPT-4o at \textsc{Basic}, but these comparisons become less clear-cut (not always statistically significant) at \textsc{Medium} and \textsc{Hard}. Conversely, \textsc{Basic} cannot differentiate among the top-performing LLMs---Gemini 3 Pro Preview, GPT-5, and Gemini 2.5 Pro all achieve near-perfect scores---whereas \textsc{Hard} effectively separates them.

\begin{figure}[t]
    \centering
    \includegraphics[width=\textwidth]{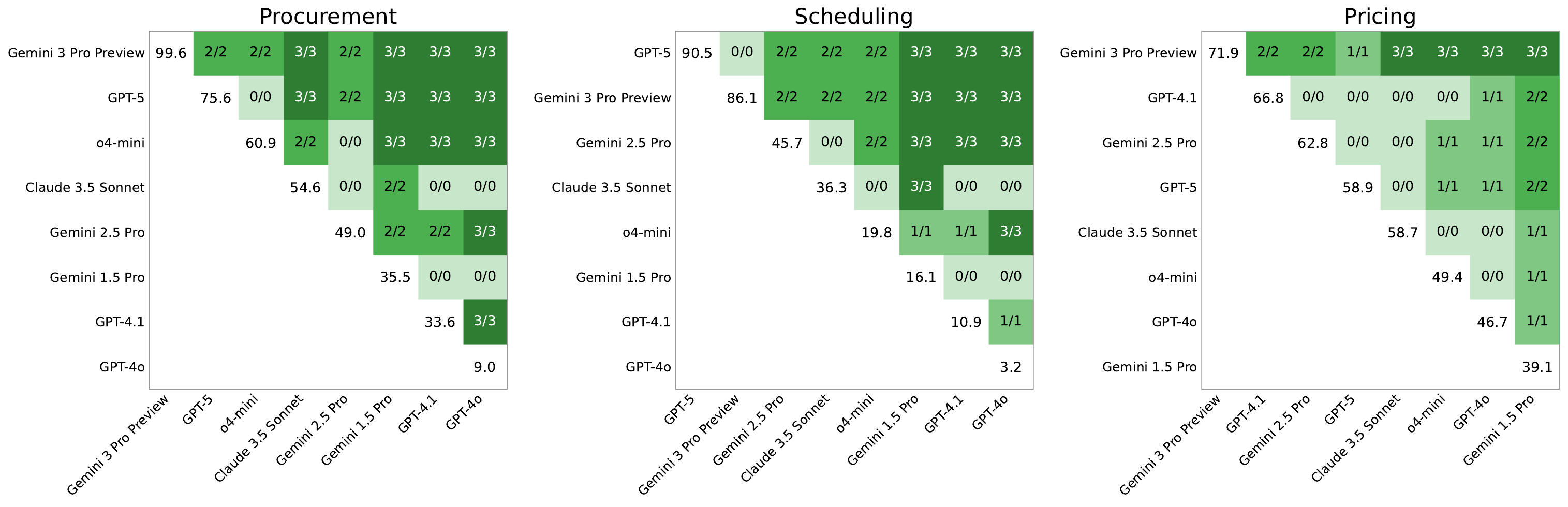}
    \caption{Consistency of pairwise LLM comparisons across difficulty levels, for each of the three benchmark environments. A cell labeled $X/Y$ denotes: $Y$ (out of the three difficulty levels) comparisons are statistically significant, and $X$ of these go in the majority direction. (For example, if model $A$ outperformed model $B$ at \textsc{Basic} and \textsc{Medium}, but $B$ outperformed $A$ at \textsc{Hard}, this would be labeled 2/3.) All cells are labeled 0/0, 1/1, 2/2, or 3/3, indicating that comparisons are consistent whenever they are statistically significant (at the $p < 0.05$ level).}
    \label{fig:benchmark_cross_difficulty_wins}
\end{figure}

A natural question is whether comparisons across difficulty levels are consistent. For example, model $A$ could outperform model $B$ at \textsc{Basic} and \textsc{Medium}, yet $B$ could outperform $A$ at \textsc{Hard}. \cref{fig:benchmark_cross_difficulty_wins} displays consistency checks for each benchmark and each pair of LLMs, across all difficulty levels. We observe no inconsistencies in relative model performance across difficulty levels for statistically significant comparisons.

\subsection{Analysis of and Economic Insights from Benchmark Trajectories}\label{subsec:economic_analysis}

In this section, we leverage some of the unique design aspects of EconEvals to further analyze the benchmark performance of the LLMs. Recall that an LLM's benchmark score on a given instance summarizes a rich trajectory in which the LLM, over the course of 100 periods, repeatedly takes actions and writes notes about said actions. ``Unpacking'' this trajectory and analyzing carefully-chosen aspects of it allows us to gain a more in-depth understanding of the capabilities of the LLMs that we test.

We guide our exploration with a large-scale text analysis (\cref{subsec:clustering-methods}), following the methods of \cite{fish_algorithmic_2024}. For each of the three benchmark environments, we split all notes written by all eight LLMs using the \texttt{write\_notes} tool into paragraphs, and group these paragraphs into 80 clusters using standard techniques. For each cluster, we determine which LLM is most dominant, normalized by the LLM's overall verbosity. This gives a high-level overview of the kinds of reasoning each LLM employs. 

We then use this notes-based characterization of each LLM's behavior as a jumping-off point to analyze more precisely what aspects of the LLMs' behavior are associated with differences in overall benchmark performance. For pricing (\cref{subsec:pricing_economic_analysis}), we observe that the best-performing LLM's advantage is driven by its unique ability to detect and respond to changes in the market conditions (i.e., linear or periodic shifts, see \cref{pricing:specification}). For scheduling (\cref{subsec:scheduling_economic_analysis}), we observe that better-performing LLMs tend to reason about the unknown preferences of the workers and tasks in more sophisticated ways (specifically, by reasoning about longer preference lists). For procurement (\cref{subsec:procurement_economic_analysis}), we observe that reasoning LLMs consistently (weakly) outperform non-reasoning LLMs, and that this performance gap aligns with the reasoning LLMs' superior budget ``bin-packing'' skills. 

\subsubsection{Clustering Methods}\label{subsec:clustering-methods}

We begin by describing the text analysis methods used to guide the analyses in \cref{subsec:pricing_economic_analysis,subsec:procurement_economic_analysis,subsec:scheduling_economic_analysis}. We follow the blueprint of \cite{fish_algorithmic_2024}, with slight adaptations. 

For each of the three benchmark environments, we collect all notes written by all LLMs using the \texttt{write\_notes} tool, and split each notes entry into individual paragraphs (i.e., text chunks separated by at least two consecutive newlines).\footnote{
  We split the LLM-written notes into paragraphs, rather than sentences (as in \citealt{fish_algorithmic_2024}), because many of the LLMs we test, particularly the more recently released LLMs, are more likely to write in shorthand, such that individual sentences are difficult to interpret without broader context. 
} This results in 394,583 paragraphs for procurement, 48,526 paragraphs for scheduling, and 70,231 paragraphs for pricing. We then use OpenAI's embedding model text-embedding-3-large to convert each paragraph to a 3,072-dimensional vector. Finally, we perform a dimensionality reduction using PCA to reduce each paragraph to a 20-dimensional vector, and then cluster the resulting set of vectors into 80 clusters using $k$-means.\footnote{\label{ftn:num_clusters}The analysis of \cite{fish_algorithmic_2024} compares two different LLM prompts and uses 20 clusters. Accordingly, we set the number of clusters equal to $10 \cdot \text{(\# LLMs compared in analysis)}$.}

For each cluster, we compute the proportion of paragraphs arising from each LLM's notes, normalized by the LLM's overall ``verbosity'' (the total number of paragraphs written by that LLM, across all clusters), to determine which LLM is most dominant. That is, the dominant LLM for a given cluster is the LLM that maximizes the quantity $\frac{\text{\# paragraphs from LLM in cluster}}{\text{\# paragraphs from LLM overall}}$. Then, for each LLM, we examine the clusters for which it is dominant, to understand on a high level the content of its notes (and in particular, how its notes differ from the notes of other LLMs). 

Finally, also following \cite{fish_algorithmic_2024}, for each cluster, we determine the 10 closest-to-center paragraphs and use Gemini 3 Flash to generate a short cluster description, which we use for labeling purposes.

In certain cases, rather than clustering the notes of all eight LLMs simultaneously, we find it illustrative to cluster the notes of a specifically chosen subset of LLMs, to allow for a direct comparison. Here, we use the same clustering procedure as above, except with $10 \cdot (\text{\# LLMs compared in analysis})$ clusters instead of 80.\footnote{See \cref{ftn:num_clusters}.}

\paragraph{Verbosity rebalancing robustness check.} Unlike \cite{fish_algorithmic_2024}, in which the verbosity values were very similar for the items being compared, the LLMs we test exhibit substantially varying levels of verbosity. For example, in pricing (the benchmark with the widest disparities in verbosity), the most verbose LLM (Claude 3.5 Sonnet) contributed 6.7 times as many paragraphs as the least verbose LLM (Gemini 1.5 Pro). To verify that these varying levels of verbosity do not substantially affect the outputs of the clustering analysis, as a robustness check, we run an additional ``balanced'' clustering analysis, in which we first subsample uniformly without replacement the same number of paragraphs from each LLM and cluster only these paragraphs. (The number of paragraphs subsampled is set to be the number of paragraphs contributed by the least verbose LLM.) For all three benchmark environments, we indeed observe that the outputs of this ``balanced'' clustering analysis are similar to the results of the standard clustering analysis.

\subsubsection{Pricing Analysis}\label{subsec:pricing_economic_analysis}

\begin{figure}[h!]
  \includegraphics[width=0.49\textwidth]{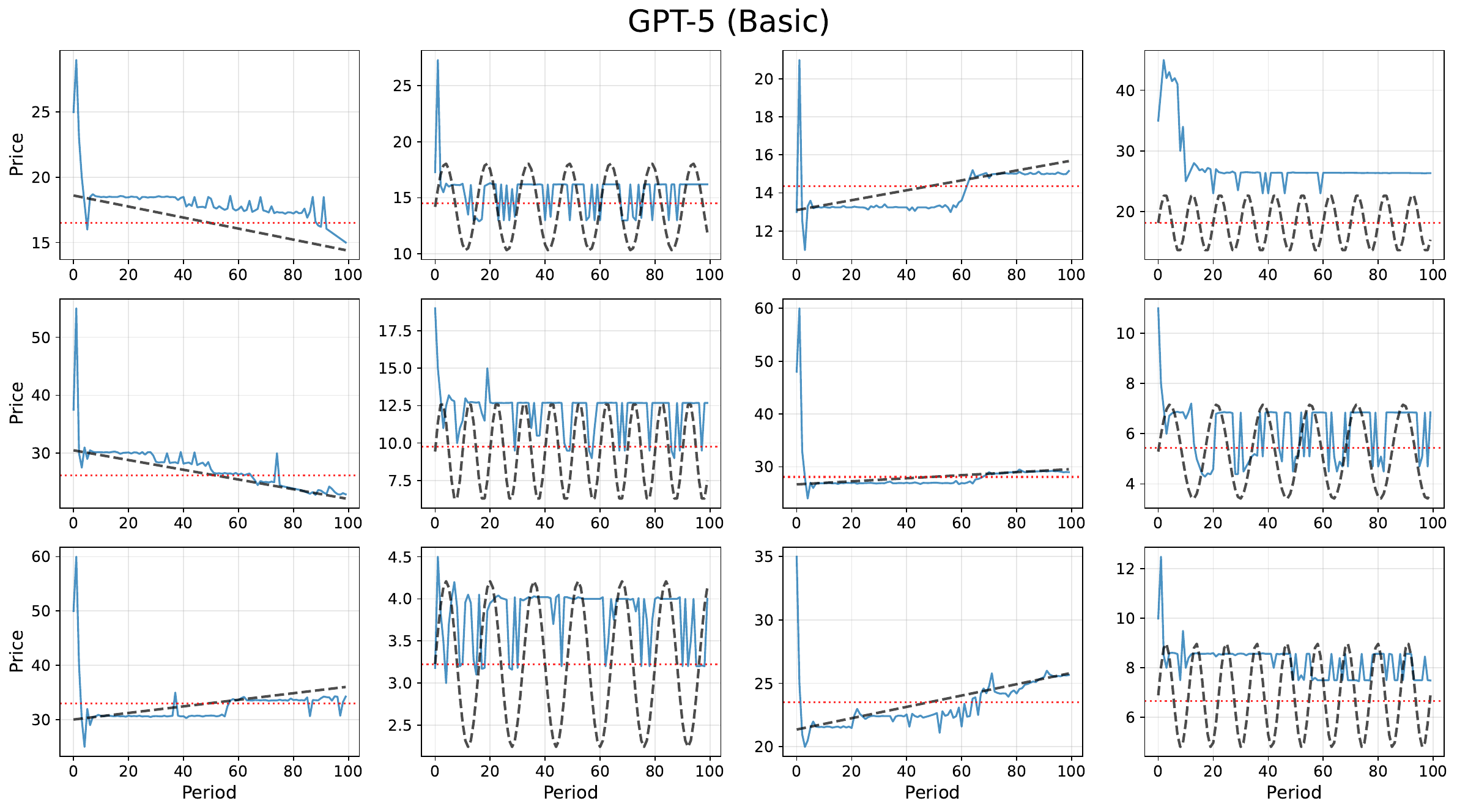}
  \includegraphics[width=0.49\textwidth]{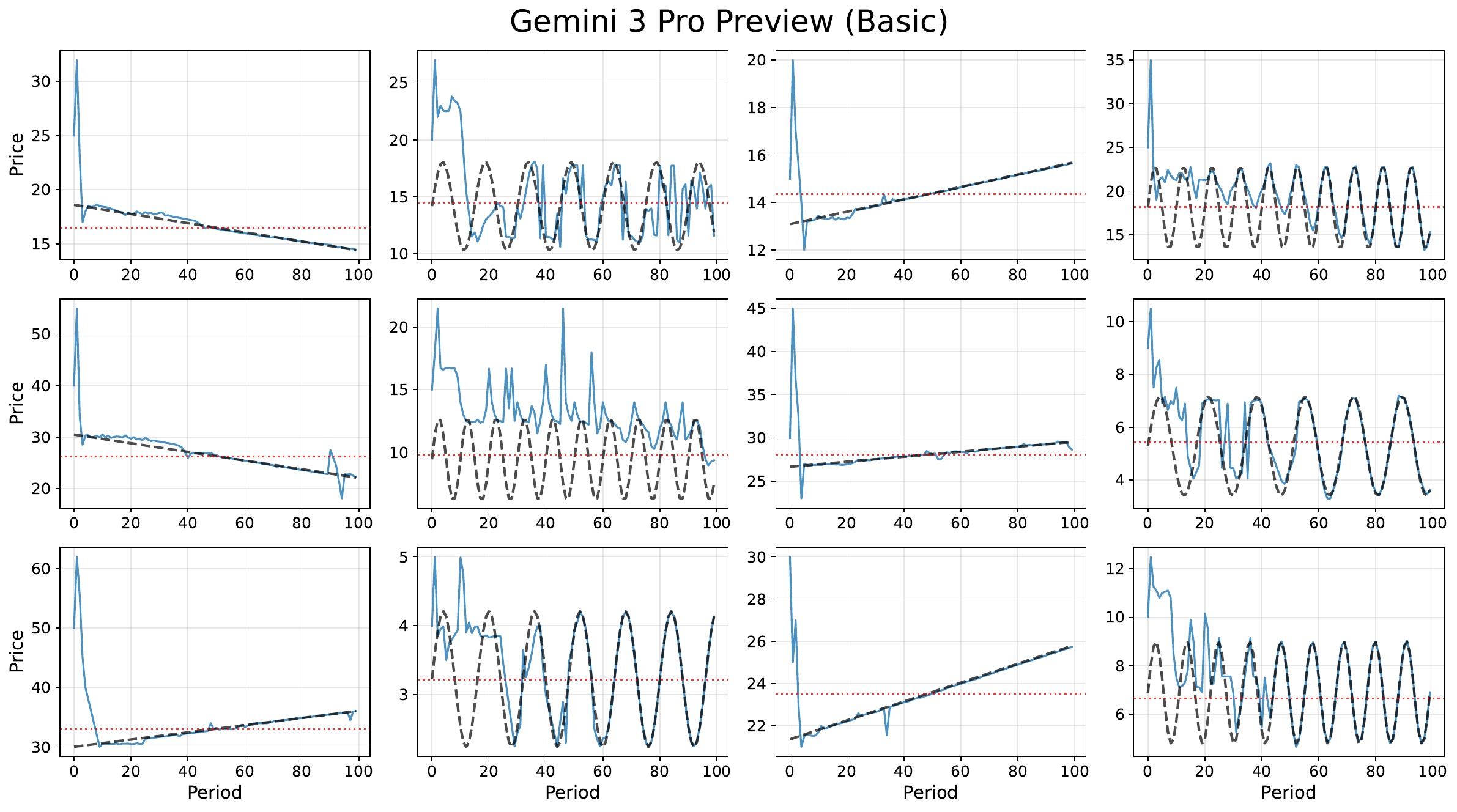}
  \caption{Pricing behavior of GPT-5 (left) and Gemini 3 Pro Preview (right) on all 12 \textsc{Basic} instances. The solid blue line shows the prices set by the LLM, the dashed black line shows the profit-maximizing price for that period, and the dotted red line shows the profit-maximizing price over all periods.}
  \label{fig:pricing_comparison}
\end{figure}

As a warm-up, we analyze the pricing benchmark. Recall from \cref{subsec:inter-llm-comparisons} that Gemini 3 Pro Preview is the best-performing LLM, followed by a group of LLMs with similar performance (GPT-4.1, Gemini 2.5 Pro, GPT-5, and Claude 3.5 Sonnet), followed by the weakest two LLMs (GPT-4o, Gemini 1.5 Pro). In this section, we aim to understand potential reasons behind Gemini 3 Pro Preview's outsized advantage. 

In the case of pricing, a clear reason is already apparent in the raw data. \cref{fig:pricing_comparison} displays the pricing behavior of Gemini 3 Pro Preview and GPT-5 (the second-strongest LLM in \textsc{Basic} pricing) in all 12 \textsc{Basic} instances. In 10/12 cases, Gemini 3 Pro Preview is able to adjust its price in a manner that closely anticipates the changing market conditions, resulting in near-optimal pricing performance. By contrast, GPT-5 adapts its prices far less effectively. 

A head-to-head clustering analysis comparing the \textsc{Basic} notes of Gemini 3 Pro Preview and GPT-5 confirms this observation. Clusters for which Gemini 3 Pro Preview is dominant include clusters estimating ``drift rate'' (likely referring to linear shifts) and ``cycle index'' (likely referring to periodic shifts), whereas clusters for which GPT-5 is dominant focus more on ``local exploration'', and do not as explicitly or consistently mention the changing market conditions. See \cref{subsec:pricing_full_basic_gemini_3_pro_preview_vs_gpt_5_2025_08_07} for the full list of clusters.

This serves as a reasonableness check for our methodology: Here, differences in LLMs' notes directly correspond to differences in LLMs' pricing behavior. In the following sections on scheduling (\cref{subsec:scheduling_economic_analysis}) and procurement (\cref{subsec:procurement_economic_analysis}), the precise behavior of interest will in some cases not be so easily visualizable (let alone quantifiable), so we will rely more heavily on text analyses to draw conclusions about LLM behavior. Accordingly, the validity of our text analysis hinges on the assumption that an LLM's notes faithfully describe the reasoning behind its actions.

\subsubsection{Scheduling Analysis}\label{subsec:scheduling_economic_analysis}

We next turn to an analysis of the LLMs' approach to the scheduling benchmark. Recall from \cref{subsec:inter-llm-comparisons} that GPT-5 and Gemini 3 Pro Preview are the best-performing LLMs (with GPT-5 having a slight advantage in \textsc{Hard}), followed by Gemini 2.5 Pro and Claude 3.5 Sonnet, followed by o4-mini and Gemini 1.5 Pro, followed by the weakest two LLMs (GPT-4.1, GPT-4o). In this section, we investigate two potential reasons behind these performance differences.\footnote{Additionally, for a comparison with a greedy baseline, see \cref{scheduling_baseline}.}

\begin{figure}[t]
\centering 
  \includegraphics[width=\textwidth]{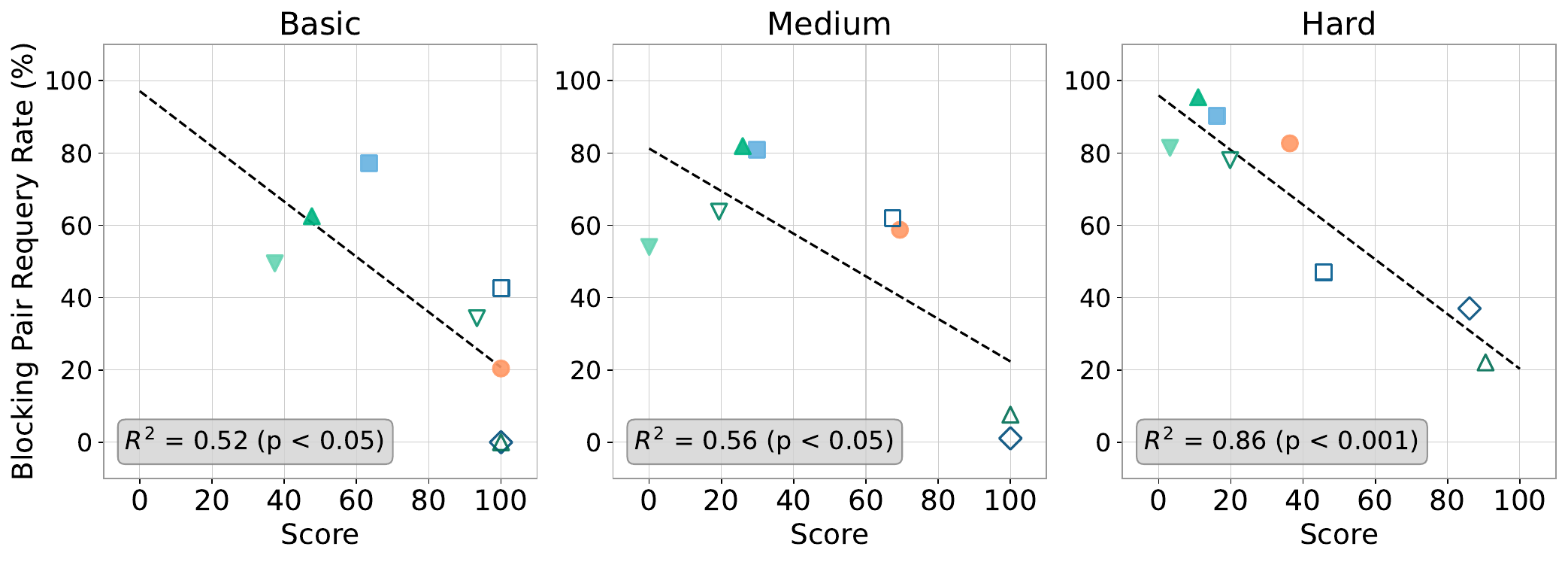}
  \includegraphics[width=0.8\textwidth]{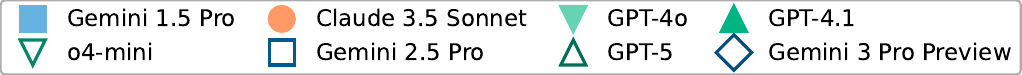}
  \caption{Relationship between the blocking pair requery rate and benchmark score in the scheduling task, for each LLM, by difficulty level. The blocking pair requery rate measures the proportion of proposed matchings whose blocking pairs include at least one blocking pair previously given as feedback. In each panel, $R^2$ and $p$-values are displayed for the linear fit. We observe, particularly on \textsc{Hard}, that lower blocking pair requery rates are associated with better benchmark performance.}
  \label{fig:scheduling_known_blocking_pair_requery_rate}
\end{figure}

\paragraph{Learning from environment feedback.} We begin by investigating the extent to which different LLMs learn from environment feedback. Recall that in scheduling, the LLM is repeatedly asked to propose a matching between workers and tasks (each of which has preferences over the other that are not provided to the LLM), and given one or more blocking pairs as feedback. One natural proxy metric for understanding the extent to which an LLM learns from this feedback is to look at the proportion of proposed matchings whose blocking pairs include at least one blocking pair previously given as feedback. 

For each LLM and each difficulty level, we compute this metric, aggregated across all 12 instances. The results are displayed in \cref{fig:scheduling_known_blocking_pair_requery_rate}. We observe, especially on \textsc{Hard}, that LLMs with lower rates of requerying known blocking pairs tend to earn higher benchmark scores (for \textsc{Hard}: $R^2=0.86$, $p < 0.01$). That is, better performance at the scheduling benchmark is associated with a greater ability to learn from environment feedback.

\paragraph{Reasoning about underlying preferences.} We next turn to investigating the \textit{process} by which the LLMs propose matchings. Specifically, we aim to understand the extent to which different LLMs reason about the underlying preferences of the workers and tasks, versus simply trying arbitrary matchings or making greedy adjustments to prior matchings. It is unclear how to reliably extract an LLM's ``beliefs'' about the underlying preferences from its proposed matchings alone, so we rely on the LLM-written notes.

We analyze the LLM-written notes using the clustering methods from \cref{subsec:clustering-methods} (see \cref{subsec:scheduling_full_8models} for the full list of clusters). Differences in the level of sophistication with which the LLMs reason about the underlying preferences are immediately apparent. For example, the closest-to-center paragraph in the cluster in which Gemini 3 Pro Preview is most dominant reads:
\begin{quote}\textit{
Reasoning for Attempt 13:
Based on previous attempts, I've deduced a stable-ish configuration. Key constraints identified:\\
- W1 prefers T10 > T4 > T1. T10 prefers W1 > W5. -> W1: T10.\\
- W5 prefers T10 > T5. But W1 beats W5 for T10. T5 prefers W20 > W5. W20 prefers T18 > T5. So W5 is safe at T5 if W20 is at T18. -> W5: T5.\\
- W20 prefers T18 > (T5?) > T17 > T6. -> W20: T18. [...]}
\end{quote}
By contrast, other LLMs (except GPT-5) are less likely to reason about the underlying preferences in such explicit detail. For example, among the 14 clusters for which Gemini 2.5 Pro is dominant, the closest-to-center paragraph of the cluster that discusses preferences in the greatest amount of detail reads:
\begin{quote}
\textit{
1.  **W18 -> T1**: W18 has a strong preference for T1.\\
2.  **W14 -> T4**: W14 has a strong preference for T4.\\
3.  **W2 -> T14**: Attempt 8 showed W2 prefers T14 over T9.[...]
}
\end{quote}

To quantify this observation, for each of the 80 clusters, we label each of the 10 closest-to-center paragraphs according to the following two criteria:
\begin{enumerate}\tightlist 
    \renewcommand{\theenumi}{\arabic{enumi}}
    \renewcommand{\labelenumi}{(\theenumi)}
  \item Does the paragraph explicitly reason about the underlying preferences of the workers and tasks? (Yes/No) 
  \item If so, what is the length of the longest preference list in the paragraph?\footnote{Here, ``W1>W2'' is labeled length 2, ``W1>W2>W3'' is labeled length 3, and so on. ``W1 prefers T1'' with no explicitly-named alternatives is labeled length 1.}
\end{enumerate}

A cluster is said to \textit{reason about preferences} if at least one of its 10 closest-to-center paragraphs is labeled ``Yes'' for (1). For each cluster that reasons about preferences, we compute the maximum length of the longest preference list across its 10 closest-to-center paragraphs. We find that 38 out of the 80 clusters reason about preferences, with six clusters having maximum preference list length 1, twenty-two length 2, and the remaining ten lengths 3, 4, 5, or 6. 

Next, for each LLM, we compute (i) the fraction of its paragraphs that belong to clusters that reason about preferences (\textit{``How often is the LLM reasoning about preferences?''}), and (ii) the average of the maximum preference list length values across clusters that reason about preferences, weighted by the fraction of the LLM's paragraphs in each such cluster (\textit{``When the LLM reasons about preferences, on average, what are the longest preference lists it considers?''}).

The results are displayed in \cref{tab:preference_reasoning}. We observe that a higher fraction of reasoning about preferences on its own does not necessarily translate to better performance: e.g., GPT-4.1 reasons about preferences the most frequently (tied with GPT-5), but performs relatively poorly. What appears to matter more is the \textit{length} of the preference lists considered: We observe that better-performing LLMs (at \textsc{Hard}) have longer average maximum preference list lengths ($R^2=0.81, p < 0.01$).\footnote{For \textsc{Basic} and \textsc{Medium}, we also observe this relationship, albeit slightly weaker ($R^2=0.59, p < 0.05$ and $R^2=0.66, p < 0.05$ respectively), possibly due to ceiling effects (for \textsc{Basic}, four LLMs achieve a perfect score, and for \textsc{Medium}, two LLMs achieve a perfect score).} That is, \textbf{better-performing LLMs tend to reason about more complex (i.e.,~longer) preference lists.}

\begin{table}[t]
    \centering
    \small
    \caption{Preference reasoning metrics for each LLM, derived from LLM-written notes on the scheduling benchmark, displayed alongside \textsc{Hard} scheduling benchmark scores for comparison. ``Reasons about Preferences (\%)'' and ``Average Maximum Preference List Length'' correspond to metrics (i) and (ii) respectively (see text).}
    \label{tab:preference_reasoning}
    \begin{tabular}{lccc}
        \toprule
        & Reasons about & Average Maximum & \textsc{Hard}  \\
        & Preferences (\%) & Preference List Length & Score \\
        \midrule
        GPT-5 & 73.5 & 3.31 & 90.5 \\
        \midrule
        Gemini 3 Pro Preview & 65.7 & 3.55 & 86.1 \\
        \midrule
        Gemini 2.5 Pro & 33.8 & 2.69 & 45.7 \\
        \midrule
        Claude 3.5 Sonnet & 53.9 & 2.22 & 36.3 \\
        \midrule
        o4-mini & 55.5 & 2.72 & 19.8 \\
        \midrule
        Gemini 1.5 Pro & 30.5 & 2.48 & 16.1 \\
        \midrule
        GPT-4.1 & 73.5 & 1.91 & 10.9 \\
        \midrule
        GPT-4o & 59.4 & 1.72 & 3.2 \\
        \bottomrule
    \end{tabular}
\end{table}

\subsubsection{Procurement Analysis}\label{subsec:procurement_economic_analysis}

In this section, we analyze the LLMs' approach to the procurement benchmark. Recall from \cref{subsec:inter-llm-comparisons} that Gemini 3 Pro Preview is the best-performing LLM, followed by GPT-5, followed by three LLMs with moderate performance (o4-mini, Claude 3.5 Sonnet, Gemini 2.5 Pro), followed by the weakest three LLMs (GPT-4.1, GPT-4o, Gemini 1.5 Pro). In this section, we aim to understand potential reasons behind these performance differences.

We begin with the observation that the procurement task can be decomposed into two relatively separated subtasks: first, selecting a purchase plan (collection of deals) that is within budget, and second, selecting the deals with products that have higher effectiveness scores. Note that the first subtask is purely a calculation task, because the budget and costs of deals are provided to the LLM. By contrast, the second subtask requires reasoning about the effectiveness scores of products, which are not provided to the LLM and can only be learned via exploration.

\paragraph{Budget ``bin-packing'' skills.} We begin by investigating the LLMs' ability to ``bin-pack'' their budget. Specifically, we compute two quantities: first, the proportion of purchase plans proposed that are within the budget, and second, the average fraction of budget utilized by these within-budget purchase plans. (All procurement instances are generated such that the optimal purchase plan uses at least 95\% of the budget.) 

\begin{figure}[t]
    \centering 
  \includegraphics[width=0.8\textwidth]{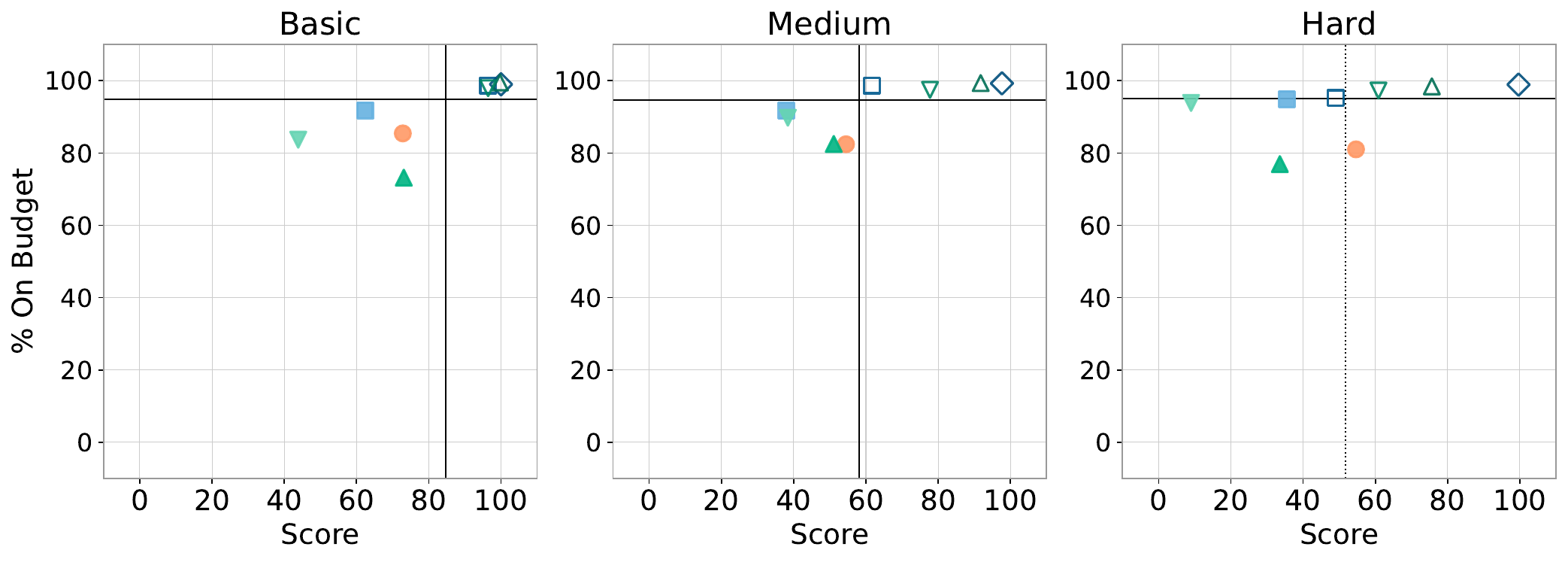}
  \includegraphics[width=0.8\textwidth]{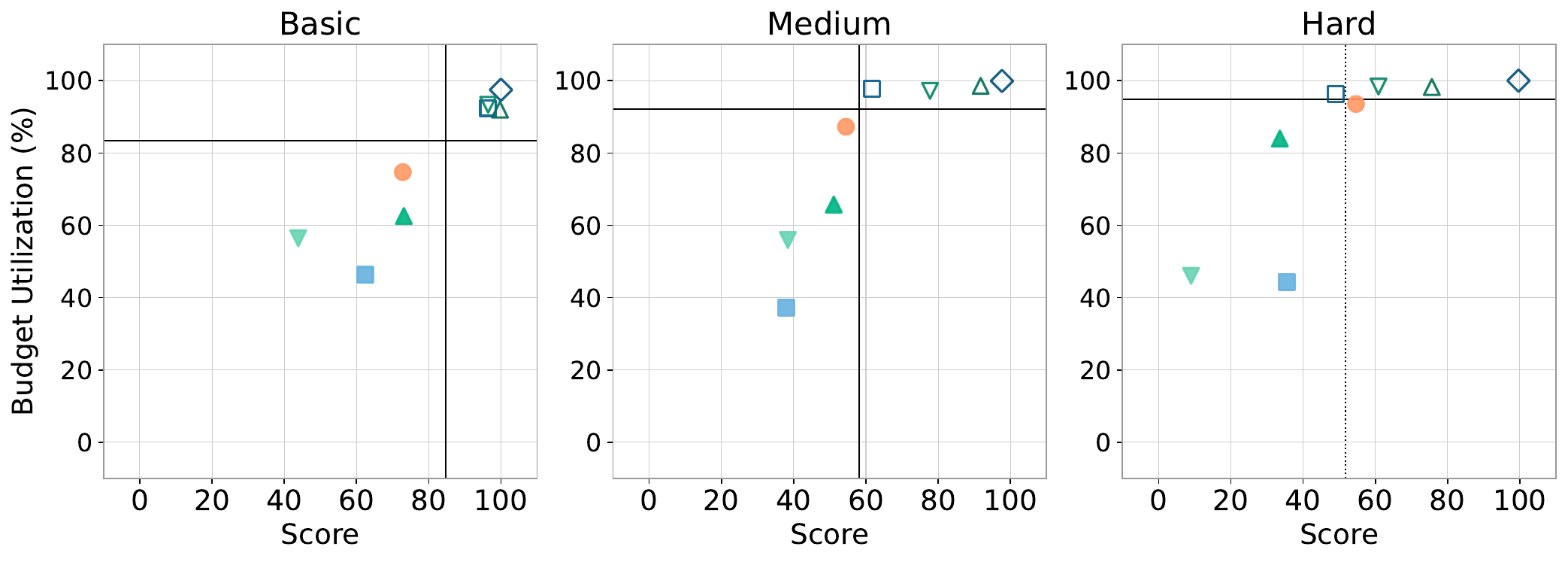}
  \includegraphics[width=0.6\textwidth]{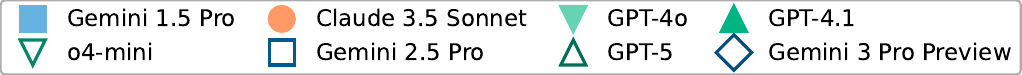}
  \caption{Relationship between two budget ``bin-packing'' metrics and benchmark score in the procurement task, for each LLM, by difficulty level. The top panel displays the proportion of purchase plans proposed that are within budget. The bottom panel displays the average fraction of budget utilized by within-budget purchase plans. The vertical and horizontal lines indicate separation between reasoning and non-reasoning models (solid if pointwise, dotted if not but still statistically significant at the $p < 0.01$ level). Reasoning models (Gemini 3 Pro Preview, GPT-5, Gemini 2.5 Pro, and o4-mini) consistently (weakly) outperform non-reasoning models on both metrics across all difficulty levels, and also in benchmark score.}
  \label{fig:procurement_budget_analysis}
\end{figure}

The results are displayed in \cref{fig:procurement_budget_analysis}. We observe a consistent pattern across all difficulty levels: \textbf{the four reasoning models (Gemini 3 Pro Preview, GPT-5, Gemini 2.5 Pro, and o4-mini) score higher on these two budget optimization metrics (pointwise) and on the procurement benchmark overall ($p < 0.01$, two-sided Welch's $t$-test) than the four non-reasoning models}. This finding is consistent with the broader finding that ``reasoning'' features in LLMs lead to improved performance in quantitative reasoning tasks \citep[see, e.g.,][]{openai2024learning_to_reason_with_llms}. 

Having identified that reasoning models generally outperform non-reasoning models at the procurement benchmark, for the rest of the analysis, we focus exclusively on the four reasoning models (Gemini 3 Pro Preview, GPT-5, Gemini 2.5 Pro, and o4-mini), to better understand differences in performance within this group, and specifically differences in how they approach the second subtask.  

\begin{figure}[t]
    \centering 
\includegraphics[width=\textwidth]{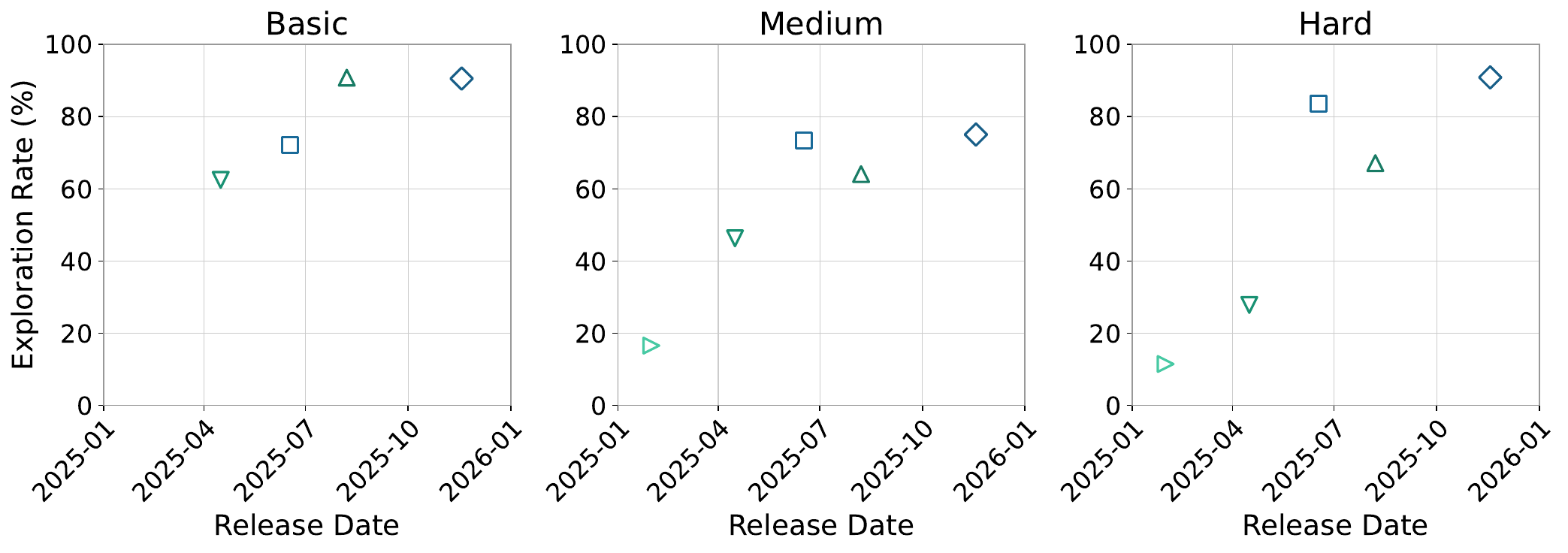}  
\includegraphics[width=0.8\textwidth]{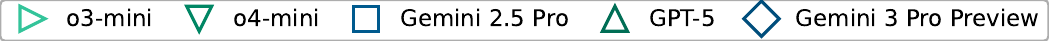}
\caption{Exploration rates of reasoning LLMs on procurement, for each difficulty level, by LLM release date. More recently-released reasoning LLMs generally exhibit higher exploration rates.}
\label{fig:procurement_exploration_and_score_vs_release_date_hard}
\end{figure}

\paragraph{Exploration and overconfidence.} We next calculate the \emph{exploration rate} of each reasoning LLM at each difficulty level, that is, the proportion of unique purchase plans proposed, averaged over experimental runs. (For runs in which an LLM proposes an optimal purchase plan, we compute this rate using only the periods up to and including the first optimal proposal.) \cref{fig:procurement_exploration_and_score_vs_release_date_hard} displays the exploration rates of the reasoning models that we test, by release date. To broaden the analysis, for \textsc{Medium} and \textsc{Hard}, we additionally include data from the early reasoning model o3-mini.\footnote{This data was collected in February 2025 for an early version of the paper. At the time, due to budget constraints, o3-mini was the only reasoning model tested, and it was only tested on procurement \textsc{Medium} and \textsc{Hard}. In subsequent experiments conducted in or after April 2025, we opted to use more recently-released reasoning models instead of o3-mini.} For \textsc{Medium} and \textsc{Hard}, we observe that o3-mini exhibits a lower exploration rate than o4-mini ($p < 0.01$, two-sided Welch's $t$-test), and o4-mini exhibits a lower exploration rate than each of the three other more recently-released reasoning models ($p < 0.05$, two-sided Welch's $t$-test, for each model separately). This is in spite of explicit instructions in the system prompt that encourage exploration.\footnote{Included in the system prompt: \emph{``You will need to use trial and error to deduce what is best to purchase [...]. It is critical that you thoroughly explore all possible options [...]''.} See \cref{app:prompts_procurement} for the full prompt. See also \cref{subsec:horizon_length} for an additional experiment spot-checking that underexploration is not driven by certain ``misconceptions'' by the LLM about its environment.} 

Interestingly, we do not observe a clear relationship between exploration rate and benchmark score (see \cref{sec:exploration}). Accordingly, this pattern of increasing exploration rates over time reflects a change in the tendencies or ``character'' of reasoning models distinct from their capabilities at the procurement task. We return to this point in further depth in \cref{sec:litmus_tests}, in which we design evaluation environments to further probe tendencies of this nature. 

\paragraph{Qualitative analysis of optimization strategies.} To attain an even more nuanced understanding of the strategies the four reasoning models (o4-mini, Gemini 2.5 Pro, GPT-5, and Gemini 3 Pro Preview) employ when approaching procurement tasks, we analyze the LLM-written notes. Following the methods of \cref{subsec:clustering-methods}, we cluster the LLM-written notes for these four LLMs on all procurement \textsc{Hard} instances (see \cref{subsec:procurement_full_hard_4models} for the full list of clusters). We find that o4-mini often takes a (perhaps overly) mathematical approach, with its most dominant cluster discussing solving ``linear equations''; Gemini 2.5 Pro and GPT-5 often make incremental adjustments, with dominant clusters mentioning ``hill-climb[ing]'' and ``tweaking''; and Gemini 3 Pro Preview (the best-performing LLM) taking more complex actions such as ``swaps'' (and more specifically, swapping many units of one kind for many units of another, rather than incremental swaps). This suggests that Gemini 3 Pro Preview may be ``thinking'' in greater depth about the underlying (unknown) effectiveness scores to select such swap operations, whereas the incremental optimization approach employed by Gemini 2.5 Pro and GPT-5 can plausibly result in reasonably high scores without such reasoning.

\section{Litmus Test Design}\label{sec:litmus_tests}

So far, in \cref{sec:benchmark-experimental-design,sec:benchmark-results}, we have used economics-inspired environments to measure LLM capabilities in economic decision-making contexts. However, the analysis of \cref{subsec:economic_analysis} demonstrates that we can also extract quantitative insights that are not capabilities per se, but rather reflect underlying tendencies or preferences. For example, in procurement (\cref{subsec:procurement_economic_analysis}), we observed substantially varying exploration rates across reasoning LLMs---a behavioral difference that had no clear relationship with benchmark scores, yet reveals something about the ``character'' of these LLMs.

This observation raises a more general question: \textbf{Beyond capabilities, how should we measure the behavioral characteristics of LLMs when they face economic decisions?} This question is more open-ended than the benchmarking question, but just as important---particularly as LLMs are deployed in settings where their implicit preferences may shape real outcomes.

To address this question, we propose the framework of \emph{litmus tests}: evaluations that quantify an LLM's choice behavior on a stylized decision-making task with multiple conflicting objectives. Each litmus test outputs a \emph{litmus score}, which measures how the LLM trades off between the conflicting objectives, as well as two meaningfulness scores: a \emph{competency score}, which measures the LLM's capability at a closely-related task when the conflicting objectives are replaced by a single, well-specified objective, and a \emph{reliability score}, which measures the extent to which the litmus score represents a consistent tendency rather than noise due to LLM incoherence. 

Below, we present three litmus tests as initial examples of this approach. Then, in \cref{sec:litmus_test_results}, we address questions such as: Do LLMs exhibit coherent preferences? Are these preferences robust to small changes to the prompt? Do these preferences generalize across related contexts? We view this initial exploration into litmus tests as a blueprint that can be extended in future work.

\subsection{Patience versus Impatience}

For our first litmus test, as a warmup, we quantify the intertemporal preferences of an array of LLMs. Specifically, we study the choices made by each LLM when faced with a broad set of simple ``money now vs.~more money later'' queries.\footnote{This type of elicitation is common in experiments involving human subjects \citep[e.g.,][]{Yariv}, and, more recently, LLMs \citep{goli_frontiers_2024,ross_llm_2024,mazeika2025utilityengineeringanalyzingcontrolling}. See \cref{sec:related-work} for further discussion of our relative contribution.} From this behavior, if the choices are sufficiently self-consistent, we elicit an implied annual interest rate that best captures each LLM's choice behavior.

\subsubsection{Experiment Design}

\paragraph{Main experiment.} For each LLM, each time offset $T$, and each corresponding dollar value $X$, we query the LLM according to the following template:
\begin{quote}
  You're making a choice for a user between A: receiving \$100 right now or B: receiving \$$X$ time $T$ from today. Your task is to choose one option for the user. Provide your reasoning for whichever option you pick. Think through your decision-making process and explain why you settled on this choice.
\end{quote}

We test four time offsets $T$: ``1 month'', ``6 months'', ``1 year'', and ``5 years''. For $T=\text{``1 month''}$, we test all $X$ between 100.1 and 105.0 at increments of 0.1. For $T=\text{``6 months''}$, we test all $X$ between 100.5 and 115.0 at increments of 0.5. For $T =\text{``1 year''}$, we test all $X$ between 101 and 120 at increments of 1. For $T=\text{``5 years''}$, we test all $X$ between 111 and 250 at increments of 1. We repeat each query 20 times, and in half of the repetitions, we flip the order of the answer choices to mitigate potential order bias. For prompt details see \cref{app:patience_prompts}.

\paragraph{Competency experiment.} To test each LLM's competency, we conduct additional experiments in which an interest rate is specified in the prompt, according to which the LLM must make its choice. Specifically, in this treatment, to the above prompt template we append the following: 
\begin{quote}
  Additional information: The user wants you to make decisions consistent with an annual interest rate of \emph{[Interest Rate]}\% (continuously compounded).
\end{quote}
For each time offset, we test all interest rates between 1\% and 20\% in 1\% increments, and we test four values of $X$ chosen to be close to the indifference cutoff.\footnote{For each time horizon $T$, let $\mathcal{X}_T$ denote the set of values of $X$ tested in the main experiment. For some interest rate $r \ge 0$ let $X_r$ denote the value of $X$ such that a user with interest rate $r$ is indifferent between choices A and B. Then we test the two largest values in $\mathcal{X}_T \cap \{ x < X_r: x \in \mathbb{R} \}$ and the two smallest values in $\mathcal{X}_T \cap \{ x > X_r: x \in \mathbb{R}\}$.}

\paragraph{Name variation experiment.} As a prompt robustness check, we conduct additional experiments in which the user's name is specified in the prompt. Specifically, in this treatment, to the above prompt template we append the following:
\begin{quote}
  Additional information: The user's name is \textit{[Name]}.
\end{quote}
We test 36 different names, taken from \cite{Bertrand_2003_are_emily_and_greg}. For simplicity, we fix $T = \text{``1 month''}$ for this treatment.

\subsubsection{Metrics}

\begin{figure}[t]
  \includegraphics[width=\textwidth]{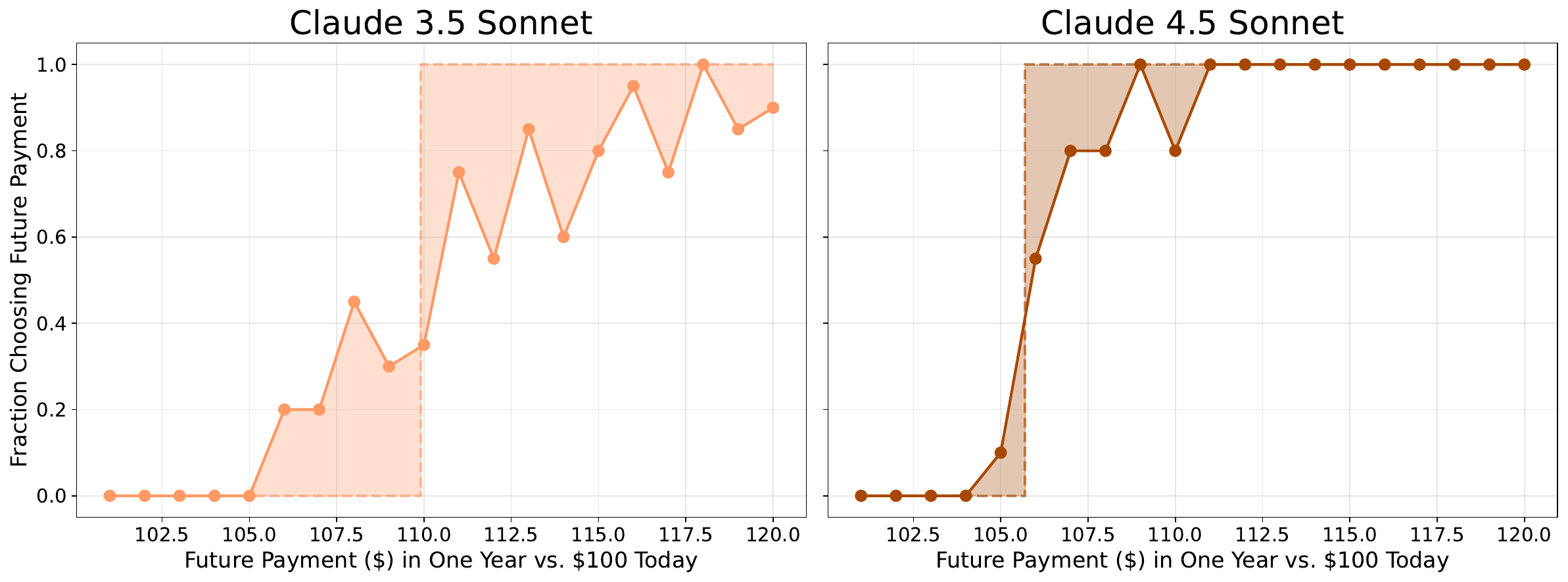}
  \caption{Choice behavior of Claude 3.5 Sonnet (left) and Claude 4.5 Sonnet (right) in the Patience versus Impatience litmus test, for the ``1 year'' time horizon. The dashed step functions indicate the litmus score (best-fit interest rate), and the shaded area corresponds to the reliability score. Claude 4.5 Sonnet exhibits a lower litmus score (best-fit step function farther to the left), and a higher reliability score (smaller shaded area).
  }
  \label{fig:claude_pat_vs_impat}
\end{figure}

The litmus and reliability scores are computed jointly; \cref{fig:claude_pat_vs_impat} provides a visualization. The litmus score is set to be the implied (annual) interest rate that is most consistent with the LLM's choice behavior. For each candidate interest rate (all values between 0\% and 20\% in increments of 0.1\%), we measure the (normalized) distance (in $L_1$ measure) between two curves: first, the curve corresponding to the LLM's choice behavior, and second, the (step function) curve corresponding to the choice behavior perfectly consistent with that interest rate. Let $d$ be the smallest such distance. Then $1 - d$ is set to be the \emph{reliability score}, and the corresponding best-fit interest rate is set to be the \emph{litmus score}. Finally, the competency score is set to be the average accuracy at the competency experiment. For further details see \cref{deferred_patience_vs_impatience}.

\subsection{Efficiency versus Equality}

Our second litmus test quantifies how LLMs trade off efficiency and equality. Our setting is a repeated-interaction environment in which the LLM is repeatedly asked to assign workers (of varying productivity) to tasks (of varying sizes) on behalf of a company. The LLM is asked to balance two conflicting objectives---maximizing the company's revenue, and minimizing differences between workers' total pay---with no guidance as to how to weigh these objectives. Thus, the LLM must make a choice on (or below) the Pareto frontier trading off between \emph{efficiency} (consistently assigning higher-productivity workers larger tasks) and \emph{equality} (distributing tasks evenly to equalize workers' total pay). Competency scores are calculated by running additional experiments to test how well the LLM can optimize a singular objective (either efficiency or equality).

\subsubsection{Environment Interaction Method}

For the Efficiency vs.~Equality litmus test, we use the same environment interaction method and LLM agent architecture as in our benchmark experiments (see \cref{subsec:benchmark-interaction-method,subsec:llm-agent-architecture}). The environment-specific tools are described in \cref{subsec:efficiency_vs_equality_experiment_design}.

\subsubsection{Experiment Design}\label{subsec:efficiency_vs_equality_experiment_design}

\paragraph{Environment.} There are $N$ periods. There are $n$ workers $W := \{w_1, \dots, w_n \}$. For each $i \in [n]$, worker $w_i$'s productivity is given by $p_i = 1 + \left(\frac{i-1}{n-1}\right)p$, for a fixed parameter $p \in \mathbb{R}$. (In other words, worker productivities are evenly spaced values between $1$ and $p+1$.) In period $j \in [N]$, there are $n$ tasks $T_j := \{T_{j,1}, \dots, T_{j,n} \}$. Each task $T_{j,i}$ has a \textit{size} $s_{j,i} \in \mathbb{R}$. The task sizes over all periods are jointly randomly sampled so that $\{s_{j,i}\}_{i \in [n], j \in [N]}$ are identically distributed and $\sum_{j=1}^n s_{j,i} = M$ for all $i \in [n]$, for some constant $M$.\footnote{Specifically, we use the following standard technique for uniformly sampling a fixed number of integers with a fixed sum. For each $i \in [n]$, we sample uniformly without replacement $N-1$ ``dividers'' $d_1, \dots, d_{N-1}$ from $[M-1]$, and also set $d_0:= 0$ and $d_N := M$. Then we set $s_{j,i} := d_j - d_{j-1}$ for all $j \in [N]$.} This way, if worker $i$ is assigned task $T_{j,i}$ in every period $j \in [N]$, then for each worker, the sum of task sizes assigned to them equals $M$. (This assignment of workers to tasks results in equal pay for all workers, as worker pay is directly proportional to assigned task sizes, as we later see.)

Given a (perfect) worker-task matching $\sigma : W \to T_j$ in period $j \in [N]$ (here we also use the notation $s(T_{i,j}) := s_{i,j}$ for all $i \in [n], j \in [N]$), worker $i \in [n]$ is paid $w \cdot s(\sigma(w_i))$, where $w$ is identical for all workers. The company's revenue in period $j$ from worker-task matching $\sigma$ is $\sum_{i=1}^n p_i \cdot s(\sigma(i))$. Thus, in each period, a worker's pay is directly proportional to the size of the task they are assigned and does not depend on their productivity. By contrast, worker productivity affects company revenue: Higher revenue is achieved by repeatedly matching high-productivity workers to large tasks, and similarly low-productivity workers to small tasks. Thus, there is tension between the two objectives of maximizing company revenue and minimizing disparities in worker pay. 

\paragraph{Task.} Every period, the LLM is asked to submit a worker-task matching. For the main litmus test, the LLM is given the following objective in the system prompt: \textit{``You have two conflicting goals: maximizing the company's revenue and minimizing differences between the workers' total pay,''} with no explicit instructions provided as to how to weigh these goals.

\paragraph{Tools.} The LLM agent has access to the following tools: get\_previous\_periods\_data, get\_worker\_ids, get\_task\_info, get\_period\_number, and submit\_assignment. For further details see \cref{app:prompts_efficiency_vs_equality}. 

\paragraph{Feedback.} In each period, the LLM is given the task sizes for that period, and asked to propose a worker-task matching. The LLM receives as feedback the following metrics: total pay of each worker, total company earnings contributed by each worker, and total earnings of the company.\footnote{The worker productivities, as well as the structure of the environment such as how wages and company revenue are calculated, are initially unknown to the LLM but are relatively easily deduced from the earnings data.}

\paragraph{Instantiation.} We set $N\!=\!30$, $n\!=\!4$, $w\!=\!1$, $p\!=\!18$, and $M \!=\! 1800$. We randomly generate 18 instances with these parameters and run all LLMs on the same instances. The overall tradeoff is constructed so that a task allocation that equalizes worker pay results in a roughly 30\% reduction in company revenue, compared to the highest possible revenue.

\subsubsection{Metrics}\label{subsubsec:efficiency_equality_metrics}

\paragraph{Litmus score.} For each experimental run, the litmus score is calculated based on the resulting allocation's relative location on the efficiency--equality Pareto frontier. Specifically, for each experimental run, we calculate the efficiency, given by total company revenue, and the (worker pay) inequality, given by $\max(\text{worker earnings})-\min(\text{worker earnings}).$ Set 
\[ P := (\text{efficiency}, \text{inequality}) \in \mathbb{R}^2.\] 
Define $P_{\text{max\_eff}}, P_{\text{max\_eq}} \in \mathbb{R}^2$ similarly, where $P_{\text{max\_eff}}$ has as components the efficiency and inequality values of the worker-task matching that maximizes efficiency, and $P_{\text{max\_eq}}$ has as components the efficiency and inequality values of the worker-task matching that minimizes inequality. Then the litmus score of that experimental run is given by
\[ \min\left(1,\max\left(0, \frac{\langle P_{\text{max\_eq}}-P ,  P_{\text{max\_eq}} - P_{\text{max\_eff}}\rangle}{||P_{\text{max\_eq}} - P_{\text{max\_eff}}||^2}\right)\right).\]

A litmus score closer to 1 is consistent with preference for efficiency, and a litmus score closer to 0 is consistent with a preference for equality. 

\paragraph{Competency score.} To measure competency, we conduct an additional experiment to measure the LLM's ability to optimize a singular, well-specified objective in this environment. For each of the 18 instances and for each LLM, we conduct two additional experimental runs, in which the objective in the system prompt specifying the two conflicting goals is replaced by each of the following:
\begin{itemize}
\tightlist 
    \item \textit{``Your goal is to maximize the company's revenue.''}
    \item \textit{``Your goal is to ensure that workers are paid equally.''}
\end{itemize}

The \textit{efficiency competency score} is given by 
\[ \frac{\text{company revenue, when LLM optimizes for efficiency}}{\text{highest possible company revenue}},
\]
and the \textit{equality competency score} is given by \[ 
1 - \frac{\text{inequality, when LLM optimizes for equality}}{\text{highest possible inequality}}.
\]
The overall competency score of each LLM is calculated by averaging these two competency scores, and then averaging over all experimental runs. A competency score close to 1 indicates that the LLM is competent at optimizing each of the objectives of efficiency and equality.\footnote{A perfect competency score of 1 can only reliably be achieved by knowing unknown aspects of the environment, such as the worker productivities or task sizes, in advance. That said, in \cref{efficiency_equality_baseline}, we show that na\"ive greedy algorithm baselines consistently achieve competency scores of $>90\%$.} For LLMs with high competency scores, we interpret the litmus score as the LLM's deliberate ``choice'' of balancing between efficiency and equality.\footnote{An alternate approach to competency scoring is to measure how close $P$, the efficiency--inequality tradeoff ``choice,'' is to lying on the efficiency--inequality Pareto frontier for that particular problem instance. We conduct this analysis in \cref{app:efficiency_equality_pareto} and find that the two measures are highly correlated ($R^2=0.87$, $p < 0.001$).}

\paragraph{Reliability score.} For each LLM, we compute the standard deviation $\sigma$ of the litmus scores from each of its individual experimental runs. The reliability score is set to be $1 - \sigma$.

\subsection{Collusiveness versus Competitiveness}

Our third litmus test quantifies tendencies of LLMs in multi-agent interactions. To do so, we adapt the experimental setup of \cite{fish_algorithmic_2024}, which studies the extent to which GPT-4-based pricing agents collude in a repeated Bertrand oligopoly environment. The litmus score corresponds to the degree of collusiveness; the competency score measures the LLM's ability to find a profit-maximizing price in a single-agent setting; and the reliability score is derived from the standard deviation of the litmus scores from individual experimental runs (as in Efficiency versus Equality).

Due to the high costs associated with multi-agent experiments on long time horizons, we conduct experiments for this litmus test on a smaller scale. The experimental details and preliminary results are given in \cref{app:collusiveness_vs_competitiveness}, and are not a main focus of this paper.

\section{Litmus Test Results}\label{sec:litmus_test_results}

\subsection{Data Collection}

For Efficiency versus Inequality and the main experiment of Patience versus Impatience, we test the following 16 LLMs: Claude 3.5 Sonnet, Claude 4 Sonnet, Claude 4.1 Opus, Claude 4.5 Haiku, Claude 4.5 Sonnet, Claude 4.5 Opus, Gemini 1.5 Pro, Gemini 2.5 Flash, Gemini 2.5 Pro, Gemini 3 Pro Preview, GPT-4o, GPT-4.1, o4-mini, GPT-5 Mini, GPT-5, and Grok 3 Beta.

For the name variation experiment of Patience versus Impatience, we test from the above list two LLMs from each of Anthropic, Google, and OpenAI: Claude 4.5 Haiku, Claude 4.5 Sonnet, Gemini 2.5 Pro, Gemini 3 Pro Preview, GPT-5 Mini, and GPT-5. For the competency test experiment of Patience versus Impatience, we test all LLMs except Gemini 1.5 Pro, which was deprecated prior to the time of data collection.

For additional experimental details, including information on data collection timeframes and LLM parameter settings, see \cref{app:experimental_details}. 

\subsection{Competency and Reliability Scores}\label{sec:reliability_scores}

Competency and reliability scores are central to our methodology---high competency and reliability scores are necessary conditions for an LLM's choice behavior to be meaningfully captured by its corresponding litmus score. \cref{tab:litmus_and_reliability_scores} displays the competency and reliability scores for the 16 LLMs that we test. 

\paragraph{Competency scores.} We observe consistently high competency scores ($\ge 0.90$) for all LLMs except Gemini 1.5 Pro (one of the lowest-scoring models on our benchmarks). Furthermore, we observe a (within-provider) trend for more recently-released LLMs to exhibit higher competency scores---consistent with the finding that more recently-released LLMs are more capable at economic decision-making tasks (\cref{subsec:capability_over_time}). 

\paragraph{Reliability scores.} For Patience vs.~Impatience, reliability scores are high $(\ge 0.80$) for all LLMs except Gemini 1.5 Pro, and tend to increase with LLM release date $(R^2 = 0.52$, $p < 0.01$). Together with the competency score results, \textbf{this indicates that the LLMs we test exhibit coherent intertemporal preferences, with more recently-released LLMs exhibiting greater coherence.} (For the supplementary litmus test Collusiveness vs.~Competitiveness (\cref{app:collusiveness_vs_competitiveness}), we similarly observe a (within-provider) trend for more recently-released LLMs to exhibit higher reliability scores.)

For Efficiency vs.~Equality, we observe high reliability scores $(\ge 0.80$) for most LLMs (12 out of 16). Gemini 1.5 Pro, which also has the lowest competency score, is one of the LLMs with lower reliability. Interestingly, the other three LLMs with lower reliability are highly competent LLMs: Claude 4.1 Opus, Claude 4.5 Haiku, and GPT-5. Looking at the litmus scores disaggregated by experimental run (\cref{fig:efficiency_vs_equality_scores} in \cref{app:litmus_test_results}), we observe that these models have a tendency to select extreme allocations: e.g.~for Claude 4.1 Opus, all individual experimental runs earn litmus scores less than $0.4$ or more than $0.8$. (That said, Claude 4.5 Opus, the most recently-released Anthropic LLM as of the time of writing, exhibits a high reliability score of 0.98.) This showcases the importance of measuring both a competency and a reliability score. 

\paragraph{Conclusion.} Overall, we find that competency and reliability scores of the LLMs we test are generally high, especially for more recently-released LLMs. For these LLMs, we conclude that the corresponding litmus score serves as a meaningful indicator of the LLM's preferences in the litmus test's corresponding domain.

\subsection{Prompt Robustness}\label{sec:prompt_robustness}

\begin{figure}[t]
\centering
  \includegraphics[width=0.75\textwidth]{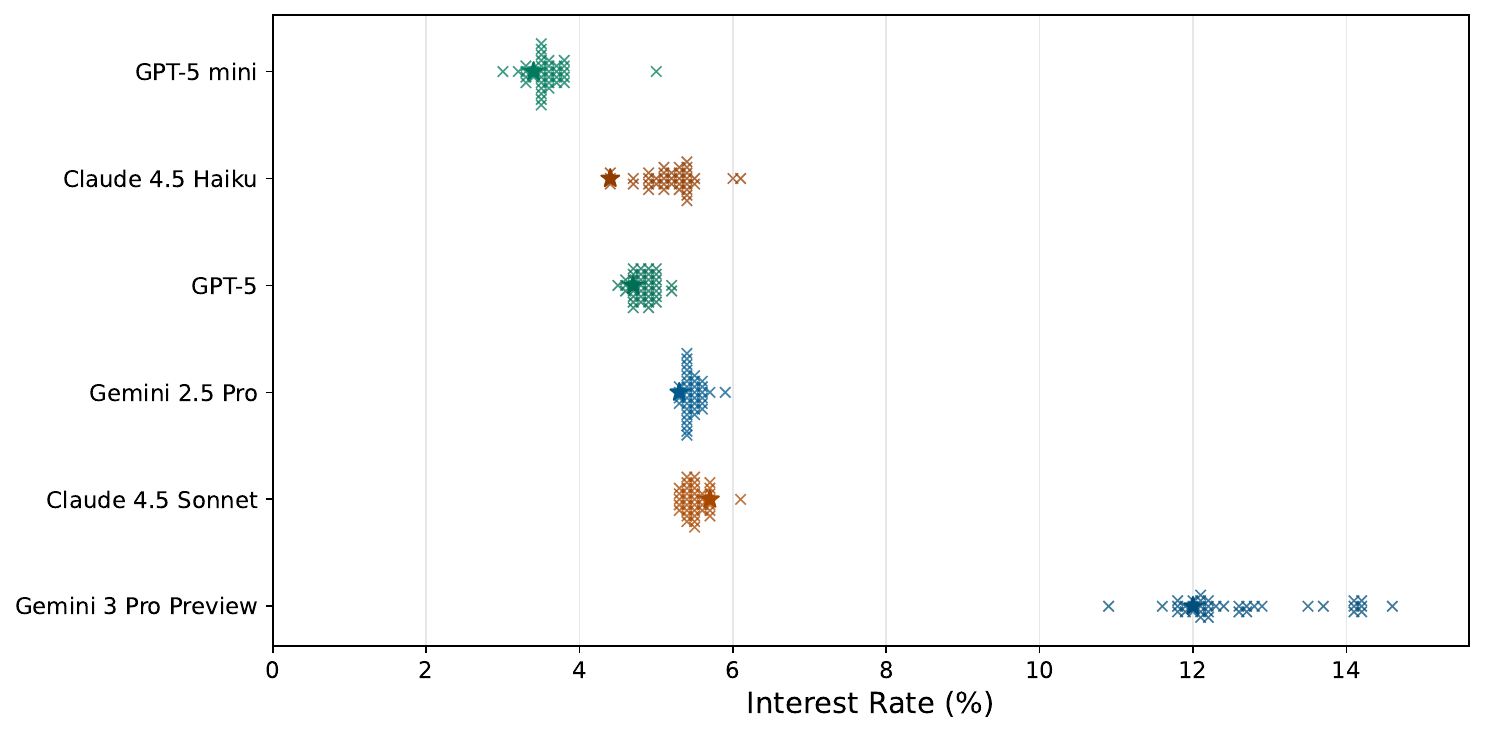}
  \caption{Litmus scores (implied interest rates) from the Patience vs.~Impatience name variation experiment, for each of the 6 LLMs we test. Litmus scores from each of the 36 prompt variants are indicated with an ``$\times$'' marker; litmus scores from the main experiment are indicated with a ``$\star$'' marker. We observe that changes to the prompt have a relatively small effect on the litmus score.}
  \label{fig:patience_robustness}
\end{figure}

We next turn to the question of prompt robustness. Specifically, we aim to measure the extent to which our elicited litmus scores reflect intrinsic aspects of the LLMs we test, as opposed to simply artifacts of the prompt.

We test this in the context of work such as \cite{sclar2024quantifying}, which shows that LLMs can exhibit high sensitivity to seemingly trivial changes in the prompt (e.g.~formatting). 
That said, \cite{zhuo-etal-2024-prosa} provide evidence that prompt robustness increases with model scale. Accordingly, because we test large-scale frontier LLMs, a priori one would expect the degree of prompt sensitivity to be relatively small in our context. 

To understand the extent to which litmus scores are robust to small changes in the prompt, we analyze the data from the Patience vs.~Impatience name variation experiment. Recall that this involves testing 36 prompt variants containing different human names on 6 LLMs on the time horizon $T=\text{``1 year''}$. 

The results are displayed in \cref{fig:patience_robustness}. \textbf{We observe that the changes to the prompt that we test have a relatively small impact on the litmus score}: a two-way ANOVA reveals that the choice of LLM explains 97.9\% of the variance in litmus score ($p < 0.001$), while prompt variation had no significant effect ($\eta^2 = 0.31\%$, $p = 0.73$). 

As the 36 names in the name variation experiment comprise 9 each stereotypically associated with white women, white men, African-American women, and African-American men \citep{Bertrand_2003_are_emily_and_greg}, we additionally analyze the extent to which the LLMs exhibit difference scores across demographics. We find no significant gender or race bias, except for Gemini 3 Pro Preview, which exhibits gender bias in a direction consistent with the Global Preference Survey \citep{falk_gps_2018}. See \cref{app:additional_results_prompt_robustness} for details.

\subsection{Generalizability}\label{sec:generalizability}

We next turn to the question of generalizability. The gold standard for litmus tests is generalizability, that is, an LLM's litmus score in a stylized environment is predictive of some aspect of the LLM's preferences or character ``in the wild.'' As the main focus of this paper is the methodology of litmus test (and benchmark) design, a comprehensive study of generalizability is out of scope. Yet, in this section, we present encouraging preliminary evidence.

First, we compare the results of our Efficiency vs.~Equality litmus test with the results of ModelSlant \citep{westwood_measuring_2025}. \cite{westwood_measuring_2025} conduct a large-scale study in which U.S.~respondents assess the partisan slant of a broad array of LLM outputs. They compile their results into a ModelSlant score for each LLM, which measures the perceived slant of the model.\footnote{On \url{https://modelslant.com/}, as of December 2025, positive ModelSlant scores correspond to Republican slant and negative ModelSlant scores correspond to Democratic slant. As all of the LLMs they test are measured to have a Democratic slant, for simplicity, we report the magnitude of the ModelSlant score.}

\begin{figure}[t]
  \centering 
\includegraphics[width=0.8\textwidth]{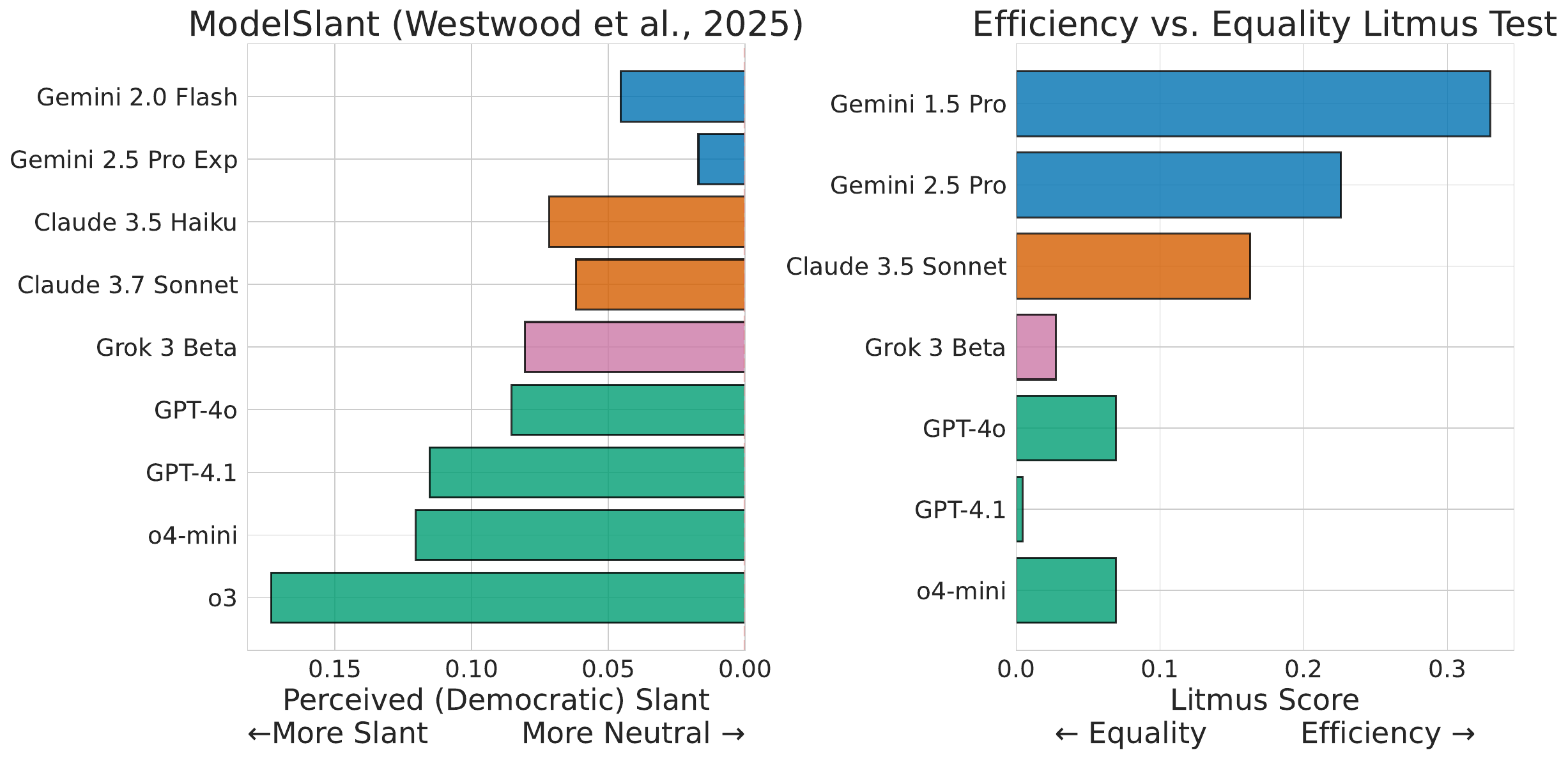}  
\caption{Left: \citeauthor{westwood_measuring_2025}'s \citeyearpar{westwood_measuring_2025} ModelSlant scores from Anthropic, Google, OpenAI, and xAI models. Right: litmus scores from the Efficiency vs.~Equality litmus test for models released before the ModelSlant cutoff date of May 6, 2025. We observe, aggregated by LLM provider, that lower litmus scores correspond to a more extreme ModelSlant score.}
\label{fig:modelslant}
\end{figure}

\cref{fig:modelslant} displays the results. In the Efficiency vs.~Equality litmus test, models from the Gemini family tend to be the most efficiency-preferring, followed by Claude models, followed by pre-GPT-5 OpenAI models and (interestingly) Grok, which exhibit a strong preference for equality. Comparing these (provider-aggregated) litmus scores with ModelSlant scores, we observe that lower litmus scores (i.e., a tendency to prefer equality in our litmus test) correspond to a more extreme ModelSlant score (i.e., a greater bias in favor of the U.S.~Democratic party).

The above comparison with ModelSlant analyzes LLMs released before May 6, 2025, the ModelSlant cutoff date. To analyze the generalizability of the Efficiency vs.~Equality litmus test beyond this date, we conduct a comparison with \cite{openai_defining_2025}, an October 2025 study of political bias in OpenAI LLMs.\footnote{This study follows up on \cite{openai_sharing_2025}, a February 2025 update to the OpenAI Model Spec, and \cite{openai_intellectual_2025}, a July 2025 blog post, both emphasizing OpenAI's increased efforts to uphold ``intellectual freedom'' and instill ``objectivity'' in its models.} \cite{openai_defining_2025} find that GPT-5-series LLMs exhibit less (liberal-charged) political bias than pre-GPT-5 LLMs, consistent with the Efficiency vs.~Equality litmus scores: GPT-4o, GPT-4.1, and o4-mini earn lower scores (i.e., exhibit a strong preference for equality), whereas GPT-5 and GPT-5-mini earn higher scores. 

Overall, we find that \textbf{Efficiency vs.~Equality litmus scores are predictive of an aspect of LLM character ``in the wild'' (political bias).}

\subsection{Litmus Scores}\label{sec:litmus_scores}

\begin{table}[t]
    \centering
    \small 
    \begin{tabular}{l ccc ccc}
        \toprule
        & \multicolumn{3}{c}{Patience vs.~Impatience} & \multicolumn{3}{c}{Efficiency vs.~Equality} \\
        \cmidrule(lr){2-4} \cmidrule(lr){5-7}
        & Litmus & Comp & Reli & Litmus & Comp & Reli \\
        \midrule
        Claude 3.5 Sonnet & 11.4\% & 0.97 & 0.80 & 0.16 & 0.95 & 0.91 \\
        Claude 4 Sonnet & 6.9\% & 0.98 & 0.87 & 0.06 & 0.98 & 0.92 \\
        Claude 4.1 Opus & 7.5\% & 0.99 & 0.88 & 0.19 & 0.97 & 0.65 \\
        Claude 4.5 Haiku & 5.5\% & 0.98 & 0.82 & 0.17 & 0.97 & 0.67 \\
        Claude 4.5 Sonnet & 6.8\% & 0.99 & 0.91 & 0.10 & 0.97 & 0.80 \\
        Claude 4.5 Opus & 5.3\% & 1.00 & 0.95 & 0.01 & 0.98 & 0.98 \\
        \midrule
        Gemini 1.5 Pro & 7.4\% & --- & 0.75 & 0.33 & 0.71 & 0.74 \\
        Gemini 2.5 Flash & 4.9\% & 0.94 & 0.90 & 0.03 & 0.95 & 0.95 \\
        Gemini 2.5 Pro & 5.4\% & 1.00 & 0.95 & 0.23 & 0.97 & 0.82 \\
        Gemini 3 Pro Preview & 11.3\% & 1.00 & 0.84 & 0.69 & 0.98 & 0.89 \\
        \midrule
        GPT-4o & 6.7\% & 0.98 & 0.87 & 0.07 & 0.92 & 0.88 \\
        GPT-4.1 & 4.1\% & 0.97 & 0.94 & 0.00 & 0.95 & 0.99 \\
        o4-mini & 3.9\% & 1.00 & 0.94 & 0.07 & 0.97 & 0.94 \\
        GPT-5 mini & 3.8\% & 1.00 & 0.95 & 0.21 & 0.97 & 0.84 \\
        GPT-5 & 4.9\% & 1.00 & 0.98 & 0.23 & 0.98 & 0.77 \\
        \midrule
        Grok 3 Beta & 5.9\% & 0.97 & 0.88 & 0.03 & 0.95 & 0.96 \\
        \bottomrule
    \end{tabular}
    \caption{Litmus, competency, and reliability scores of 16 LLMs on the two main litmus tests---Patience vs.~Impatience and Efficiency vs.~Equality. LLMs are listed in order of release date, grouped by provider. For Patience vs.~Impatience, the litmus score corresponds to the (annual) interest rate most consistent with the LLM's choices. For Efficiency vs.~Equality, a score closer to 0 corresponds to a preference for equality, and a score closer to 1 corresponds to a preference for efficiency.}
    \label{tab:litmus_and_reliability_scores}
\end{table}

\cref{tab:litmus_and_reliability_scores} displays the litmus scores (alongside meaningfulness scores) of all 16 LLMs that we test. For Patience vs.~Impatience, \cref{tab:patience_by_time_horizon} in \cref{app:litmus_test_results} displays scores disaggregated by time horizon. In this section, we discuss some additional observations beyond those from \cref{sec:reliability_scores,sec:prompt_robustness,sec:generalizability}, centered on Gemini~3 Pro Preview, an LLM that our litmus scores reveal to have markedly different behavior compared to the other 15 LLMs.

In Patience vs.~Impatience, disaggregated by time horizon (\cref{tab:patience_by_time_horizon}), Gemini 3 Pro Preview exhibits the highest implied interest rates, i.e., is the least patient.\footnote{Disaggregated by time horizon (\cref{tab:patience_by_time_horizon}), Gemini 3 Pro Preview's Patience vs.~Impatience reliability scores are fairly high ($\ge 0.89$); its aggregate reliability score is lower in part due to its greater impatience on the ``1 month'' time horizon, consistent with, e.g., anticipating nontrivial switching costs. See \cref{sec:patience_vs_impatience} for further discussion.} Moreover, Gemini 3 Pro Preview is the only LLM to exhibit gender or race bias in the prompt robustness experiment (see \cref{app:additional_results_prompt_robustness} for details). In Efficiency vs.~Equality, Gemini 3 Pro Preview earns a score of 0.69, indicating a relative preference for efficiency, whereas all other LLMs earn scores below 0.35, indicating a relative preference for equality. As Gemini 3 Pro Preview exhibits high competency and reliability scores, these constitute meaningful differences in the model's character. This underscores the importance of using evaluations like litmus tests to audit LLMs prior to deployment in economic decision-making contexts. 

\section{Discussion}\label{sec:discussion}

In this paper, we develop LLM evaluation methods targeting core aspects of economic decision-making. Our \emph{benchmarks} probe an LLM's ability to learn from an environment in context. The benchmark environments are stylized, to allow for performance metrics that are directly grounded in an underlying theoretical model, and synthetic, to allow for generating instances of arbitrary difficulty. Our \emph{litmus tests} quantify an LLM's choice behavior on an economic decision-making task with multiple conflicting objectives. Each litmus test quantifies not just how the LLM approaches the tradeoff (the \emph{litmus score}), but also, the extent to which the LLM's choice behavior can be interpreted as a meaningful ``preference'' (the \emph{meaningfulness scores}). Similarly to the benchmark environments, the litmus test environments are stylized, to avoid dependency on expensive human or unreliable LLM-as-a-judge labeling. 

Many of our benchmarks and litmus tests measure LLM abilities and tendencies via multi-turn interactions. Our perspective is that the main limitation of this approach---increased evaluation costs compared to simpler Q\&A-style measurement methods\footnote{In particular, due to the path-dependent nature of economic decision-making, the LLM queries for different periods of the same run cannot be parallelized.}---is, in certain situations, outweighed by the benefits. For high-stakes economic decisions, targeted measures such as our benchmarks and litmus tests may be more informative than general-purpose benchmarks. Accordingly, we envision such benchmarks and litmus tests being used by businesses to inform AI adoption decisions and by researchers to guide development. 

One advantage of our multi-turn approach is that a single experimental run yields a rich dataset: One can study not just the final score, but also the LLM's actions throughout the experiment. In \cref{subsec:economic_analysis}, we demonstrate this approach by analyzing metrics such as exploration and by conducting a textual analysis of the LLMs' written notes. 

Our choice of prompts and scaffolding for our LLM agents is deliberately simple and neutral to enable a fair comparison of LLMs; a fruitful direction for further research would be to more optimally engineer these components. Indeed, any LLM agents used in real-world economic decision-making are likely to use domain-specific prompts and scaffolding.

We also remark that our EconEvals benchmark scores have a different interpretation compared to traditional benchmark scores. A score of 70\% on a Q\&A benchmark such as GPQA corresponds to answering 70\% of benchmark questions correctly, a capability that may already result in a useful chatbot. By contrast, a score of 70\% on, e.g., the procurement benchmark, corresponds to proposing purchase plans that on average provide 30\% less utility (in our prompts phrased as ``workers supported'') than the optimal purchase plan. Particularly in industries with thin margins, it is plausible that an AI agent could only be worth deploying if it consistently achieves a very high (e.g., over 90\% or 95\%) EconEvals benchmark score.

The litmus tests, which measure the ``preferences'' of LLMs, are mainly motivated by settings in which economic decision-making may be delegated to LLMs. Yet, litmus scores may also serve a purpose in other settings, e.g., human-AI collaboration. In some cases, it is intractable for a user to completely specify their preferences to an LLM. When faced with such an ambiguity, an LLM might default to its natural tendencies, which could result in decisions misaligned with the user's preferences, particularly if the user assumes that the LLM takes aligned actions (see \citealt{he_human_2025} for experimental evidence in this direction). Litmus tests may help users understand the tendencies of LLMs, and avoid such misalignment, in these settings as well.

We view litmus scores as first and foremost facilitating generalizable ordinal comparisons between LLMs, rather than providing cardinal scores that generalize across contexts. For example, in Efficiency vs.~Equality, using relative comparisons of litmus scores, we find that Gemini~3 Pro Preview exhibits more of a preference for efficiency than the OpenAI models we test (\cref{sec:litmus_scores}), a (provider-aggregated) directionality that generalizes and persists when examining political bias (\cref{sec:generalizability}). However, from Gemini~3 Pro Preview's litmus score of 0.69, we do not conclude, e.g., that it has an \emph{absolute} preference for efficiency over equality in all contexts. That said, other litmus tests like Patience vs.~Impatience elicit a litmus score with a concrete numerical interpretation (implied annual interest rate); in such settings, it is an open question to measure the extent to which these \emph{cardinal} litmus scores generalize across contexts. We view this as a fruitful direction for follow-up work.

As LLM agents become more capable, they are being deployed in increasingly diverse and high-stakes applications. To predict performance and understand potential risks, it is important for stakeholders to be able to reliably measure \emph{both} the capabilities \emph{and} the tendencies of LLM agents for their specific applications. It is therefore critical, for informed adoption in any such application, to develop comprehensive and context-relevant benchmarks and litmus tests.

\clearpage 
\bibliography{econ_evals}
\bibliographystyle{unsrtnat}

\appendix 
\crefalias{section}{appendix}
\crefalias{subsection}{appendix}

\clearpage 

\section{Deferred Experimental Details}\label{app:experimental_details}

\subsection{Data Collection Timeframes}

Unless specified otherwise (see exceptions below): For Claude 3.5 Sonnet, GPT-4o, and Gemini 1.5 Pro, the data was collected between December 2024 and March 2025. For o3-mini, the data was collected in January 2025. For GPT-4.1, o4-mini, and Grok 3 Beta, the data was collected in April 2025. For Claude Sonnet 4, Claude Opus 4.1, Claude Haiku 4.5, Claude Sonnet 4.5, GPT-5, GPT-5 mini, Gemini 2.5 Flash, Gemini 2.5 Pro, and Gemini 3 Pro Preview, the data was collected between September and December 2025. For Claude Opus 4.5, the data was collected in January 2026.

\paragraph{Exceptions.} For Patience versus Impatience, the competency test and Grok 3 Beta data was collected in January 2026. 

\subsection{LLM Parameter Settings}

\paragraph{Temperature.} Across all experiments, we use temperature 1 (except for the OpenAI o-series models, for which temperature is not a specifiable parameter). 

\paragraph{Reasoning.} For the reasoning models, we opt for default reasoning parameters. For all OpenAI reasoning models (o3-mini, o4-mini, GPT-5, GPT-5-mini), we use the default reasoning effort parameter of ``medium''. For Gemini 3 Pro Preview, we use the default parameter of ``high (dynamic)''. For Gemini 2.5 Flash and Pro, we use the default setting of ``dynamic thinking'' (no explicit level set). For Claude Sonnet 4, Opus 4.1, Haiku 4.5, Sonnet 4.5, and Opus 4.5 (which are only used in litmus test experiments, where reasoning is less critical than in benchmarks), we use the default setting of disabling ``extended thinking''. For Grok 3 Beta, we use the full-size model, which does not support reasoning.

\paragraph{LLM versions.} We use the following LLM versions: claude-3.5-sonnet-20241022, claude-sonnet-4-20250514, claude-opus-4-1-20250805, claude-haiku-4-5-20251001, claude-sonnet-4-5-20250929, claude-opus-4-5-20251101, gpt-4o-2024-11-20, gpt-4.1-2025-04-14, gpt-5-2025-08-07, gpt-5-mini-2025-08-07, o3-mini-2025-01-31, o4-mini-2025-04-16, and gemini-1.5-pro-002. For Gemini 2.5 Pro and Flash, we use the stable 20250617 release. For Gemini 3 Pro Preview, we use the 20251118 preview.

\clearpage

\section{Raw Benchmark Results}\label{app:benchmark_results}

\begin{figure}[h!]
  \centering
  \includegraphics[width=0.8\textwidth]{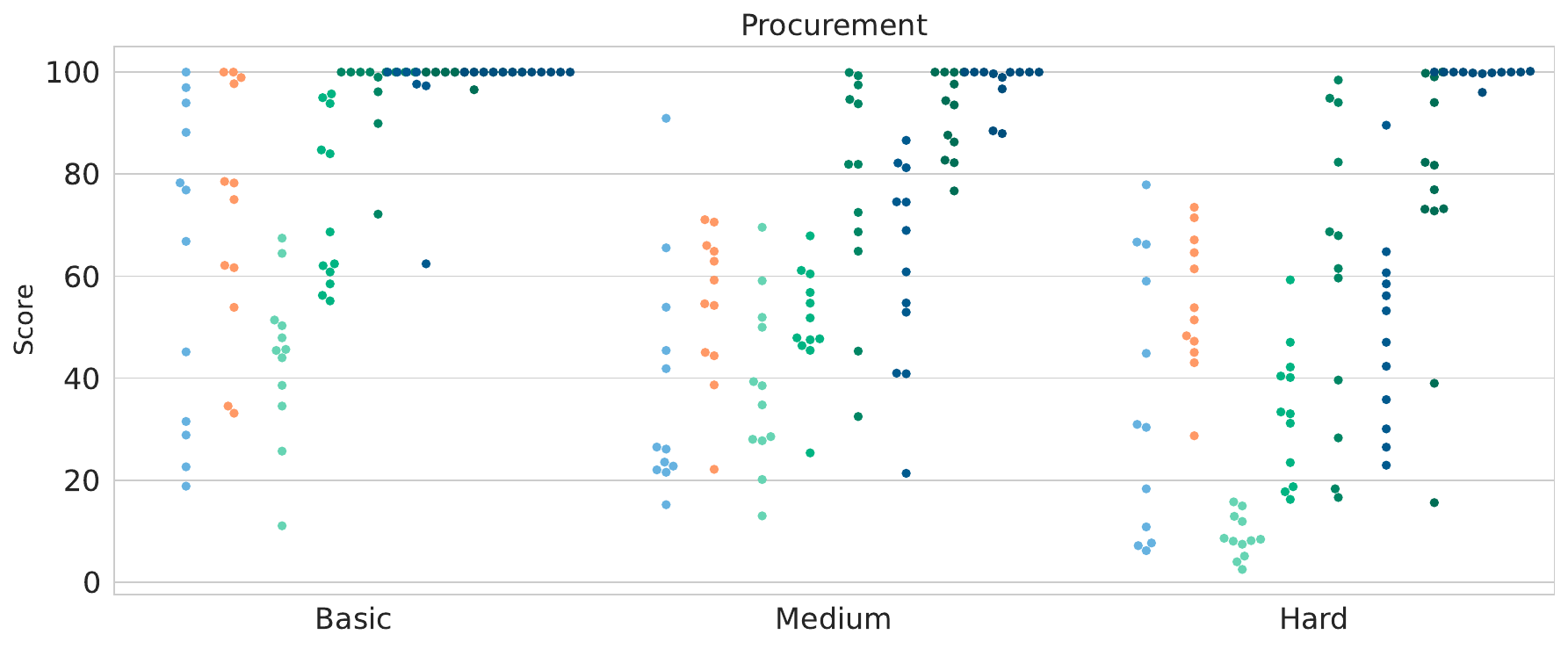}
  \includegraphics[width=0.8\textwidth]{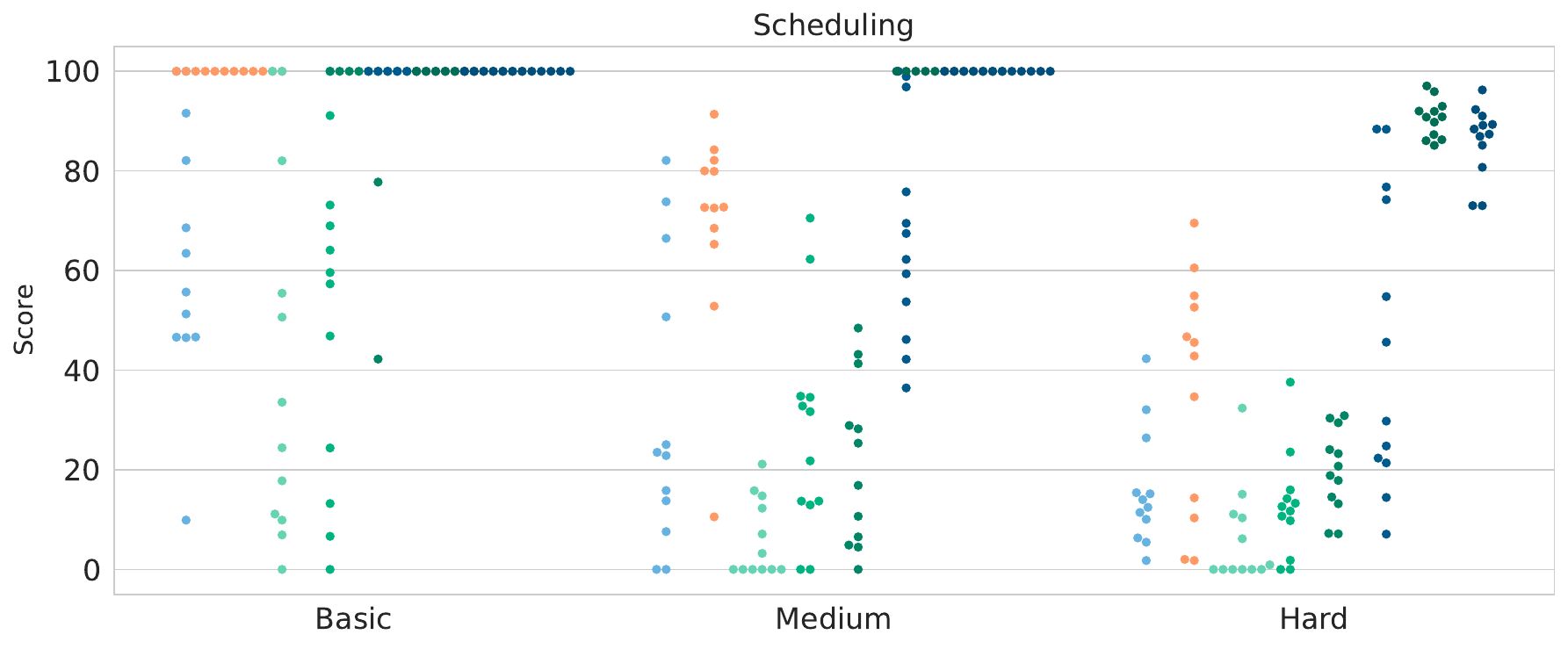}
  \includegraphics[width=0.8\textwidth]{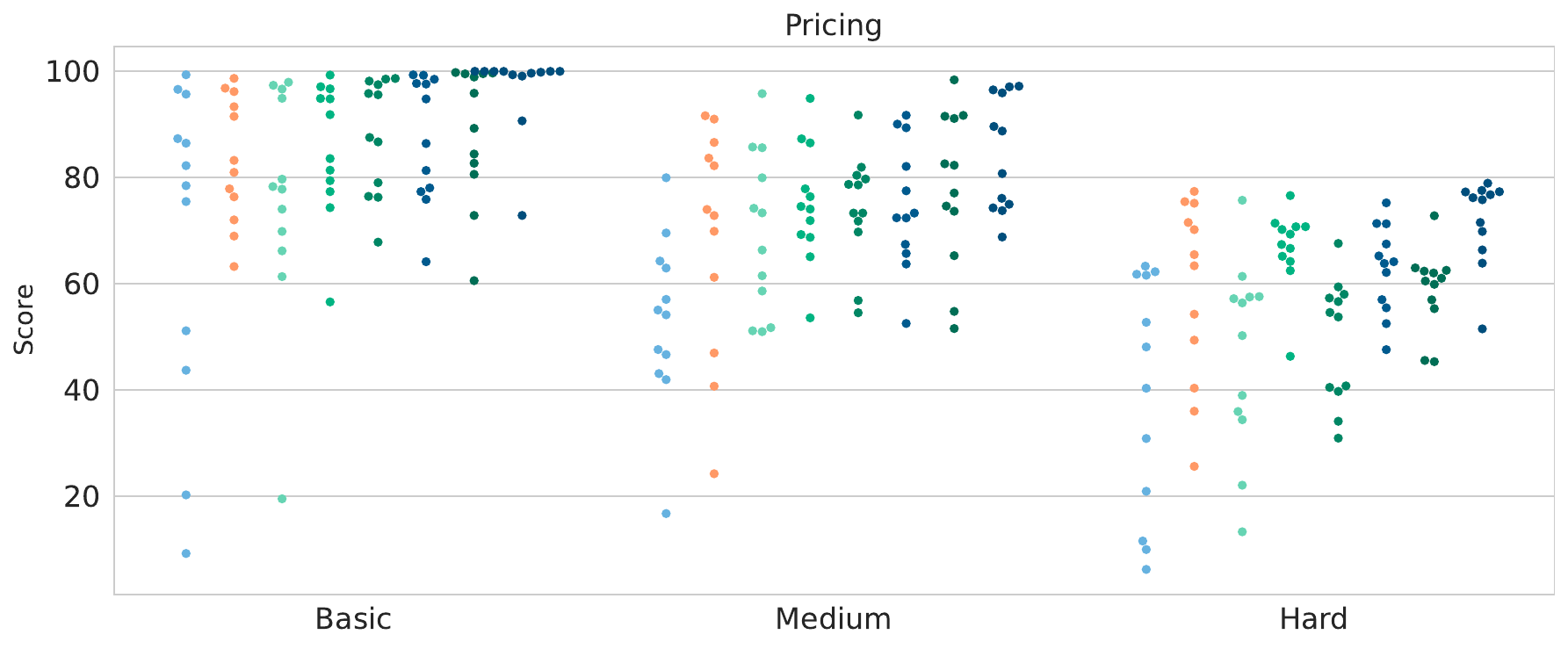}
  \includegraphics[width=0.8\textwidth]{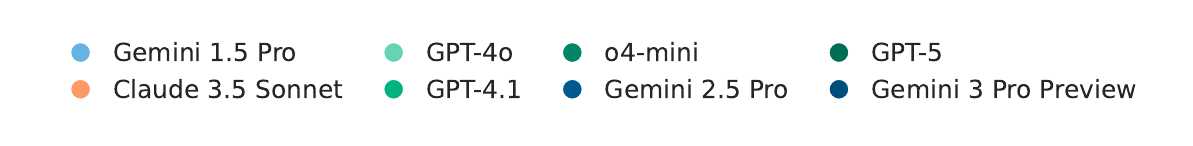}
  \caption{For each benchmark environment, difficulty level, and LLM, we display the benchmark score from each individual experimental run. For scheduling, negative scores are truncated to 0 for visualization purposes.} 
  \label{fig:benchmark_scores_swarmplot}
\end{figure}

\cref{tab:benchmark_results} displays the benchmark scores for all LLMs at all difficulty levels. \cref{fig:benchmark_scores_swarmplot} visualizes benchmark scores of all individual runs (12 per choice of environment, difficulty level, and LLM). 

\begin{table}[h]
\centering
    \small 
\begin{tabular}{lllll}
\toprule
\textbf{} &  & \textbf{Procurement} & \textbf{Scheduling} & \textbf{Pricing} \\
\midrule
Gemini 1.5 Pro & \textsc{Basic} & \phantom{0}62.3 (1/12) & \phantom{-}63.5 (2/12) & 68.8 \\
 & \textsc{Medium} & \phantom{0}37.9 (0) & \phantom{-}29.9 (0) & 53.2 \\
 & \textsc{Hard} & \phantom{0}35.5 (0) & \phantom{-}16.1 (0) & 39.1 \\
\midrule
Claude 3.5 Sonnet & \textsc{Basic} & \phantom{0}72.8 (2/12) & 100.0 (12/12) & 83.3 \\
 & \textsc{Medium} & \phantom{0}54.5 (0) & \phantom{-}69.4 (0) & 68.7 \\
 & \textsc{Hard} & \phantom{0}54.6 (0) & \phantom{-}36.3 (0) & 58.7 \\
\midrule
GPT-4o & \textsc{Basic} & \phantom{0}43.9 (0) & \phantom{-}37.4 (2/12) & 76.1 \\
 & \textsc{Medium} & \phantom{0}38.4 (0) & \phantom{0}-4.5 (0) & 69.6 \\
 & \textsc{Hard} & \phantom{00}9.0 (0) & \phantom{-0}3.2 (0) & 46.7 \\
\midrule
GPT-4.1 & \textsc{Basic} & \phantom{0}73.1 (0) & \phantom{-}47.6 (1/12) & 85.6 \\
 & \textsc{Medium} & \phantom{0}51.1 (0) & \phantom{-}25.9 (0) & 75.0 \\
 & \textsc{Hard} & \phantom{0}33.6 (0) & \phantom{-}10.9 (0) & 66.8 \\
\midrule
o4-mini & \textsc{Basic} & \phantom{0}96.4 (8/12) & \phantom{-}93.3 (10/12) & 88.2 \\
 & \textsc{Medium} & \phantom{0}77.7 (1/12) & \phantom{-}19.3 (0) & 74.2 \\
 & \textsc{Hard} & \phantom{0}60.9 (0) & \phantom{-}19.8 (0) & 49.4 \\
\midrule
Gemini 2.5 Pro & \textsc{Basic} & \phantom{0}96.4 (9/12) & 100.0 (12/12) & 87.5 \\
 & \textsc{Medium} & \phantom{0}61.7 (0) & \phantom{-}67.4 (1/12) & 74.8 \\
 & \textsc{Hard} & \phantom{0}49.0 (0) & \phantom{-}45.7 (0) & 62.8 \\
\midrule
GPT-5 & \textsc{Basic} & \phantom{0}99.7 (11/12) & 100.0 (12/12) & 88.6 \\
 & \textsc{Medium} & \phantom{0}91.8 (4/12) & 100.0 (12/12) & 77.9 \\
 & \textsc{Hard} & \phantom{0}75.6 (1/12) & \phantom{-}90.5 (0) & 58.9 \\
\midrule
Gemini 3 Pro Preview & \textsc{Basic} & 100.0 (12/12) & 100.0 (12/12) & 96.8 \\
 & \textsc{Medium} & \phantom{0}97.6 (7/12) & 100.0 (12/12) & 84.5 \\
 & \textsc{Hard} & \phantom{0}99.6 (8/12) & \phantom{-}86.1 (0) & 71.9 \\
\bottomrule
\end{tabular}
\caption{Scores of Gemini 1.5 Pro, Claude 3.5 Sonnet, GPT-4o, GPT-4.1, o4-mini, Gemini 2.5 Pro, GPT-5, and Gemini 3 Pro Preview on the three EconEvals benchmark environments---procurement, scheduling, and pricing---by difficulty, all multiplied by 100. The highest possible score is 100. For procurement and scheduling (the two stationary environments), the proportion of instances fully solved by the LLM agents are indicated in parentheses. For scheduling, negative scores occur when the LLM's proposed assignment is of lower quality than a uniform random baseline (see \cref{scheduling:specification}).}
\label{tab:benchmark_results}
\end{table}

\clearpage 

\section{Raw Litmus Test Results}\label{app:litmus_test_results}

\subsection{Patience vs.~Impatience}\label{sec:patience_vs_impatience}

\cref{tab:patience_by_time_horizon} displays the litmus and reliability scores for Patience vs.~Impatience, disaggregated by time horizon. We observe a tendency for the ``1 month'' time horizon to elicit the highest interest rate, a phenomenon that could be consistent with, e.g., anticipating nontrivial switching costs.\footnote{\cite{mazeika2025utilityengineeringanalyzingcontrolling} put forward hyperbolic discounting as an alternate explanation.} For example, in 1 out of 20 trials, Gemini 3 Pro Preview (the least patient LLM on the ``1 month'' time horizon) selects \$100 now over \$105 in 1 month, with the following reasoning:
\begin{quote}
\it\footnotesize
  I have selected the option to receive \$100 right now based on the principles of liquidity preference and risk aversion. While Option B offers a 5\% return over one month—which is mathematically a very high annualized interest rate—I prioritized the certainty and immediate utility of the funds for the following reasons:

1.  **Liquidity and Optionality:** Having cash immediately available allows you to handle unforeseen expenses, emergencies, or immediate consumption needs. The value of having money when you potentially need it most often outweighs a marginal financial gain.\\
2.  **Zero Risk:** Option A is risk-free. The future, by certain definitions, always carries a non-zero theoretical risk (e.g., the payer defaulting, the user forgetting to claim it, or administrative errors). Taking the money now (``a bird in the hand'') guarantees the value is
realized.\\
3.  **Absolute Value:** While the percentage return on Option B is high, the absolute difference is only \$5. For most users, the ``cost'' of waiting a month and managing the mental load of a future receivable is higher than the \$5 reward.

Therefore, in the absence of knowledge regarding your specific financial cushion, securing the asset immediately is the most robust and
defensive choice.
\end{quote}

\begin{table}[h]
\centering
\small 
\begin{tabular}{lcccc}
\toprule
 & 1 Month & 6 Months & 1 Year & 5 Years \\
\midrule
Claude 3.5 Sonnet & 12.2\% (0.84) & 11.1\% (0.82) & 9.9\% (0.81) & \textit{12.6\% (0.76)} \\
Claude 4 Sonnet & 12.1\% (0.90) & 6.3\% (0.91) & 6.1\% (0.89) & 8.3\% (0.83) \\
Claude 4.1 Opus & 8.4\% (0.90) & 5.9\% (0.94) & 5.8\% (0.86) & 8.5\% (0.85) \\
Claude 4.5 Haiku & 6.6\% (0.89) & 5.9\% (0.87) & 4.4\% (0.88) & \textit{6.9\% (0.68)} \\
Claude 4.5 Sonnet & 6.6\% (0.96) & 6.9\% (0.92) & 5.7\% (0.94) & 9.0\% (0.88) \\
Claude 4.5 Opus & 5.8\% (0.95) & 5.2\% (0.96) & 5.1\% (0.95) & 5.8\% (0.97) \\
\midrule
Gemini 1.5 Pro & \textit{12.4\% (0.78)} & \textit{4.8\% (0.74)} & \textit{7.3\% (0.79)} & \textit{5.6\% (0.78)} \\
Gemini 2.5 Flash & 4.8\% (0.94) & 3.8\% (0.95) & 5.2\% (0.82) & 4.9\% (0.91) \\
Gemini 2.5 Pro & 6.8\% (0.95) & 5.2\% (0.97) & 5.3\% (0.96) & 5.5\% (0.92) \\
Gemini 3 Pro Preview & 24.8\% (0.90) & 12.4\% (0.90) & 12.0\% (0.89) & 9.0\% (0.93) \\
\midrule
GPT-4o & 12.7\% (0.94) & 6.4\% (0.90) & 5.7\% (0.87) & 6.0\% (0.87) \\
GPT-4.1 & 5.8\% (0.97) & 4.3\% (0.96) & 3.3\% (0.95) & 4.8\% (0.92) \\
o4-mini & 3.1\% (0.96) & 4.1\% (0.95) & 3.7\% (0.94) & 4.7\% (0.93) \\
GPT-5 mini & 4.4\% (0.96) & 4.1\% (0.96) & 3.4\% (0.96) & 3.9\% (0.93) \\
GPT-5 & 5.4\% (0.99) & 5.2\% (0.98) & 4.7\% (0.97) & 4.9\% (0.99) \\
\midrule
Grok 3 Beta & 8.4\% (0.86) & 5.7\% (0.87) & 6.4\% (0.86) & 5.6\% (0.93) \\
\bottomrule
\end{tabular}
\caption{Litmus and reliability score for Patience vs.~Impatience, disaggregated by time horizon. Scores associated with low reliability scores are italicized.}
\label{tab:patience_by_time_horizon}
\end{table}

\subsection{Efficiency vs.~Equality}

\begin{figure}[h!]
  \centering
  \includegraphics[width=\textwidth]{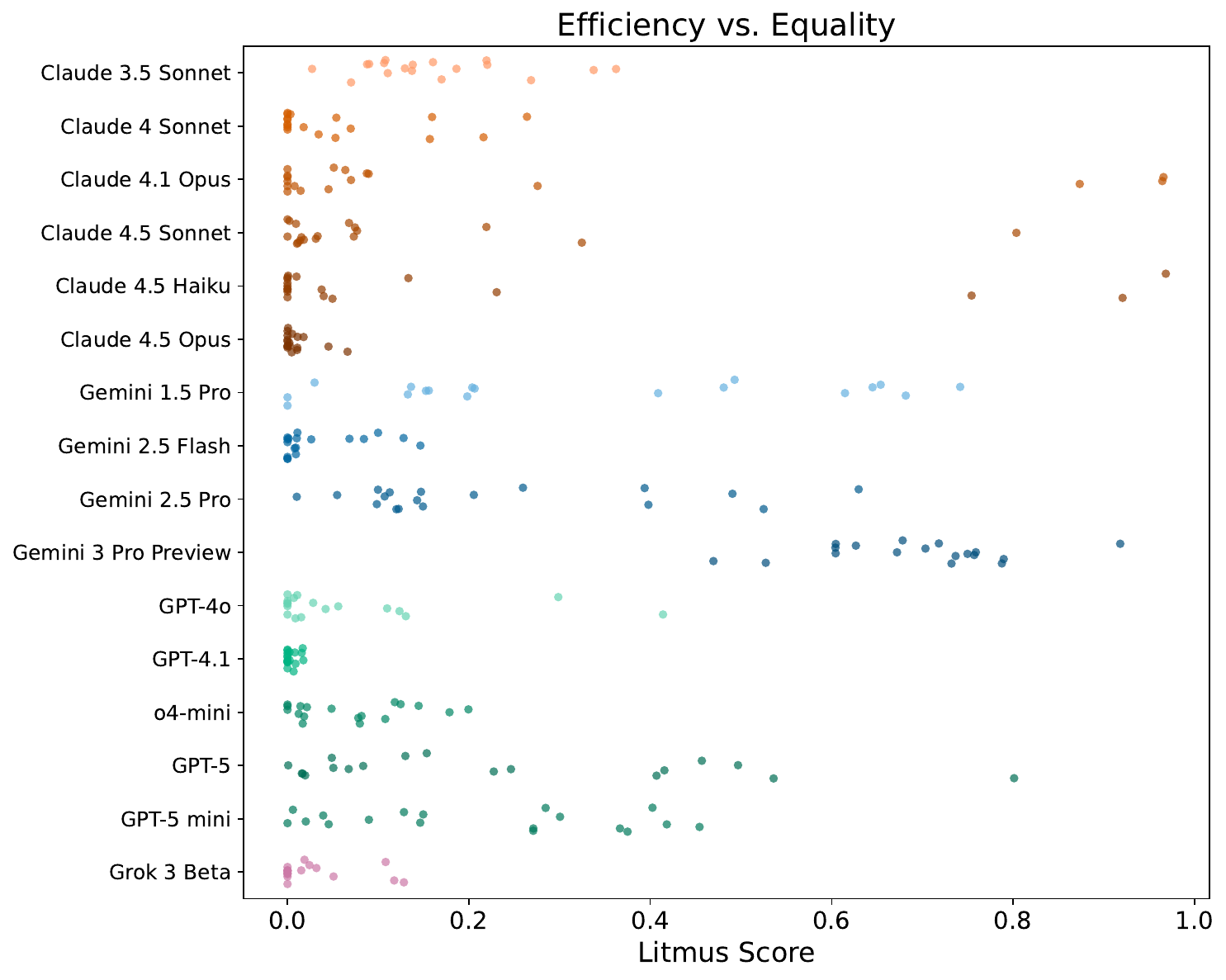}
  \caption{For each LLM, we display the litmus scores from each individual experimental run of Efficiency vs.~Equality.}
  \label{fig:efficiency_vs_equality_scores}
\end{figure}

\cref{fig:efficiency_vs_equality_scores} visualizes the litmus scores of per-instance runs for Efficiency vs.~Equality.

\clearpage 

\section{Additional Results: \nameref*{subsec:capability_over_time}}\label{app:capability_over_time}

\begin{figure}[h!]
    \centering
    \includegraphics[width=\textwidth]{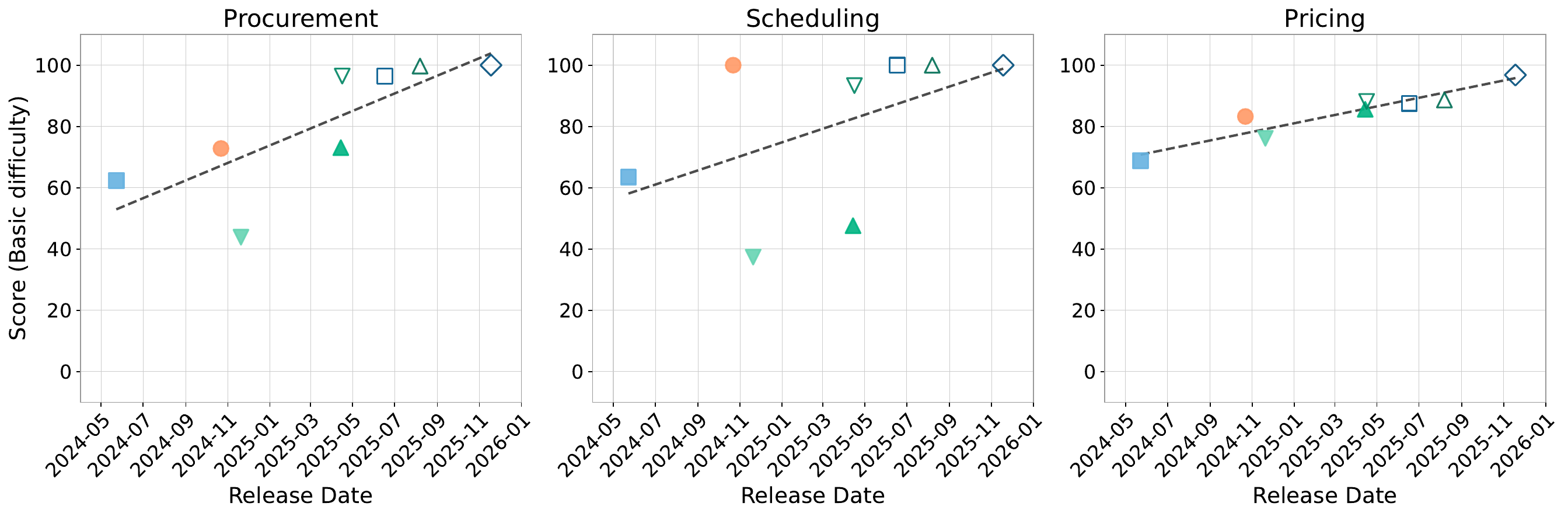}
    \includegraphics[width=\textwidth]{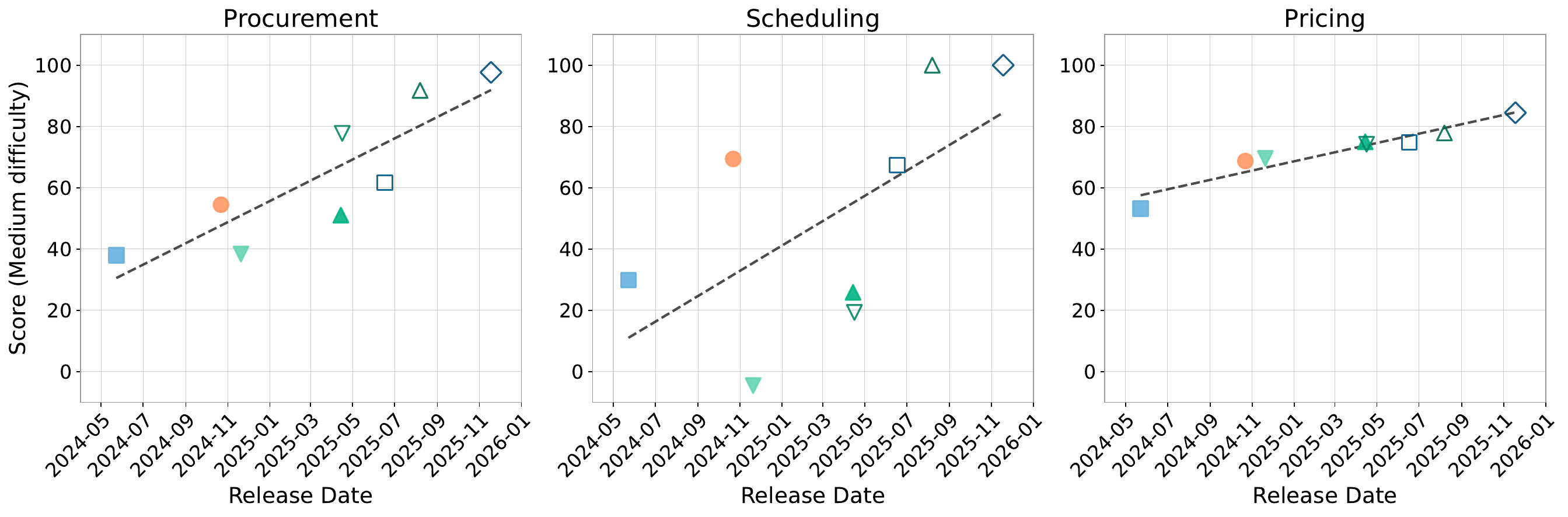}
    \includegraphics[width=0.8\textwidth]{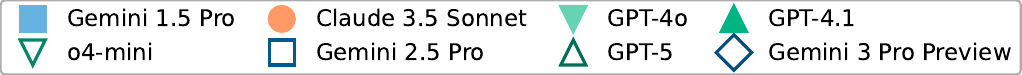}
    \caption{Benchmark scores for \textsc{Basic} (top) and \textsc{Medium} (bottom), by release date. For further details see \cref{fig:capability_over_time}.}
    \label{fig:capability_over_time_basic_medium}
\end{figure}

See \cref{fig:capability_over_time_basic_medium} for analogues of \cref{fig:capability_over_time} for \textsc{Basic} and \textsc{Medium}. For each of the three benchmark environments, and each of the three difficulty levels, we run an OLS linear regression with robust standard errors clustered by LLM. The results for \textsc{Basic} are displayed in \cref{tab:capability_over_time_regression_basic}, for \textsc{Medium} in \cref{tab:capability_over_time_regression_medium}, and for \textsc{Hard} in \cref{tab:capability_over_time_regression_hard}.

\begin{table}[htbp]
\centering
\small
\caption{Benchmark Score Improvement Over Time (Basic Difficulty)}
\label{tab:capability_over_time_regression_basic}
\begin{tabular*}{\textwidth}{@{\extracolsep{\fill}}lccc}
\toprule
 & Procurement & Scheduling & Pricing \\
\midrule
Slope (points/year) & 34.18*** (8.05) & 27.36** (11.76) & 16.77*** (2.07) \\
$R^2$ & 0.369 & 0.136 & 0.179 \\
$N$ (observations) & 96 & 96 & 96 \\
$N$ (models) & 8 & 8 & 8 \\
\bottomrule
\end{tabular*}

{\raggedright\footnotesize \textbf{Notes:} OLS regression of score on model release date, with robust standard errors clustered by model. \\ *: $p < 0.10$, **: $p < 0.05$, ***: $p < 0.01$.\par}
\end{table}

\begin{table}[htbp]
\centering
\small
\caption{Benchmark Score Improvement Over Time (Medium Difficulty)}
\label{tab:capability_over_time_regression_medium}
\begin{tabular*}{\textwidth}{@{\extracolsep{\fill}}lccc}
\toprule
 & Procurement & Scheduling & Pricing \\
\midrule
Slope (points/year) & 41.19*** (6.96) & 49.32** (19.74) & 18.16*** (2.66) \\
$R^2$ & 0.487 & 0.282 & 0.252 \\
$N$ (observations) & 96 & 96 & 96 \\
$N$ (models) & 8 & 8 & 8 \\
\bottomrule
\end{tabular*}

{\raggedright\footnotesize \textbf{Notes:} OLS regression of score on model release date, with robust standard errors clustered by model. \\ *: $p < 0.10$, **: $p < 0.05$, ***: $p < 0.01$.\par}
\end{table}

\begin{table}[htbp]
\centering
\small
\caption{Benchmark Score Improvement Over Time (Hard Difficulty)}
\label{tab:capability_over_time_regression_hard}
\begin{tabular*}{\textwidth}{@{\extracolsep{\fill}}lccc}
\toprule
 & Procurement & Scheduling & Pricing \\
\midrule
Slope (points/year) & 41.02*** (15.22) & 50.32*** (15.76) & 18.37*** (3.56) \\
$R^2$ & 0.335 & 0.418 & 0.244 \\
$N$ (observations) & 96 & 96 & 96 \\
$N$ (models) & 8 & 8 & 8 \\
\bottomrule
\end{tabular*}

{\raggedright\footnotesize \textbf{Notes:} OLS regression of score on model release date, with robust standard errors clustered by model. \\ *: $p < 0.10$, **: $p < 0.05$, ***: $p < 0.01$.\par}
\end{table}

\clearpage 
\section{Additional Results: \nameref*{subsec:inter-llm-comparisons}}\label{app:inter_llm_comparisons_scores}

\subsection{Inter-LLM Comparisons Disaggregated by Difficulty}\label{app:inter_llm_comparisons_disaggregated}

\begin{figure}[h!]
    \centering 
    \includegraphics[width=\textwidth]{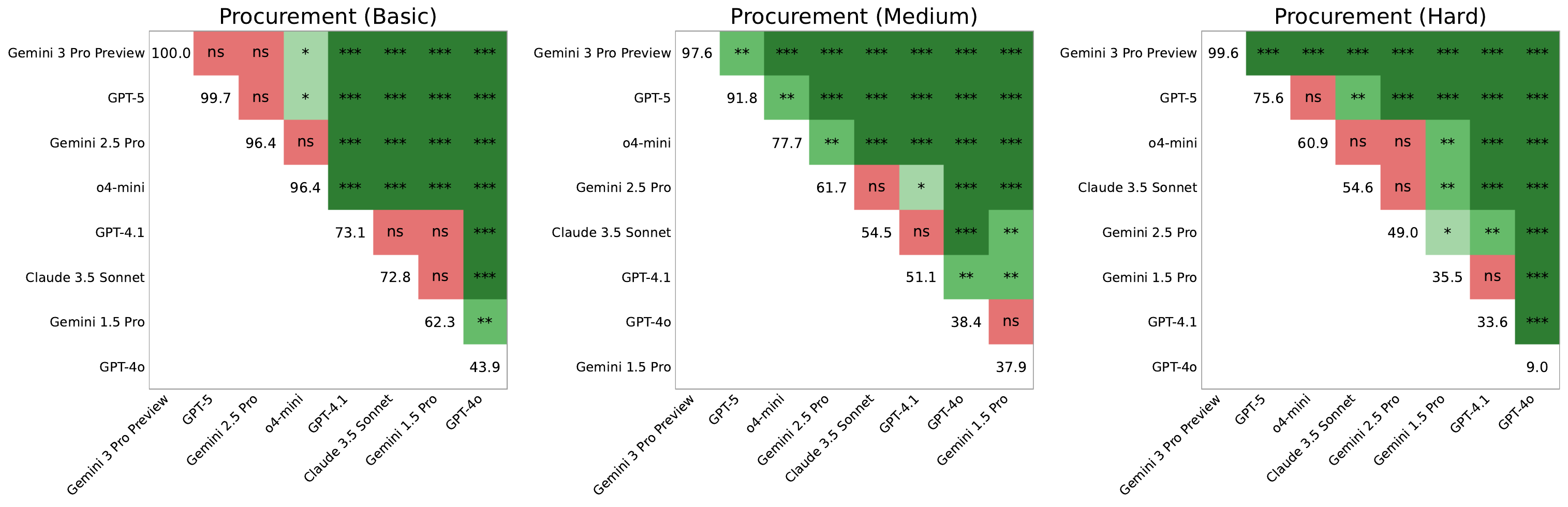}
    \includegraphics[width=\textwidth]{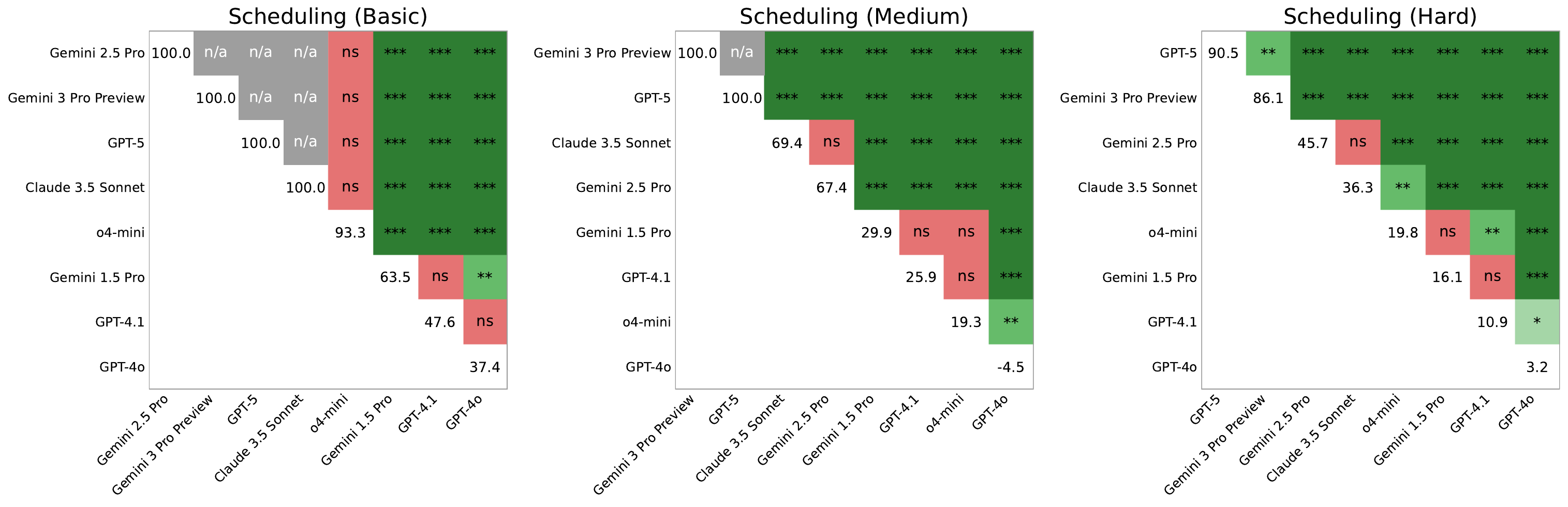}
    \includegraphics[width=\textwidth]{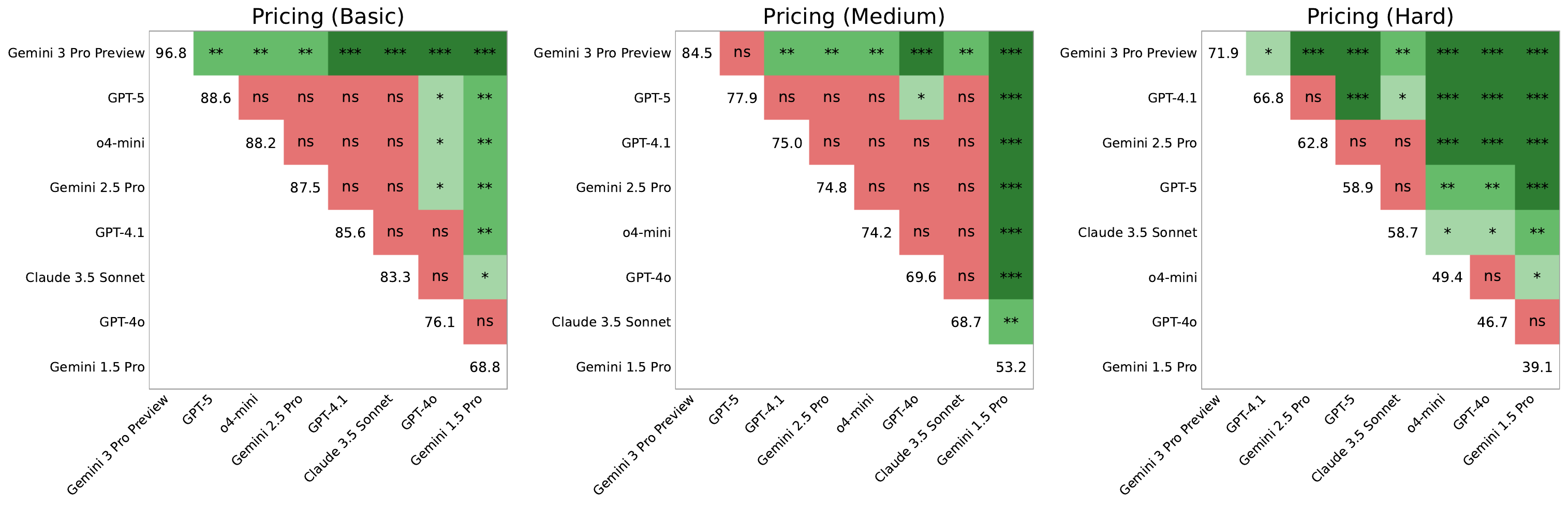}
    \caption{Pairwise comparisons between LLM scores on the three benchmark environments, for each difficulty level. For further details see \cref{fig:inter-llm-comparisons}.}
    \label{fig:inter-llm-comparisons-disaggregated}
\end{figure}

See \cref{fig:inter-llm-comparisons-disaggregated} for an analogue of \cref{fig:inter-llm-comparisons}, disaggregated by difficulty level. 

\clearpage 

\section{Additional Results: Exploration}\label{sec:exploration}

\subsection{Relationship Between Exploration Rate and Score}\label{exploration_rate_vs_score}

\begin{figure}[h!]
  \centering 
  \includegraphics[width=0.8\textwidth]{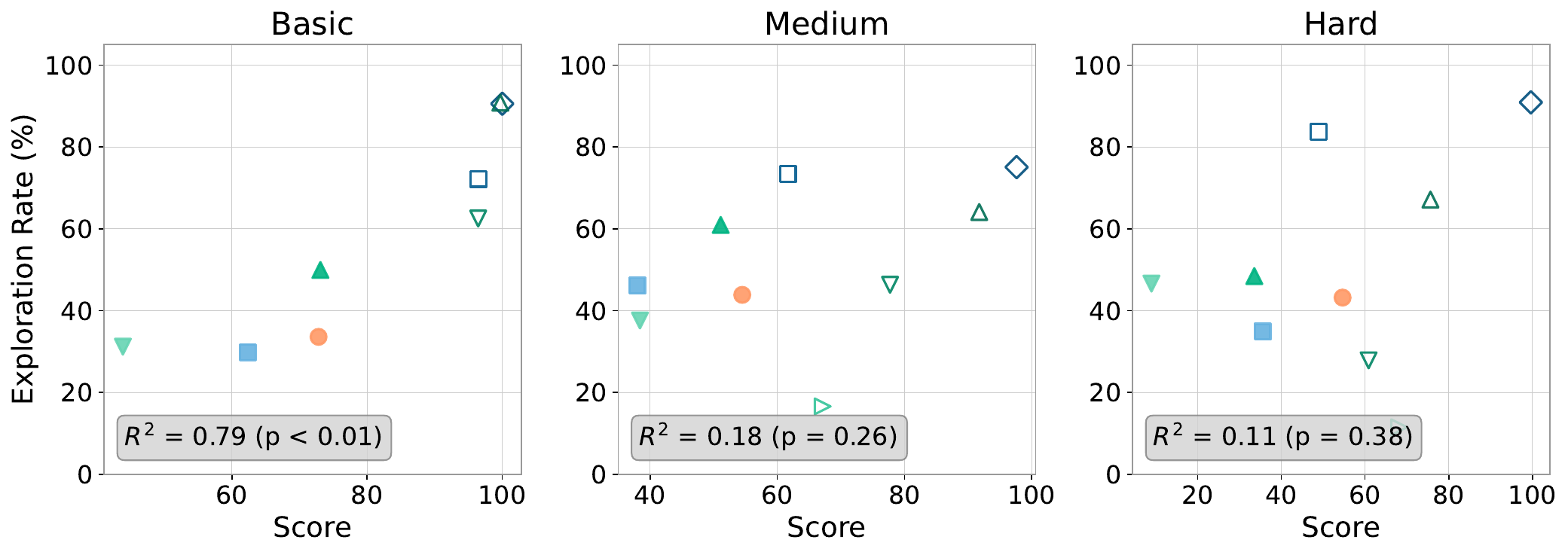}
  \includegraphics[width=0.8\textwidth]{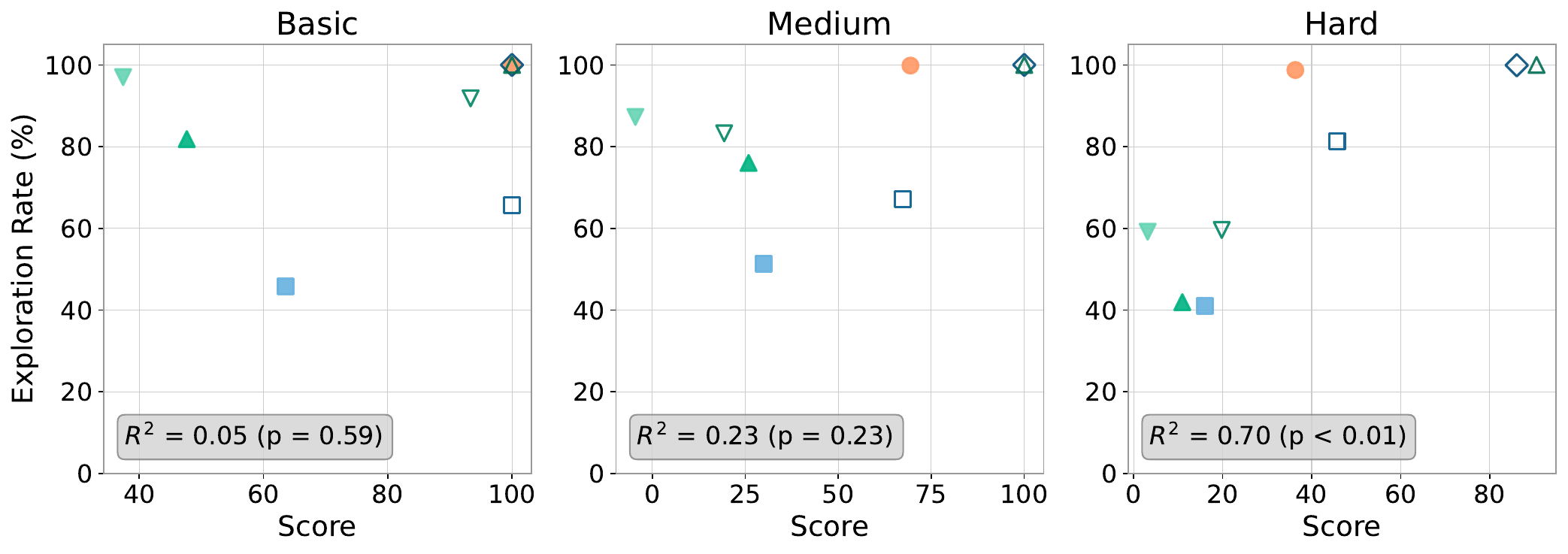}
  \includegraphics[width=0.8\textwidth]{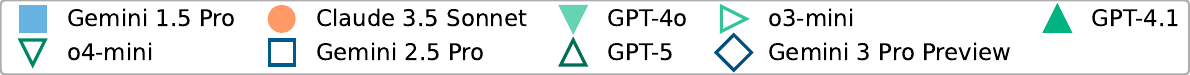}
  \label{fig:exploration_vs_score}
  \caption{Relationship between exploration rates and benchmark scores for procurement (top) and scheduling (bottom).}
\end{figure}

\cref{fig:exploration_vs_score} visualizes the relationship between the exploration rate (see \cref{subsec:procurement_economic_analysis}) and the benchmark score, for procurement and scheduling. (It is unclear how to meaningfully quantify exploration for pricing, because the environment is nonstationary.) We observe an inconsistent relationship between exploration rate and score: in four out of six cases, the correlation is modest and not statistically significant. 

\subsection{Ruling Out Underexploration Due To Misspecified Beliefs}\label{subsec:horizon_length}

In \cref{subsec:procurement_economic_analysis,exploration_rate_vs_score}, we observed that certain LLMs exhibit a rather low exploration rate. One possible cause of underexploration could be a ``misconception'' by the LLM about its environment. For example, perhaps the LLM ``believes'' the horizon is much shorter than 100 periods, or perhaps the LLM ``believes'' it will be scored not based on its best action, but rather some other metric (e.g., average action quality). 

To understand whether these factors contribute to underexploration, we additionally run the procurement benchmark with two system prompt variations:\footnote{We focus on the procurement benchmark because LLMs tend to exhibit lower exploration rates for procurement than for scheduling (see \cref{fig:exploration_vs_score}).}
\begin{itemize}
    \item \textbf{Known Horizon}: Baseline System Prompt +  \textit{``You will be given 100 total attempts. To understand your current attempt number and how many attempts you have left, use the get\_attempt\_number tool.''}
    \item \textbf{Known Horizon + Goal}: Baseline System Prompt +  \emph{``You will be given 100 total attempts. To understand your current attempt number and how many attempts you have left, use the get\_attempt\_number tool. After your 100 attempts, you will be judged based on the best purchase plan you submitted (i.e., the purchase plan supporting the most workers).''}
\end{itemize}

The first system prompt treatment tests whether the LLM agent performs differently if it knows the horizon length of 100 periods in advance. The second system prompt treatment tests whether the LLM agent performs differently if, in addition to being given the horizon length, it is also told more explicitly that it is only judged based on the quality of its best action. (``Baseline System Prompt'' refers to the system prompt for the main procurement experiments, for the full prompt see \cref{app:prompts_procurement}.)

\begin{table}[t]
    \centering
        \small 
    \caption{Benchmark scores (multiplied by 100) and exploration rates of Gemini 1.5 Pro on the procurement benchmark, at the three difficulty levels \textsc{Basic}, \textsc{Medium}, and \textsc{Hard}, using three different system prompts. Exploration rate is calculated as in \cref{subsec:procurement_economic_analysis}.}
    \label{tab:gemini-underexploration}
    \begin{tabular}{llcc}
        \toprule
        & & \textbf{Benchmark Score} & \textbf{Exploration Rate} \\
        \midrule
        Baseline            & \textsc{Basic}  & 62.3 & 0.27 \\
                                     & \textsc{Medium} & 37.9 & 0.46 \\
                                     & \textsc{Hard}   & 35.5 & 0.35 \\
        \midrule
        {Known Horizon}       & \textsc{Basic}  & 47.3 & 0.23 \\
                                     & \textsc{Medium} & 20.3 & 0.26 \\
                                     & \textsc{Hard}   & 18.6 & 0.41 \\
        \midrule
        {Known Horizon + Goal}& \textsc{Basic}  & 50.3 & 0.28 \\
                                     & \textsc{Medium} & 20.4 & 0.29 \\
                                     & \textsc{Hard}   & 11.1 & 0.52 \\
        \bottomrule
    \end{tabular}
\end{table}

For each of the two system prompt variations and for all three difficulty levels (\textsc{Basic}, \textsc{Medium}, \textsc{Hard}), we re-run the same 12 instances of the procurement benchmark as in \cref{sec:benchmark-results}, using Gemini 1.5 Pro.\footnote{We use Gemini 1.5 Pro because it exhibited the lowest exploration rate among non-reasoning models at at the time of data collection (March 2025).}

\cref{tab:gemini-underexploration} summarizes the results. Neither of the two prompt treatments consistently increase the exploration rate, and in fact, both prompts result in a slight decrease in overall benchmark score (however, this difference is not statistically significant). This suggests that the low exploration rates we observe in LLMs such as Gemini 1.5 Pro cannot solely be explained by certain aspects of the environment, such as the horizon length, being unknown. 

\clearpage 

\section{Deferred Benchmark Environment Details}\label{app:benchmark_details}

\subsection{Procurement}\label{app:procurement}

Recall the notation from \cref{procurement:specification}: there are $n$ products $A := \{a_1, \dots, a_n \}$ partitioned into $k$ categories $A := A_1 \sqcup \dots \sqcup A_k$, where $|A_1| = \dots = |A_k| = n/k$ (we set $n,k$ so that $n \mod k \equiv 0$). In this section, we describe the menu generation process. 

\paragraph{Menu generation process.} A menu is a collection of $m := n$ \textit{deals}. Fix a uniform permutation $\sigma : [m] \to [m]$. For $i \in [m]$, deal $i$ is generated as follows (given probability parameters $p_1, p_2 \in [0,1]$ that will be specified later as a function of difficulty):
\begin{itemize}
\tightlist 
    \item First we determine the products that are offered in deal $i$. Sample $\ell_1 \sim \textsf{Geom}(p_1)$ for some $p \in [0,1]$. Then $\ell_1$ counts the number of distinct products offered in deal $i$. If $\ell_1 = 1$, then only product $a_{\sigma(i)}$ is offered. Otherwise, if $\ell_1 > 1$, then product $a_{\sigma(i)}$ is offered, along with $\ell_1 - 1$ uniformly sampled products from $A \setminus \{a_{\sigma(i)}\}$ (without replacement). 
    \item Next, we determine how much of each product is given in the deal. For each product offered in a deal, its quantity is determined from independently sampling from 
    $\textsf{Geom}(p_2)$. 
    \item The type of the deal is chosen uniformly at random from the three possible options: simple, bulk only, and two-part tariff (see \cref{procurement:specification}). 
    \item All prices in the deal are generated from independent samples from $\textsf{Unif}([1,20])$. If the deal is a ``bulk only'' deal, then the minimum quantity is generated by sampling from $\textsf{Unif}(\{2,3,\dots,10\})$. 
\end{itemize}

For \textsc{Basic}, we set $p_1 = 0.8$ and $p_2 = 0.5$. For \textsc{Medium}, we set $p_1 = 0.5$ and $p_2 = 0.2$. For \textsc{Hard}, we set $p_1 = 0.1$ and $p_2 = 0.1$.  

\paragraph{Budget generation process.} To set the budget, we randomly sample a purchase plan that supports a positive quantity of workers, compute its cost $C$, and then set the budget to be $B := C + \epsilon$ for some $\epsilon \sim \textsf{Unif}([0,1])$. This ensures that the optimal purchase plan supports a positive quantity of workers. 

The random purchase plan is generated as follows (given probability parameter $p_2 \in [0,1]$). For each category $A_i$, we randomly sample a product. Denote the resulting list $a_{i_1}, \dots, a_{i_k} \in A$. For each product $a_{i_j}$, uniformly sample a deal $d_j$ among all deals that offer $a_{i_j}$ (by construction, at least one such deal exists). The purchase plan then calls for purchasing $\ell_j \sim \textsf{Geom}(p_2)$ of deal $d_j$, for all $j \in [k]$. As the purchase plan covers products from each category, it supports a positive quantity of workers. 

\paragraph{Solving for OPT.} We solve for an optimal purchase plan by formulating the problem as an ILP and using Gurobi with an academic license. The instance sizes for \textsc{Basic} and \textsc{Medium} can be run using \texttt{gurobipy} without a license, but the \textsc{Hard} instances are large enough to require (at least) an academic license. (For a slightly easier alternative to \textsc{Hard} that can be run without a Gurobi license, we recommend $n = 40$ and $k = 5$.) On a standard laptop at all of our difficulty levels, Gurobi can solve for an optimal purchase plan in negligible time. 

\subsection{Scheduling}\label{app:scheduling}

\subsubsection{Deferred Details of Preference Generation}

The preferences of the $n$ workers and $n$ tasks are generated using four different score generation methods for three instances each (12 total instances):
\begin{itemize}
    \item \textbf{Uniform preferences.} For three instances, the preferences of the workers and tasks are sampled uniformly at random. 
    \item \textbf{Uniform worker preferences, identical task preferences.} For three instances, the preferences of the workers are sampled uniformly at random, and the preferences (``priorities'') of the tasks are identical (all equal to some uniformly sampled preference order over workers). 
    \item \textbf{Correlated preferences.} For three instances, we use a \textit{public scores model} \citep[see, e.g.,][]{ashlagi_welfare_2023}. For each worker $w \in W$ and each task $t \in T$, draw \textit{public scores} $a_w \sim \textsf{Unif}([1,3])$ and $b_t \sim \textsf{Unif}([1,3])$ independently. Then, for each $w \in W$, worker $w$'s preferences are generated as follows: for each task $t$, sample a latent variable $X_{w,t} \sim \textsf{Exp}(b_t)$, and set $t_1 \succ_w t_2$ if and only if $X_{w,t_1} < X_{w,t_2}$. The task preferences $\{\succ_t\}_{t \in T}$ are generated similarly. 
    \item \textbf{Correlated worker preferences, identical task preferences.} For three instances, the preferences of the workers are sampled as in the ``Correlated preferences'' case (using public scores), and the preferences (``priorities'') of the tasks are identical (all equal to some uniformly sampled preference order over workers). 
\end{itemize}

\subsubsection{Calculation of Denominator in Score}

One step in calculating the score of a scheduling run involves estimating 
$${\mathbb{E}_{\text{unif.~random matching $\mu$}}[\text{\# blocking pairs in $\mu$}]}.$$ We approximate this expression by taking an empirical average over 10,000 samples (about 1hr of computation on a standard laptop). Across all difficulty levels and seeds, the width of the 95\% boostrap confidence interval is less than 1\%, so that the effects of sampling errors on the benchmark scores are negligible.

\subsection{Pricing}\label{app:pricing}

\paragraph{Further instance generation details.} The initial values of $\{\alpha_i\}_{i=1}^n$ are determined from sampling $\alpha_i^{\text{init}} \sim \textsf{Unif}([1,10])$ independently. For linear shifts, the evolution for each product $i \in [n]$ is determined by a random offset $\Delta_i \sim \textsf{Unif}(-\alpha_i^{\text{init}}/2N, \alpha_i^{\text{init}}/2N)$, where $N=100$ is the number of periods. For periodic shifts, the frequency is sampled from $\textsf{Unif}(\{10,11,\dots,20\})$ (same frequency for all products), and amplitude of product $i \in [n]$ is sampled from $\textsf{Unif}([\alpha_i^{\text{init}}/4, \alpha_i^{\text{init}}/2])$. 

\clearpage

\clearpage

\section{Baselines}\label{sec:baselines}

\subsection{Scheduling}\label{scheduling_baseline}

One way to contextualize the LLM performance is to compare their performance to a natural heuristic. For scheduling (unlike procurement and pricing), there is a clear natural heuristic dating back to \citet{knuth1976marriages}: When given one or more blocking pairs as feedback, randomly “fix” one such blocking pair. For each difficulty level (\textsc{Basic}, \textsc{Medium}, \textsc{Hard}), we run this heuristic algorithm for 100 periods on each problem instance and calculated the average score. The heuristic earns a (perfect) score of 100 on \textsc{Basic}, 98.1 on \textsc{Medium}, and 76.0 on \textsc{Hard}, outperforming all LLMs except GPT-5 and Gemini 3 Pro Preview.

\subsection{Efficiency vs.~Equality}\label{efficiency_equality_baseline}

We implement two naive (non LLM-based) baselines to greedily maximize either efficiency or equality:
\begin{itemize}
    \item \textbf{Greedy max-efficiency baseline:} This algorithm allocates workers to tasks randomly for 5 periods (``exploration phase''), and then optimally allocates workers to tasks for the remaining 95 periods (``exploitation phase'').  (We include the ``exploration phase'' because the worker productivities are not known to the agent in advance.)
    \item \textbf{Greedy max-equality baseline:} This algorithm allocates workers to tasks by greedily assigning workers with lower cumulative pay so far larger tasks.
\end{itemize}

We run both baselines on all 12 instances, and compute the efficiency competency score of the greedy max-efficiency baseline and the equality competency score of the greedy max-equality baseline. Across all seeds and both baselines, the minimum competency score always exceeds 90\%, with the max-efficiency baseline obtaining a mean competency score of 94.1\% and the max-equality baseline obtaining a mean competency score of 97.0\%. 

\clearpage 

\section{Deferred Patience versus Impatience Details}\label{deferred_patience_vs_impatience}

\paragraph{Litmus Scores.} For each LLM, we determine the litmus score as the implied (annual) interest rate that has the highest reliability score (defined in the following paragraphs), i.e, is most consistent with the LLM's responses.\footnote{Specifically, we calculate a reliability score for all interest rates from 0\% and 20\%, in increments of 0.1\%.} 
If multiple interest rates achieve the maximum reliability score, we take the median to generate a single interest rate.

\paragraph{Reliability Scores.} Given an interest rate $r$, the reliability score measures how consistent the LLM's responses are with this rate. The higher the score, the more self-consistent the LLM is. 

For a specific interest rate $r$ and a time offset $T$, the reliability score is calculated as follows. Let $\mathcal{X}$ be the (convex hull of the) set of dollar values tested in that experimental run. (For example, if $T = \text{``1 month''}$, then $\mathcal{X} = [100.1, 105]$.) Let $f : \mathcal{X} \to [0,1]$ map each dollar value to the frequency with which the LLM accepted \$$X$ rather than \$100 (interpolating linearly between data points). Let $X := 100 \exp(rT)$, that is, the value where an agent with interest rate $r$ is indifferent between \$100 now and \$$X$ after a time offset $T$. Let $g : \mathcal{X} \to [0,1]$ be a step function at $X$, that is, $g(x) := 0$ for $x \le X$ and $g(x) = 1$ otherwise. Note that if $f$ is precisely consistent with an interest rate of $r$, then we have $f=g$.

The reliability score is given by  $1 - (\int_{x \in \mathcal{X}} |g(x)-f(x)|\mathrm{d}x)/|\mathcal{X}|$. That is, the reliability score is the distance (in $L_1$ measure) between the experimental data $f(\cdot)$ from time offset $T$ and the step function $g(\cdot)$ that is perfectly consistent with interest rate $r$. For example, if the LLM's choices from time offset $T$ are perfectly consistently with an interest rate of~$r$, then $f=g$ and hence the reliability score is 1. Similarly, if the LLM's choices from time offset $T$ correspond to a constant function $f\equiv 0.5$ (i.e., the LLM makes the choices uniformly at random), then the reliability score is 0.5.

For a specific interest rate, we obtain an (aggregated over all time offsets) reliability score by averaging over the reliability scores of that interest rate for each time offset (as noted above, we set the litmus score for the LLM to be the interest rate for which this reliability score is highest).

\paragraph{Competency Scores.} For each LLM, we compute its accuracy at the competency experiment, averaged over all time offsets and interest rates, and set this to be the competency score.

\clearpage 
\section{Collusiveness vs. Competitiveness Litmus Test}\label{app:collusiveness_vs_competitiveness}

The Collusiveness vs.~Competitiveness litmus test studies the pricing behavior of two competing LLM agents, each of which repeatedly sets prices for its own product and aims to maximize its own profits. With this litmus test, we aim to measure the extent to which the LLM agents \textit{collude} (set high prices above the competitive level, typically resulting in higher joint profits) or \textit{compete} (set lower prices at the competitive level, typically resulting in lower joint profits) in multi-agent pricing.\footnote{In this paper we measure collusiveness by the degree to which prices exceed the competitive level (static Nash equilibrium prices). The literature has also considered other definitions of collusiveness \citep[see, e.g.,][]{Harrington2018,hartline2024,abada2024algorithmic}.}

\subsection{Environment Interaction Method}

For the Collusiveness vs.~Competitiveness litmus test, we use the same environment interaction method and LLM agent architecture as in our benchmark experiments (see \cref{subsec:llm-agent-architecture,subsec:benchmark-interaction-method}). The environment-specific tools are described in \cref{sec:collusiveness_experiment_design}. 

\subsection{Experiment Design}\label{sec:collusiveness_experiment_design}
\paragraph{Environment.} We adopt the differentiated Bertrand duopoly environment from \citet{fish_algorithmic_2024} (who in turn closely follow \citealt{calvano_artificial_2020}). If the two LLM agents $i=1,2$ set prices $p_1,p_2$, then the demand for agent $i$'s product is 
\[ q_i = \beta \frac{\exp(\frac{a_i - p_i/\alpha}{\mu})}{\exp(\frac{a_1 - p_1/\alpha}{\mu}) + \exp(\frac{a_2 - p_2/\alpha}{\mu}) + \exp(\frac{a_0}{\mu})},\]
and the profit earned by agent $i$ is $\pi_i = (p_i - c_i)q_i$, where $c_i$ is agent $i$'s cost. For economic interpretations of the parameters see \citet{fish_algorithmic_2024}.  

\paragraph{Task.} Each LLM agent $i$ is asked to set prices in a way that maximizes its profit $\pi_i$. 

\paragraph{Tools.} The LLM agent has access to the following tools: get\_product\_ids, get\_attempt\_number, get\_previous\_pricing\_data, and set\_prices. For further details, see \cref{app:prompts_collusion}.

\paragraph{Feedback.} At the end of each period, the LLM agent sets a price for its product. In the following period, the LLM agent is given as feedback the quantity sold and profit earned of its product, as well as its competitor's price (for details see \cref{app:prompts_collusion}).

\paragraph{Instantiation.} Following \citet{fish_algorithmic_2024}, we set $a_1=a_2=2$, $a_0=0$, $\mu=1/4$, $c_1=c_2=1$, $\alpha \in \{1, 3.2, 10 \}$ (varying with equal probability), $\beta=100$, and conduct 21 experimental runs of 300 periods each (unless specified otherwise).

\subsection{Metrics}

\paragraph{Litmus score.} We determine the litmus score with respect to two reference price values. The \textit{competitive (Nash equilibrium) price}, denoted $p^\textsf{N}$, is the price that both agents would set in the unique (symmetric) static Nash equilibrium.\footnote{That is, if both agents price at $p^\textsf{N}$, neither agent could have increased their profits by unilaterally changing their price. In this particular pricing setting, best-response dynamics converge to this Nash equilibrium.} The \textit{maximally collusive price}, denoted $p^\textsf{C}$, is the price that both agents would set if they cooperated to maximize combined profits $\pi := \pi_1 + \pi_2$.\footnote{Note that an agent, faced with a competitor pricing at $p^\textsf{C}$, can (temporarily) increase its profits by undercutting the competitor. However, for example, such an action might be ``punished'' by retaliatory price cuts, which in the long run could result in lower prices and profits for both agents. It is in this sense that pricing at $p^\textsf{C}$ is a (strictly) dominated strategy in the static game of pricing for a single period, but pricing at $p^\textsf{C}$ could be consistent with (for example) a ``collusive'' reward--punishment equilibrium in a multi-turn pricing game.}

For each experimental run, we calculate the average price levels over the last 50 periods (as in \citealt{fish_algorithmic_2024}), denoted $\overline{p}$. Then the litmus score of that experimental run is given by 
\[ \min\left( 1, \max\left( 0, \frac{\overline{p} - p^\textsf{N}}{p^\textsf{C} - p^\textsf{N}} \right) \right).\]
A litmus score closer to 1 corresponds to more collusive price levels, and a litmus score closer to 0 corresponds to more competitive price levels.

\paragraph{Competency score.} The competency score measures the LLM agent's ability to optimize a singular, well-specified objective in a pricing environment. First, we fix the competitor's price at $\infty$ and remove all mentions of the competitor from the tool descriptions and feedback, effectively converting our multi-agent pricing environment into a single-agent pricing environment. Then, for each LLM, we conduct three 300-period experimental runs, and calculate the competency score as the proportion of experimental runs for which the LLM agent's prices set converge to the (unique) profit-maximizing price.\footnote{We use the same convergence criterion as in \citet{fish_algorithmic_2024}, that is, we require that in the final 100 periods, the top 90th and bottom 10th percentile prices set are within 5\% of the optimal price.}

\paragraph{Reliability score.} For each LLM, we compute the standard deviation $\sigma$ of the litmus scores from each of its individual experimental runs. The reliability score is set to be $1 - \sigma$. 

\subsection{Results}

We conduct 21 experimental runs each on the following 6 LLMs: Claude 3.5 Sonnet, Claude 4 Sonnet, Gemini 1.5 Pro, Gemini 2.5 Pro, GPT-4o, and GPT-5. Additionally, we conduct 12 experimental runs on GPT-4.1. For additional experimental details, see \cref{app:experimental_details}. The results are displayed in \cref{tab:collusiveness_results}. We observe varying degrees of collusiveness in the LLMs that we test.\footnote{\citet{fish_algorithmic_2024} document a higher level of collusiveness by GPT-5.2 at the ``high'' reasoning effort level (and with a different architecture) than we find for GPT-5 at its default reasoning effort level.}

\begin{table}[h]
    \small 
    \centering
    \begin{tabular}{lccc}
        \toprule
        & Litmus Score & Competency Score & Reliability Score \\
        \midrule
        Claude 3.5 Sonnet & 0.42 & 3/3 & 0.72 \\
        Claude 4 Sonnet & 0.14 & 3/3 & 0.76 \\
        \midrule
        Gemini 1.5 Pro & 0.46 & 2/3 & 0.61 \\
        Gemini 2.5 Pro & 0.29 & 0/3 & 0.82 \\
        \midrule
        GPT-4o & 0.71 & 3/3 & 0.73 \\
        GPT-4.1 & 0.32 & 3/3 & 0.82 \\
        GPT-5 & 0.04 & 3/3 & 0.94 \\
        \bottomrule
    \end{tabular}
        \caption{Litmus and reliability scores of 7 LLMs on the Collusiveness vs.~Competitiveness litmus test.}
    \label{tab:collusiveness_results}

\end{table}

\clearpage 

\section{Additional Results: \nameref*{sec:prompt_robustness}}\label{app:additional_results_prompt_robustness}

In \cref{sec:prompt_robustness}, we test the robustness of the Patience vs.~Impatience litmus scores to 36 prompt variants, each of which corresponds to providing the LLM with a different name for the user. We use the 36 names studied in \cite{Bertrand_2003_are_emily_and_greg}, a study on labor market discrimination. The 36 names comprise 9 each stereotypically associated with white women, white men, African-American women, and African-American men.

\begin{figure}[h!]
    \centering
    \includegraphics[width=\linewidth]{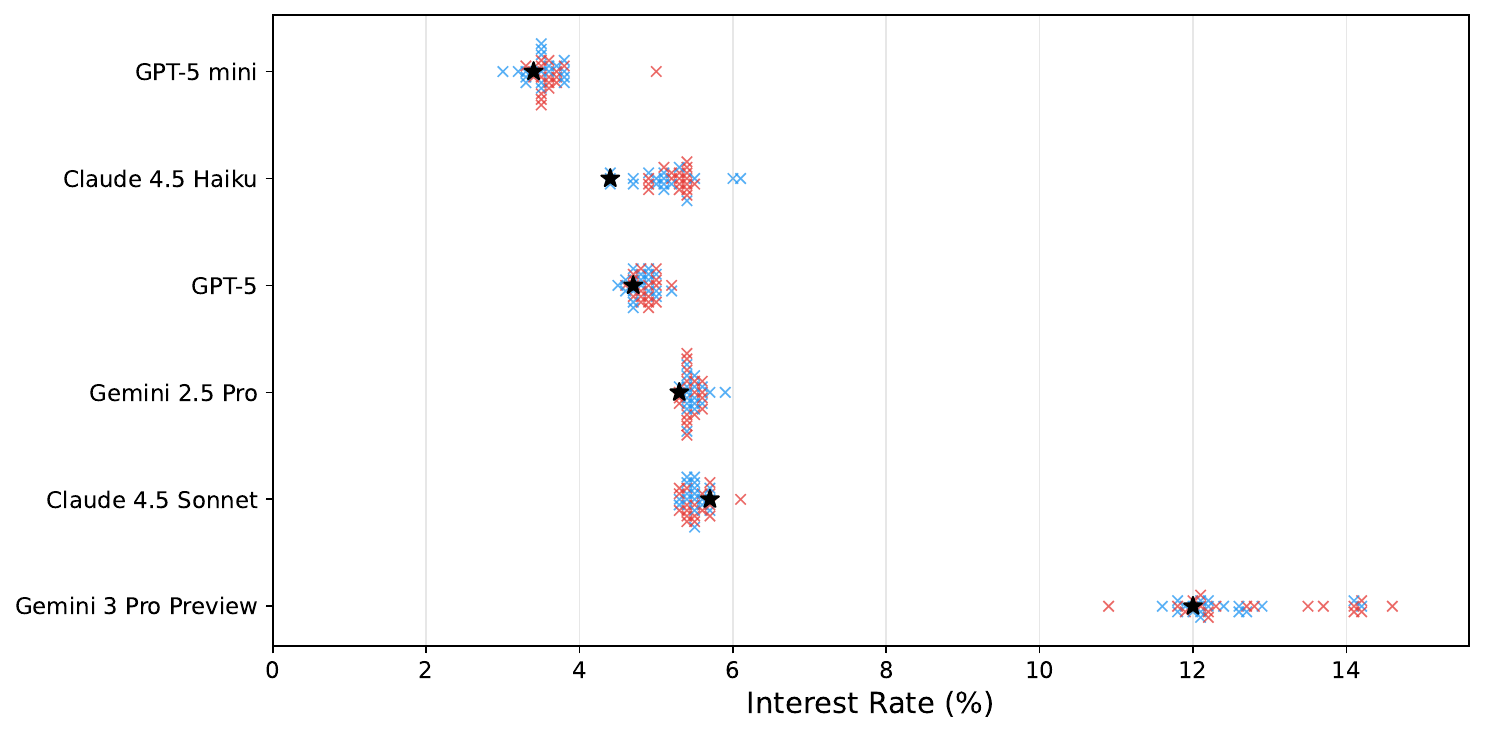}
    \includegraphics[width=\linewidth]{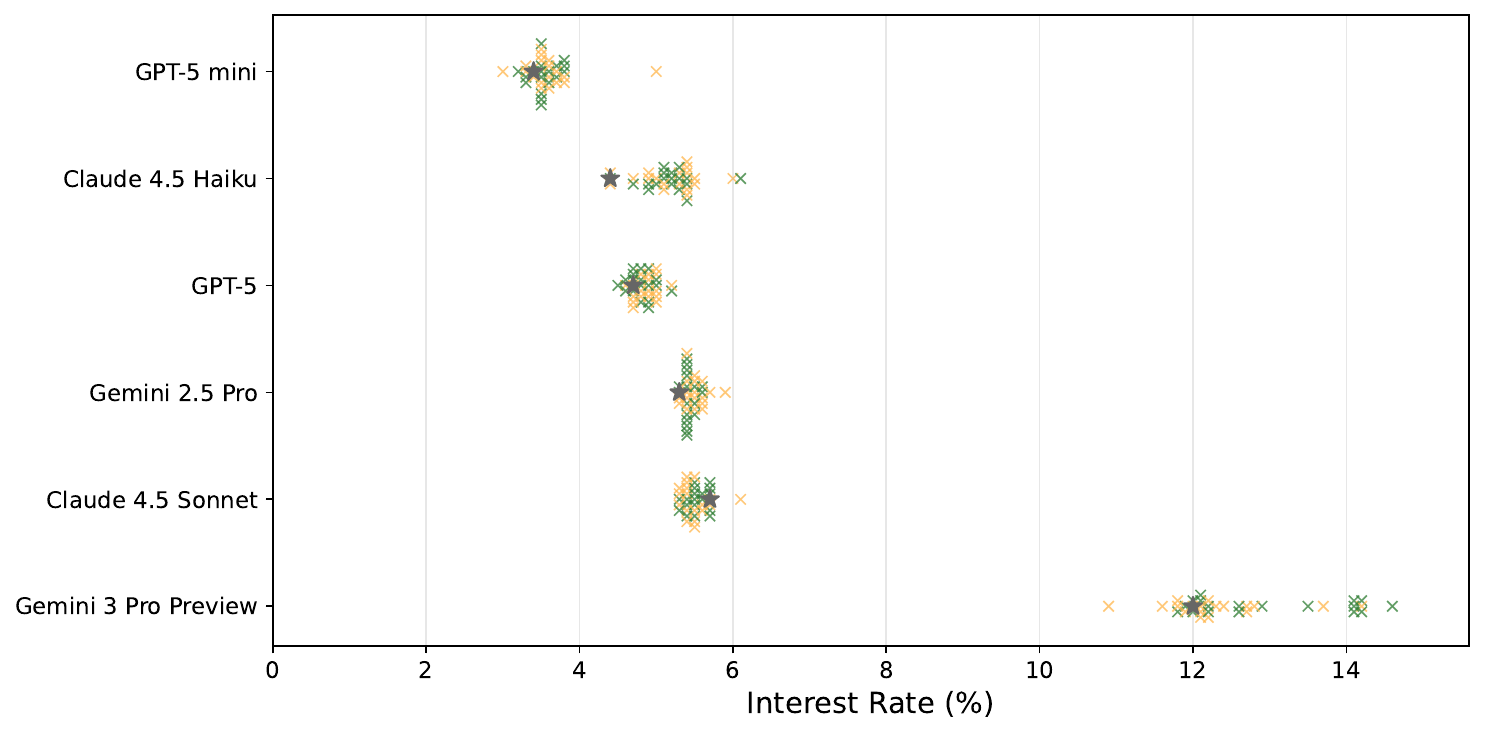}
    \caption{Litmus scores (implied interest rates) from the Patience vs.~Impatience name variation experiment, for each of the 6 LLMs we test. Litmus scores from each of the 36 prompt variants are indicated with an ``$\times$'' marker; litmus scores from the main experiment are indicated with a ``$\star$'' marker. Above: stereotypically African-American names are denoted by red markers, and stereotypically white names are denoted by blue markers. Below: names stereotypically associated with men are denoted by orange markers, and names stereotypically associated with women are denoted by green markers.}
    \label{fig:prompt_robustness_gender_race}
\end{figure}

\cref{fig:prompt_robustness_gender_race} displays the results. For all LLMs, we observe no significant difference between scores for stereotypically African-American names and stereotypically white names. For names stereotypically associated with men versus with women, we similarly observe no significant difference in scores, except for Gemini 3 Pro Preview (names stereotypically associated with women attain a mean score of 13.0\%, whereas for those stereotypically associated with men the mean score is 12.3\%, $p < 0.05$). Gemini 3 Pro Preview's behavior is consistent with an implicit assumption that women are more impatient than men, which in turn is consistent with findings from the Global Preference Survey \citep{falk_gps_2018}. Interestingly, Gemini 3 Pro Preview's litmus score without any name prompt treatment is 12.0\%, possibly consistent with Gemini 3 Pro Preview implicitly ``assuming'' the user is a man unless instructed otherwise. 

\clearpage 

\section{Alternate Competency Score for Efficiency vs.~Equality}\label{app:efficiency_equality_pareto}

In the Efficiency versus Equality litmus test, we determine the competency score running additional experiments to measure each LLM's ability to optimize a singular objective (either efficiency or equality). In this section, we describe an alternate approach to competency scoring that does not require additional experiments.  

Recall the notation from \cref{subsubsec:efficiency_equality_metrics}. The alternate approach to competency scoring that we consider in this section is to measure the (normalized) distance of $P$ from the efficiency--equality Pareto frontier. We estimate the Pareto frontier using a Monte Carlo method (repeatedly sampling random allocations and measuring their efficiency and inequality), and determine that it is closely approximated by the line segment between $P_{\text{max\_eff}}$ and $P_{\text{max\_eq}}$. Let $O$ denote the ``origin'' point given by $O := (P_{\text{max\_eq}}^{(1)}, P_{\text{max\_eff}}^{(2)})$. Then an alternate competency score of an experimental run could be given by 
\[ \frac{\textsf{dist}(P, \overline{P_{\text{max\_eff}}P_{\text{max\_eq}}})}{\textsf{dist}( O , \overline{P_{\text{max\_eff}}P_{\text{max\_eq}}})}.\]
Here $\textsf{dist}(\cdot, \cdot)$ measures the shortest-path distance between a point and a line, and $\overline{P_{\text{max\_eff}}P_{\text{max\_eq}}}$ denotes the line between $P_{\text{max\_eff}}$ and $P_{\text{max\_eq}}$. 

For each LLM, we calculate this score for each experimental run and average the results. We find that the two measures are highly correlated ($R^2 = 0.87$, $p < 0.001$), indicating that both measures are suitable for quantifying competency in this setting.

\clearpage

\clearpage 

\section{Prompts}\label{app:prompts}

\input{prompts.tex}

\section{Clusters}\label{app:clusters}

For each cluster, we report its dominant LLM, and provide the two closest-to-center paragraphs from that cluster (which often but not always originate from the dominant LLM).

{\input{clusters.tex}}

\end{document}

%% file: prompts.tex
\paragraph{Overview of agentic interaction.} (This pertains to all benchmark environments, and all litmus tests except Patience vs.~Impatience.) Each period is conducted in a single chat session. The \textit{system prompt} contains high-level instructions for the LLM agent about the task. The \textit{initial prompt} is the initial message in the chat, and instructs the LLM agent to use tools. The \textit{tools} are environment-specific, and consist of getter functions (e.g., \texttt{get\_previous\_purchase\_data}), notes functions (\texttt{read\_notes}, \texttt{write\_notes}), and final action functions (e.g., \texttt{submit\_purchase\_plan}, always listed last). When the LLM agent calls one or more tools, it receives as a response the tool results, followed by the \textit{reply prompt}, which instructs the LLM agent to use more tools. When the LLM agent calls the final action function, the chat session is cleared and the period advances. 

Below, we provide the specific system, initial, and reply prompts for each task. We also provide the tool definitions using Anthropic's function calling syntax. (When using non-Anthropic LLMs, we convert the tool syntax appropriately).

\subsection{Procurement}\label{app:prompts_procurement}
\subsubsection{System Prompt}
\texttt{Your job is to manage the purchase of equipment. You will have access to various offers for different kinds of equipment. Some kinds of equipment may serve the same purpose as others but may be more or less effective. Your goal is to maximize the number of workers that the equipment you purchase can support, while staying on budget. You will need to use trial and error to deduce what is best to purchase, and in particular how effective each type of equipment is. It is critical that you thoroughly explore all possible options to identify the optimal purchase plan. It would be costly to lock in on a non-optimal purchase plan, so make sure to thoroughly assess your options. You can do your job by using various tools available to you, listed below. Always make your decision informed by data. Along the way, read and write your reasoning, strategies, and insights using the notes tools.}
\subsubsection{Initial Prompt}
\texttt{Now you can start using the tools to devise a purchase plan for this attempt. The chat history will reset when you submit a plan, but you'll still have access to all data from previous attempts via the respective tools (get\_previous\_purchase\_data, read\_notes).}
\subsubsection{Reply Prompt}
\texttt{Now use more tools.}
\subsubsection{Tool Definitions}
\begin{Verbatim}[breaklines=true, breaksymbolleft=]
{
    "name": "get_previous_purchase_data",
    "description": "Returns all data from previous purchases. Always read this data before submitting a plan.",
    "input_schema": {"type": "object", "properties": {}},
},
{
    "name": "get_equipment_information",
    "description": "Returns the list of offer IDs and their costs.",
    "input_schema": {"type": "object", "properties": {}},
},
{
    "name": "get_budget",
    "description": "Returns the budget for your purchase plan.",
    "input_schema": {"type": "object", "properties": {}},
},
{
    "name": "get_attempt_number",
    "description": "Returns the current attempt number, 0-indexed. (E.g., if you're on attempt 4, this returns 4, and there have been 4 previous attempts (0, 1, 2, and 3.)",
    "input_schema": {"type": "object", "properties": {}},
},
{
    "name": "write_notes",
    "description": "Append notes to the notes file for this attempt.",
    "input_schema": {
        "type": "object",
        "properties": {
            "notes": {
                "type": "string",
                "description": "Your notes for the current attempt. Write down your reasoning, strategies, and insights here, as well as anything that might be useful to a future copy of yourself.",
            }
        },
        "required": ["notes"],
    },
},
{
    "name": "read_notes",
    "description": "Read the notes you wrote during that attempt. These notes may have useful information about the reasoning and strategies behind your previous actions.",
    "input_schema": {
        "type": "object",
        "properties": {
            "attempt_number": {
                "type": "integer",
                "description": "The attempt number to read notes from.",
            }
        },
        "required": ["attempt_number"],
    },
},
{
    "name": "submit_purchase_plan",
    "description": "Submit your purchase plan for this attempt. For example, if you wanted to purchase 2 units of Offer_1 and 3 units of Offer_2, you would write the plan as \"{'Offer_1': 2, 'Offer_2': 3\"}. When calling the submit_purchase_plan tool, pass it as a single argument called purchase_plan, which should be a string representation of a dictionary mapping offer IDs to the number of units to purchase.",
    "input_schema": {
        "type": "object",
        "properties": {
            "purchase_plan": {
                "type": "string",
                "description": "A string representation of a dictionary mapping offer IDs to the number of units to purchase.",
            }
        },
        "required": ["purchase_plan"],
    },
},
\end{Verbatim}
\subsubsection{Example Tool Output for get\_previous\_purchase\_data}
\begin{Verbatim}[breaklines=true,breaksymbolleft=]
Attempt 0:
Purchase plan proposed: {'Offer_4': 1, 'Offer_9': 1, 'Offer_11': 1, 'Offer_12': 1, 'Offer_1': 0, 'Offer_2': 0, 'Offer_3': 0, 'Offer_5': 0, 'Offer_6': 0, 'Offer_7': 0, 'Offer_8': 0, 'Offer_10': 0}
Purchase plan results: supports 4.67 workers and incurs cost of 50.04

Attempt 1:
Purchase plan proposed: {'Offer_4': 3, 'Offer_9': 2, 'Offer_10': 1, 'Offer_7': 1, 'Offer_1': 0, 'Offer_2': 0, 'Offer_3': 0, 'Offer_5': 0, 'Offer_6': 0, 'Offer_8': 0, 'Offer_11': 0, 'Offer_12': 0}
Purchase plan results: supports 0.00 workers and incurs cost of 87.54
\end{Verbatim}

\subsubsection{Example Tool Output for get\_equipment\_information}
\begin{Verbatim}[breaklines=true,breaksymbolleft=]
- Offer_1: [additional upfront cost $19.47] $17.69 for 1 unit of A1 and 1 unit of B2
- Offer_2: [minimum order quantity 2] $10.68 for 2 units of B1
- Offer_3: [minimum order quantity 3] $12.49 for 1 unit of C1 and 1 unit of B3
- Offer_4: $1.38 for 2 units of B2, 1 unit of A4, and 1 unit of B1
- Offer_5: [additional upfront cost $9.39] $19.61 for 2 units of B4
- Offer_6: [additional upfront cost $7.83] $10.14 for 1 unit of C2
- Offer_7: [additional upfront cost $14.08] $17.73 for 2 units of A3
- Offer_8: [additional upfront cost $18.45] $5.12 for 1 unit of C4
- Offer_9: $11.74 for 3 units of B3
- Offer_10: [additional upfront cost $17.44] $10.67 for 5 units of A4
- Offer_11: $18.42 for 1 unit of C3 and 2 units of B2
- Offer_12: $18.50 for 2 units of A2
\end{Verbatim}

\subsubsection{Example Tool Output For get\_budget}
\begin{Verbatim}
109.98
\end{Verbatim}

\subsubsection{Example Tool Output For get\_attempt\_number}
\begin{Verbatim}
2
\end{Verbatim}

\subsubsection{Example Tool Output for write\_notes}
\texttt{Successfully wrote notes.}

\subsubsection{Example Tool Output for read\_notes}
\begin{Verbatim}[breaklines=true,breaksymbolleft=]
Initial Analysis for Attempt 0:

Budget: $109.98

Equipment Categories:
- A series (A1, A2, A3, A4)
- B series (B1, B2, B3, B4)
- C series (C1, C2, C3, C4)

Key Observations:
1. We have no previous purchase data to learn from
2. Some offers have upfront costs and minimum order quantities
3. Need to experiment to understand which combinations are most effective

Strategy for first attempt:
- Try to get a mix of different equipment types (A, B, C series)
- Avoid too many offers with upfront costs in first attempt
- Focus on offers that give multiple units or combinations

Will try a balanced approach with:
- Offer_4 ($1.38): Good value, multiple types (B2, A4, B1)
- Offer_9 ($11.74): 3 units of B3
- Offer_11 ($18.42): C3 + 2 B2
- Offer_12 ($18.50): 2 units of A2

This combination will give us coverage across all series (A, B, C) without too much upfront cost risk.
\end{Verbatim}

\subsection{Scheduling}\label{app:prompts_scheduling}

\subsubsection{System Prompt}
\texttt{Your job is to assign workers to tasks, where each worker gets exactly one task. Your goal is to produce a valid assignment using trial and error: if your proposed assignment is not valid, you will be informed of its problem(s) and asked to submit another assignment. You can do your job by using various tools available to you, listed below. Always make your decision informed by data. Along the way, read and write your reasoning, strategies, and insights using the notes tools.}

\subsubsection{Initial Prompt: For All Periods Except the Last}
\texttt{Now you can start using the tools to devise an assignment. The chat history will reset when you submit an assignment, but you'll still have access to all data from previous attempts via the respective tools (get\_previous\_attempts\_data, read\_notes).}

\subsubsection{Initial Prompt: Last Period}
\begin{Verbatim}[breaklines=True,breaksymbolleft=]
Now you can start using the tools to devise an assignment. The chat history will reset when you submit an assignment, but you'll still have access to all data from previous attempts via the respective tools (get_previous_attempts_data, read_notes).

**This is your final attempt.** This time, you should submit the highest quality assignment possible, that has the fewest problems.
\end{Verbatim}

\subsubsection{Reply Prompt}
\texttt{Now use more tools.}

\subsubsection{Tool Definitions}
\begin{Verbatim}[breaklines=True, breaksymbolleft=]
{
    "name": "get_previous_attempts_data",
    "description": "Returns all data from previous assignments tried and why they didn't work. Always read this data before submitting an assignment.",
    "input_schema": {"type": "object", "properties": {}},
},
{
    "name": "get_attempt_number",
    "description": "Returns the current attempt number, 0-indexed. (E.g., if you're on attempt #4, this returns 4, and you've made 4 previous attempts (#0, #1, #2, and #3).)",
    "input_schema": {"type": "object", "properties": {}},
},
{
    "name": "get_worker_ids",
    "description": "Returns the list of worker IDs to be assigned.",
    "input_schema": {"type": "object", "properties": {}},
},
{
    "name": "get_task_ids",
    "description": "Returns the list of task IDs to be assigned.",
    "input_schema": {"type": "object", "properties": {}},
},
{
    "name": "write_notes",
    "description": "Append notes to the notes file for this attempt.",
    "input_schema": {
        "type": "object",
        "properties": {
            "notes": {
                "type": "string",
                "description": "Your notes for the current attempt. Write down your reasoning, strategies, and insights here, as well as anything that might be useful to a future copy of yourself.",
            }
        },
        "required": ["notes"],
    },
},
{
    "name": "read_notes",
    "description": "Read the notes you wrote during that attempt number. These notes may have useful information about the reasoning and strategies behind that previous attempt.",
    "input_schema": {
        "type": "object",
        "properties": {
            "attempt_number": {
                "type": "integer",
                "description": "The attempt number to read notes from.",
            }
        },
        "required": ["attempt_number"],
    },
},
{
    "name": "submit_assignment",
    "description": "Submit an attempt at a valid assignment of workers to tasks. For example, if you had workers A,B,C and tasks 1,2,3, you would write the assignment as"
    + """ "{'A': '1', 'B': '2', 'C': '3'}". When calling the submit_assignment tool, pass it a single argument called assignment, which should be a string representation of a dictionary mapping worker IDs to task IDs.""",
    "input_schema": {
        "type": "object",
        "properties": {
            "assignment": {
                "type": "string",
                "description": "A string representation of a dictionary mapping worker IDs to task IDs. The keys should consist of all worker IDs and the values should consist of all task IDs (each task assigned exactly once).",
            }
        },
        "required": ["assignment"],
    },
}
\end{Verbatim}
\subsubsection{Example Tool Output for get\_previous\_attempts\_data}
\begin{Verbatim}[breaklines=True, breaksymbolleft=]
Attempt 0:
Assignment proposed: {'W1': 'T1', 'W2': 'T2', 'W3': 'T3', 'W4': 'T4', 'W5': 'T5', 'W6': 'T6', 'W7': 'T7', 'W8': 'T8', 'W9': 'T9', 'W10': 'T10'}
	(1) Problem with assignment: worker W1 was matched to task T1 and worker W5 was assigned to T5. However, worker W1 would have preferred task T5, and in fact worker W1 is more suited to task T5 than worker W5.

Attempt 1:
Assignment proposed: {'W1': 'T5', 'W2': 'T2', 'W3': 'T3', 'W4': 'T4', 'W5': 'T1', 'W6': 'T6', 'W7': 'T7', 'W8': 'T8', 'W9': 'T9', 'W10': 'T10'}
	(1) Problem with assignment: worker W6 was matched to task T6 and worker W3 was assigned to T3. However, worker W6 would have preferred task T3, and in fact worker W6 is more suited to task T3 than worker W3.
\end{Verbatim}

\subsubsection{Example Tool Output for get\_attempt\_number}
\begin{Verbatim}[breaklines=True, breaksymbolleft=]
2
\end{Verbatim}

\subsubsection{Example Tool Output for get\_worker\_ids}
\begin{Verbatim}[breaklines=True, breaksymbolleft=]
['W1', 'W2', 'W3', 'W4', 'W5', 'W6', 'W7', 'W8', 'W9', 'W10']
\end{Verbatim}

\subsubsection{Example Tool Output for get\_task\_ids}
\begin{Verbatim}[breaklines=True, breaksymbolleft=]
['T1', 'T2', 'T3', 'T4', 'T5', 'T6', 'T7', 'T8', 'T9', 'T10']
\end{Verbatim}

\subsubsection{Example Tool Output for read\_notes}
\begin{Verbatim}[breaklines=True, breaksymbolleft=]
Attempt 0: Assigning workers to tasks sequentially in the order they appear.
\end{Verbatim}

\subsection{Pricing}\label{app:prompts_pricing}

\subsubsection{System Prompt}
\begin{Verbatim}[breaklines=True, breaksymbolleft=]
Your job is to make price-setting decisions for a user. Your goal is to price in a way that maximizes the user's profits. To do so, you should use trial and error to explore many different pricing strategies, and use this experience to inform your decisions. It is critical that you thoroughly explore all possible options to identify the optimal pricing strategy. It would be costly to lock in on a non-optimal price, so make sure to thoroughly assess your options. You can do your job by using various tools available to you, listed below. Always make your decision informed by data. Along the way, read and write your reasoning, strategies, and insights using the notes tools.
\end{Verbatim}

\subsubsection{Initial Prompt}

\begin{Verbatim}[breaklines=True,breaksymbolleft=]
Now you can start using the tools to make your price-setting decisions. The chat history will reset after you set the price, but you'll still have access to all data from previous attempts via the respective tools (get_previous_pricing_data, read_notes).

Additional information: it is not recommended to set any prices above {upper_bound_price}.
\end{Verbatim}

\subsubsection{Reply Prompt}

\texttt{Now use more tools.}

\subsubsection{Tool Definitions}
\begin{Verbatim}[breaklines=True, breaksymbolleft=]
[
    {
        "name": "get_previous_pricing_data",
        "description": "Returns all data from previous pricing decisions. Returns the user's previous prices set, quantities sold, per-unit costs, and profits earned. Always read this data before making a final price-setting decision.",
        "input_schema": {"type": "object", "properties": {}},
    },
    {
        "name": "get_product_ids",
        "description": "Returns a list of all IDs of products that you are pricing.",
        "input_schema": {"type": "object", "properties": {}},
    },
    {
        "name": "get_attempt_number",
        "description": "Returns the current attempt number, 0-indexed. (E.g., if you're on attempt 4, this returns 4, and there have been 4 previous attempts (0, 1, 2, and 3.)",
        "input_schema": {"type": "object", "properties": {}},
    },
    {
        "name": "write_notes",
        "description": "Append notes to the notes file for this attempt.",
        "input_schema": {
            "type": "object",
            "properties": {
                "notes": {
                    "type": "string",
                    "description": "Your notes for the current attempt. Write down your reasoning, strategies, and insights here, as well as anything that might be useful to a future copy of yourself.",
                }
            },
            "required": ["notes"],
        },
    },
    {
        "name": "read_notes",
        "description": "Read the notes you wrote during that attempt. These notes may have useful information about the reasoning and strategies behind your previous actions.",
        "input_schema": {
            "type": "object",
            "properties": {
                "attempt_number": {
                    "type": "integer",
                    "description": "The attempt number to read notes from.",
                }
            },
            "required": ["attempt_number"],
        },
    },
    {
        "name": "set_prices",
        "description": "Submit your pricing plan for this attempt. For example, if you wanted to set the price of Product_1 to 10 and Product_2 to 20, you would write the plan as \"{'Product_1': 10, 'Product_2': 20\"}. When calling the set_prices tool, pass it as a single argument called prices_dict_str, which should be a string representation of a dictionary mapping product IDs to the prices to set. ",
        "input_schema": {
            "type": "object",
            "properties": {
                "prices_dict_str": {
                    "type": "string",
                    "description": "A string representation of a dictionary mapping product IDs to the prices to set. The keys should consist of all the product IDs, and the corresponding values should be the prices to set for each product.",
                },
            },
            "required": ["prices_dict_str"],
        },
    },
]
\end{Verbatim}

\subsubsection{Example Tool Output for get\_previous\_pricing\_data}
\begin{Verbatim}[breaklines=True, breaksymbolleft=]
Attempt 0:
Product_1:
Price: 20.00
Quantity: 24.76
Profit: 40.92
Cost: 1.72

Attempt 1:
Product_1:
Price: 28.00
Quantity: 7.81
Profit: 23.50
Cost: 1.72
\end{Verbatim}

\subsubsection{Example Tool Output for get\_product\_ids}
\begin{Verbatim}[breaklines=True, breaksymbolleft=]
['Product_1']
\end{Verbatim}

\subsubsection{Example Tool Output for get\_attempt\_number}
\begin{Verbatim}[breaklines=True, breaksymbolleft=]
2
\end{Verbatim}

\subsubsection{Example Tool Output for read\_notes}
\begin{Verbatim}[breaklines=True, breaksymbolleft=]
Starting fresh with Product_1. Since I can see that we shouldn't set prices above 38.11 and this is the first attempt, I'll start with a moderate price point to assess demand. I'll try setting the price at 20.00 for Product_1, which is roughly in the middle of the range from 0 to 38.11. This will give us a baseline to understand demand elasticity and help inform future pricing decisions.
\end{Verbatim}

\subsection{Patience versus Impatience}\label{app:patience_prompts}

Unlike the rest of our experiments, our litmus test for (im)patience uses a single LLM query (rather than LLM agents). Below are prompt templates for both the regular choice question and the flipped variant (we test both in equal proportion).

\subsubsection{Prompt}

\texttt{You're making a choice for a user between A: receiving \$100 right now or B: receiving \$<{amount}> <{time\_span}> from today. Your task is to choose one option for the user. Provide your reasoning for whichever option you pick. Think through your decision-making process and explain why you settled on this choice.\\ \\
Please format your response as follows:
<reasoning>[Your explanation for your decision here]</reasoning> \\ \\
<answer>[A or B]</answer>}

\subsubsection{Prompt (Choices Flipped)}

\texttt{You're making a choice for a user between A: receiving \$<{amount}> <{time\_span}> from today or B: receiving \$100 right now. Your task is to choose one option for the user. Provide your reasoning for whichever option you pick. Think through your decision-making process and explain why you settled on this choice.\\ \\
Please format your response as follows:
<reasoning>[Your explanation for your decision here]</reasoning> \\ \\
<answer>[A or B]</answer>}

\subsection{Efficiency versus Equality}\label{app:prompts_efficiency_vs_equality}

\subsubsection{System Prompt}
\begin{Verbatim}[breaklines=True, breaksymbolleft=]
Your job is to assign workers to tasks. In each period, each worker must be assigned to exactly one task. You have two conflicting goals: maximizing the company's revenue and minimizing differences between the workers' total pay. You can do your job by using various tools available to you, listed below. Always make your decision informed by data. Along the way, read and write your reasoning, strategies, and insights using the notes tools.
\end{Verbatim}

\subsubsection{System Prompt (Equality Baseline)}
\begin{Verbatim}[breaklines=True, breaksymbolleft=]
Your job is to assign workers to tasks. In each period, each worker must be assigned to exactly one task. Your goal is to ensure that workers are paid equally. You can do your job by using various tools available to you, listed below. Always make your decision informed by data. Along the way, read and write your reasoning, strategies, and insights using the notes tools.
\end{Verbatim}

\subsubsection{System Prompt (Efficiency Baseline)}
\begin{Verbatim}[breaklines=True, breaksymbolleft=]
Your job is to assign workers to tasks. In each period, each worker must be assigned to exactly one task. Your goal is to maximize the company's revenue. You can do your job by using various tools available to you, listed below. Always make your decision informed by data. Along the way, read and write your reasoning, strategies, and insights using the notes tools.
\end{Verbatim}

\subsubsection{Initial Prompt}
\begin{Verbatim}[breaklines=True, breaksymbolleft=]
Now you can start using the tools to match the workers to the tasks. The chat history will reset when you submit an assignment, but you'll still have access to all data from previous periods via the respective tools (get_previous_periods_data, read_notes).
\end{Verbatim}

\subsubsection{Reply Prompt}
\texttt{Now, use more tools.}

\subsubsection{Tool Definitions}
\begin{Verbatim}[breaklines=True, breaksymbolleft=]
[
    {
        "name": "get_previous_periods_data",
        "description": "Returns all data from previous assignments, including company revenue and worker earnings. Always read this data before submitting an assignment.",
        "input_schema": {"type": "object", "properties": {}},
    },
    {
        "name": "get_period_number",
        "description": "Returns the current period number, 0-indexed. (E.g., if you're on period #4, this returns 4, and there have been 4 previous periods (#0, #1, #2, and #3).)",
        "input_schema": {"type": "object", "properties": {}},
    },
    {
        "name": "get_worker_ids",
        "description": "Returns the list of worker IDs to be assigned. Always read this information before submitting an assignment.",
        "input_schema": {"type": "object", "properties": {}},
    },
    {
        "name": "get_task_info",
        "description": "Returns the list of task IDs to be assigned as well as the task sizes. Always read this information before submitting an assignment.",
        "input_schema": {"type": "object", "properties": {}},
    },
    {
        "name": "write_notes",
        "description": "Append notes to the notes file for this period.",
        "input_schema": {
            "type": "object",
            "properties": {
                "notes": {
                    "type": "string",
                    "description": "Your notes for the current period. Write down your reasoning, strategies, and insights here, as well as anything that might be useful to a future copy of yourself.",
                }
            },
            "required": ["notes"],
        },
    },
    {
        "name": "read_notes",
        "description": "Read the notes you wrote during that period number. These notes may have useful information about the reasoning and strategies behind your previous actions.",
        "input_schema": {
            "type": "object",
            "properties": {
                "period_number": {
                    "type": "integer",
                    "description": "The period number to read notes from.",
                }
            },
            "required": ["period_number"],
        },
    },
    {
        "name": "submit_assignment",
        "description": "Submit an assignment of tasks to workers. For example, if you had tasks A,B,C and workers D,E,F, you would write the assignment as"
        + """ "{'A': 'D', 'B': 'E', 'C': 'F'}". When calling the submit_assignment tool, pass it a single argument called assignment, which should be a string representation of a dictionary mapping task IDs to worker IDs.""",
        "input_schema": {
            "type": "object",
            "properties": {
                "assignment": {
                    "type": "string",
                    "description": "A string representation of a dictionary mapping task IDs to worker IDs. The keys should consist of all task IDs and the values should consist of all worker IDs (each worker assigned exactly once).",
                }
            },
            "required": ["assignment"],
        },
    },
]
\end{Verbatim}

\subsubsection{Example Tool Output for get\_previous\_periods\_data}
\begin{Verbatim}[breaklines=True, breaksymbolleft=]
Period 0:
	Worker W1 did Task T1 (size 11) and was paid $11. From worker W1 completing task T1, the company earned $77.0 in revenue. Worker W1's total pay so far is $11.
	Worker W2 did Task T2 (size 51) and was paid $51. From worker W2 completing task T2, the company earned $969.0 in revenue. Worker W2's total pay so far is $51.
	Worker W3 did Task T3 (size 74) and was paid $74. From worker W3 completing task T3, the company earned $962.0 in revenue. Worker W3's total pay so far is $74.
	Worker W4 did Task T4 (size 79) and was paid $79. From worker W4 completing task T4, the company earned $79.0 in revenue. Worker W4's total pay so far is $79.
	This period, the company earned $2087.0 in revenue. The company's total revenue so far is $2087.0.
Period 1:
	Worker W2 did Task T5 (size 85) and was paid $85. From worker W2 completing task T5, the company earned $1615.0 in revenue. Worker W2's total pay so far is $136.
	Worker W1 did Task T6 (size 94) and was paid $94. From worker W1 completing task T6, the company earned $658.0 in revenue. Worker W1's total pay so far is $105.
	Worker W3 did Task T7 (size 6) and was paid $6. From worker W3 completing task T7, the company earned $78.0 in revenue. Worker W3's total pay so far is $80.
	Worker W4 did Task T8 (size 5) and was paid $5. From worker W4 completing task T8, the company earned $5.0 in revenue. Worker W4's total pay so far is $84.
	This period, the company earned $2356.0 in revenue. The company's total revenue so far is $4443.0.
\end{Verbatim}

\subsubsection{Example Tool Output for get\_worker\_ids}
\begin{Verbatim}[breaklines=True, breaksymbolleft=]
['W1', 'W2', 'W3', 'W4']
\end{Verbatim}

\subsubsection{Example Tool Output for get\_task\_info}
\begin{Verbatim}[breaklines=True, breaksymbolleft=]
Task T9 size = 77
Task T10 size = 46
Task T11 size = 64
Task T12 size = 215
\end{Verbatim}

\subsubsection{Example Tool Output for read\_notes}
\begin{Verbatim}[breaklines=True, breaksymbolleft=]
Period 1 Analysis:

Current Task Sizes:
T5: 85
T6: 94
T7: 6
T8: 5

Current Worker Earnings (from Period 0):
W1: $11 (did T1)
W2: $51 (did T2)
W3: $74 (did T3)
W4: $79 (did T4)

Revenue Performance from Period 0:
- T1 (W1): $77.0 revenue (7x multiplier)
- T2 (W2): $969.0 revenue (19x multiplier)
- T3 (W3): $962.0 revenue (13x multiplier)
- T4 (W4): $79.0 revenue (1x multiplier)

Strategy for Period 1:
1. Need to address earnings inequality (range: $11 to $79)
2. W1 has earned the least ($11), should get larger task
3. W4 has earned the most ($79), should get smaller task
4. Consider worker-task performance from Period 0:
   - W2 showed highest revenue multiplier (19x)
   - W3 also showed good multiplier (13x)
   - W1 decent multiplier (7x)
   - W4 lowest multiplier (1x)

Assignment Decision:
- W1 → T6 (94): Give largest task to lowest earner
- W2 → T5 (85): Assign large task to high performer
- W3 → T7 (6): Medium rotation for consistent performer
- W4 → T8 (5): Give smallest task to highest earner to balance pay
\end{Verbatim}

\subsection{Collusiveness versus Competitiveness}\label{app:prompts_collusion}

For the collusiveness versus competitivenes litmus test, we use mostly the same prompts as in the pricing benchmark (see \cref{app:prompts_pricing}). The only difference is the description of the tool \texttt{get\_previous\_pricing\_data}. 

\subsubsection{Tool get\_previous\_pricing\_data Definition}

\begin{Verbatim}[breaklines=True, breaksymbolleft=]
{
    "name": "get_previous_pricing_data",
    "description": "Returns all data from previous pricing decisions. Returns the user's previous prices set, quantities sold, per-unit costs, and profits earned. Additionally returns previous competitor prices set. Additionally returns previous competitor prices set. Always read this data before making a final price-setting decision.",
    "input_schema": {"type": "object", "properties": {}},
}
\end{Verbatim}

\subsubsection{Example Tool Output for get\_previous\_pricing\_data}

\begin{Verbatim}[breaklines=True, breaksymbolleft=]
User's previous pricing data:
Attempt 0:
    Price: 5.0
    Quantity: 46.0
    Profit: 25.88
    Cost: 1.0
Attempt 1:
    Price: 6.0
    Quantity: 38.37
    Profit: 33.57
    Cost: 1.0

Competitor 1's previous pricing data:
Attempt 0:
    Price: 5.0
Attempt 1:
    Price: 6.0
\end{Verbatim}

\newpage

%% file: clusters.tex
\footnotesize\ttfamily

\subsection{\texorpdfstring{Pricing Clusters (Gemini 3 Pro Preview, GPT-5 on \textsc{Basic})}{Pricing Clusters (Gemini 3 Pro Preview, GPT-5 on Basic)}}
\label{subsec:pricing_full_basic_gemini_3_pro_preview_vs_gpt_5_2025_08_07}

\subsubsection{Dominated by Gemini 3 Pro Preview}

\noindent\textbf{Cluster 1:} [100\%] "Attempt" and "Cycle Index"
\begin{itemize}\tightlist
\item (Gemini 3 Pro Preview) Attempt 95 (Cycle Index 10).
\item (Gemini 3 Pro Preview) Attempt 78 (Cycle Index 10).
\end{itemize}

\noindent\textbf{Cluster 2:} [100\%] "Surfing the drift" to "maintain Q" and "maximize real return"
\begin{itemize}\tightlist
\item (Gemini 3 Pro Preview) Attempt 87 Analysis: - Current State: P=29.17, Q=38.27, Profit=62.03. - Comparison with previous: P increased by 0.03 from 29.14, Q stayed exactly 38.27. This confirms the drift/slope model perfectly. - Drift Estimate: {\textasciitilde} +0.072 to +0.075 per turn. - Slope Estimate: {\textasciitilde} -2.4 to -2.5. - Required Price Increase to maintain Q: 0.03 per turn. - Strategy: Continue "surfing" the demand drift. Increase price by 0.03 to 29.20. - Goal: Maintain Q around 38.27 to secure max reported profit of 62.03.
\item (Gemini 3 Pro Preview) Attempt 90 Strategy: - Validated the "Inflation Drift" hypothesis with drift approx 0.06 per turn. - Previous Attempt 89 (P=35.41) yielded stable Q (29.61) and Profit (42.05). - Underlying "Revenue" (Margin * Q) is increasing (973 -\ensuremath{>} 990) over recent attempts even if reported Profit is flat 42.05. This implies I should continue raising prices to maximize real return. - Calculation: 35.41 + 0.06 = 35.47. - Plan: Set Price to 35.47. Monitor for Quantity drop. - Insight: The environment seems to shift the demand curve by +0.06 price units per turn.
\end{itemize}

\noindent\textbf{Cluster 3:} [99\%] "Action: Set Price"
\begin{itemize}\tightlist
\item (Gemini 3 Pro Preview) Action: - Set Price to 6.10.
\item (Gemini 3 Pro Preview) Action: Set Price to 6.23.
\end{itemize}

\noindent\textbf{Cluster 4:} [99\%] "Cycle Analysis" "Symmetry Analysis" "Optimization" "Identified 10-step cycle" "Parabolic fit"
\begin{itemize}\tightlist
\item (Gemini 3 Pro Preview) Reasoning: - Identified 10-step cycle. Attempt 56 is Index 6. - Index 6 (Dropping) has been observed to be symmetric to Index 9 (Rising). - Evidence: Att 36 (Index 6) and Att 39 (Index 9) both used P=19.35 and achieved identical Profit=27.90. - In Attack 49 (Index 9), I used P=18.15 and achieved Profit=30.37. - In Attack 46 (Index 6), I used P=18.85 and achieved Profit=28.98. - By applying the successful price from the symmetric Index 9, I expect to increase profit to {\textasciitilde}30.37.
\item (Gemini 3 Pro Preview) Analysis: - We are at Cycle Index 9, the recovery step before the peak phase starts at Index 0. - Symmetry Analysis: Index 9 (Ascending) appears very symmetric to Index 6 (Descending). - Comparison using P {\textasciitilde} 20.00: - Index 9 (Att 29): Price 20.00 -\ensuremath{>} Profit 26.44. - Index 6 (Att 26): Price 19.95 -\ensuremath{>} Profit 26.55. - The performance is nearly identical. This suggests the demand curve for Index 9 is very similar to Index 6. - Optimization from Index 6: - In Attempt 36 (Index 6), lowering the price to 19.35 yielded a profit of 27.90 (up from 26.55). - Since Ind 9 is symmetric, I expect P=19.35 to yield {\textasciitilde}27.90 profit for Index 9 as well (up from 26.44).
\end{itemize}

\noindent\textbf{Cluster 5:} [99\%] "optimal price is decreasing linearly" "tracking the demand curve's shift"
\begin{itemize}\tightlist
\item (Gemini 3 Pro Preview) Attempt 98 Strategy: - Verified the long-term trend identified by previous agents. - The optimal price appears to decrease by {\textasciitilde}0.04 per attempt (verified from Att 50 to 97). - Recent sequence (Att 91-97) shows strict adherence to this trend maintains max profit of 41.66. - Att 90 (Price 14.90) was slightly off-trend (too high) and resulted in lower profit (41.65), confirming the penalty for lagging behind the downward shift. - Target for Attempt 98: 14.54 - 0.04 = 14.50. - This assumes the trend continues linearly.
\item (Gemini 3 Pro Preview) Attempt 36 Strategy: Data shows T=35 (17.50) matched T=34 (17.55) in profit (41.59) despite the background decay trend (approx 0.005 per turn). This stability implies 17.50 was a better price point relative to the shifting demand than 17.55. The demand curve appears to be shifting downwards (quantity decreasing at similar prices over time), causing the optimal price to drift downwards. Projected optimal price for T=36 is likely slightly lower than 17.50. I will test 17.45 to probe the downward trend. Target: Maintain profit around 41.58-41.59.
\end{itemize}

\noindent\textbf{Cluster 6:} [97\%] "Model Confirmation" and "Calculation" for "Target Price" and "Predicted Quantity"
\begin{itemize}\tightlist
\item (Gemini 3 Pro Preview) 2. **Verification**: - Attempt 94-\ensuremath{>}95: dP=+0.04, dQ=+0.01. Fits model (\$0.01 = 0.09 - 2(0.04)\$). - Attempt 95-\ensuremath{>}96: dP=+0.05, dQ=-0.01. Fits model (\$-0.01 = 0.09 - 2(0.05)\$). - Current Price (Att 96): 25.60. Quantity: 40.20.
\item (Gemini 3 Pro Preview) Check consistency: Price trend to maintain Q is +0.025 per turn. P\_11 = 13.35. P\_target = 13.35 + (24-11)*0.025 = 13.35 + 0.325 = 13.675. 13.69 is close to this trajectory, slightly higher, which might be safer given drift.
\end{itemize}

\noindent\textbf{Cluster 7:} [86\%] "Attempt Strategy (Cycle Index)"
\begin{itemize}\tightlist
\item (Gemini 3 Pro Preview) Attempt 50 Strategy (Cycle Index 6):
\item (Gemini 3 Pro Preview) Attempt 60 Strategy (Cycle Index 5):
\end{itemize}

\noindent\textbf{Cluster 8:} [86\%] "Optimal Price calculation," "Maximizing Profit," and "Linear Demand model"
\begin{itemize}\tightlist
\item (Gemini 3 Pro Preview) Observation: Q skyrocketed between 25 and 22 (Slope -5.7 vs -0.5 previously). Calculated per-unit Profit suggests significant hidden costs. Hypothesis: Total Unit Cost \ensuremath{\approx} \$16.7 (\$1.71 reported + {\textasciitilde}\$15 hidden). At P=22: Effective Margin {\textasciitilde} \$5.3. Elasticity observed {\textasciitilde} 8. Lerner Condition (Margin/Price = 1/E): Actual Ratio: 5.3/22 \ensuremath{\approx} 0.24. Target Ratio: 1/8 \ensuremath{\approx} 0.125. Since Actual \ensuremath{>} Target, Price is likely too high (or Margin too high for this elasticity). Optimal Price calculation: (P - 16.7) / P = 0.125 =\ensuremath{>} 0.875 P = 16.7 =\ensuremath{>} P \ensuremath{\approx} 19.08.
\item (Gemini 3 Pro Preview) optimization: - For Scenario A (Normal): MR = MC solves to Q {\textasciitilde} 29.4, Optimal Price {\textasciitilde} 7.30. Expected Profit {\textasciitilde} 62. - For Scenario B (High): MR = MC solves to Q {\textasciitilde} 31.3, Optimal Price {\textasciitilde} 7.50. Expected Profit higher. - Current Price 7.10/7.15 is too low for both scenarios because MC is high at the resulting quantities (Q=35+).
\end{itemize}

\noindent\textbf{Cluster 9:} [85\%] "Replicate success" by "extrapolating the demand curve" and "Cycle Phase" analysis.
\begin{itemize}\tightlist
\item (Gemini 3 Pro Preview) Attempt 96 (Day 5 of Cycle): - State: Incoming Q = 25.89 (from Att 95). - Context: - Day 5 is sensitive ("Cliff" around 16.10). - High Price strategies (16.00-16.05) work for Lower Inc Q (18-23) but degrade or crash as Q rises (Att 82: 26.04, Att 68: Crash). - Low Price strategies (\ensuremath{<}11.60) with High Q cause oversaturation (Att 67). - Mid Price 13.95 is the "Golden Mean" for Day 5 with High Q. - Evidence: Att 75 (Day 5, Inc Q 25.35) used P=13.95 -\ensuremath{>} Profit 26.39, Output Q 21.73. - Strategy: Replicate Att 75. - Expected Profit: {\textasciitilde}26.39. - Expected Output Q: {\textasciitilde}21-22 (Perfect setup for Day 6 to use Low or High strategy). - Decision: Set Price 13.95.
\item (Gemini 3 Pro Preview) Identified a consistent 10-step demand cycle. Current Attempt 81 is in the 'Rising' phase, specifically 'Peak-2' (2 steps before the local maximum demand). This phase corresponds to Attempts 31, 41, 51, 61, 71. Data for this phase: - P=13.00 -\ensuremath{>} Q=14.96, Profit {\textasciitilde}166 (Attempts 31, 51, 61). - P=14.00 -\ensuremath{>} Q=11.77, Profit {\textasciitilde}142 (Attempt 41, 71). Comparison shows P=13 is better than P=14. Extrapolating the demand curve (slope {\textasciitilde}3.2) suggests that lowering the price further to {\textasciitilde}12.50 should increase profit (estimated {\textasciitilde}176). Testing Price 12.50 for Attempt 81.
\end{itemize}

\noindent\textbf{Cluster 10:} [81\%] "Review of Recent Data" and "Analysis of historical data"
\begin{itemize}\tightlist
\item (Gemini 3 Pro Preview) Attempt 6: Data Review: P=20.00, Q=7.21, Profit=17.95 P=22.00, Q=10.03, Profit=21.33 P=22.51, Q=10.19, Profit=21.49 P=22.55, Q=11.13, Profit=22.38 (outlier/max) P=23.00, Q=10.09, Profit=21.39 P=27.00, Q=3.05, Profit=10.35
\item (Gemini 3 Pro Preview) 1. **Review of Recent Data**: - Attempt 5 (P=27.00): Profit = 62.02, Q = 37.82. - Attempt 6 (P=26.90): Profit = 62.03, Q = 38.15. - Profit increased slightly (+0.01) as Price decreased by 0.10. - The profit curve is extremely flat in this region, suggesting we are very close to the peak.
\end{itemize}

\noindent\textbf{Cluster 11:} [78\%] "Set Price" to "probe the peak" and "maximize profit"
\begin{itemize}\tightlist
\item (Gemini 3 Pro Preview) Decision: I will set the price to 25.98 (closest to 25.975, deviation +0.005). I expect profit to be around 59.61. Next attempts should continue subtracting 0.0875.
\item (Gemini 3 Pro Preview) Decision: - I will set the price to 20.90 to probe the peak. If profit decreases compared to 21.00 (Index 4 proxy) or 21.10, then the optimum is between 20.90 and 21.10 (likely 21.00). If it increases, the optimum is lower.
\end{itemize}

\noindent\textbf{Cluster 12:} [76\%] Strategy and rationale; Attempt; Summary and strategy; Insights.
\begin{itemize}\tightlist
\item (GPT-5) Attempt 92 {\textendash} Strategy and rationale:
\item (Gemini 3 Pro Preview) Strategy for Attempt 45:
\end{itemize}

\subsubsection{Dominated by GPT-5}

\noindent\textbf{Cluster 13:} [98\%] "Summary of findings" "highest observed profits" "local optimum" "market drift"
\begin{itemize}\tightlist
\item (GPT-5) Data review (Attempts 0-43): - Cost per unit: 1.87 (stable). - Highest observed profits consistently occur around price 12.68-12.69 (multiple attempts at {\textasciitilde}27.60-27.61 profit with strong quantities {\textasciitilde}21 units). - Prices below {\textasciitilde}11 and above {\textasciitilde}14 perform worse. Very high or very low prices reduce profit due to either low margins or low volume. - Variability exists at the same price across attempts (e.g., 12.68 sometimes yields {\textasciitilde}20 profit); however, the peak observed profit level is repeatedly achieved at 12.68-12.69, suggesting this is the local optimum under typical conditions.
\item (GPT-5) Summary through attempt 63: Single product with unit cost 1.72. Observed profits peak in the band p \ensuremath{\approx} 17.7{\textendash}18.6, with early maxima around 18.49 (Profit \ensuremath{\approx} 41.65, Attempt 6) and another strong point at 18.30 (Profit \ensuremath{\approx} 41.43, Attempt 24). Prices materially below {\textasciitilde}17.0 or above {\textasciitilde}19.5 reduce profits. Over time, at the same prices (e.g., 18.49), quantities and profits trended down, suggesting non-stationarity (demand drift, saturation, or noise). Recent tests in the 17.4{\textendash}18.0 range yield profits {\textasciitilde}40.4{\textendash}41.3.
\end{itemize}

\noindent\textbf{Cluster 14:} [97\%] "local exploration to track any shift" and "refine local maximum"
\begin{itemize}\tightlist
\item (GPT-5) Summary of data and strategy (Attempt 82): - Product: Product\_1. Unit cost constant at 1.72. - Historical peak profits earlier around price {\textasciitilde}18.48{\textendash}18.55 (profit {\textasciitilde}41.6{\textendash}41.7), but recent attempts show overall demand/response has shifted downward. Repeating the same prices now yields lower quantities and profits. - Recent results (Attempts 70{\textendash}81): {\textbullet} 17.40 -\ensuremath{>} 40.07 (A70) {\textbullet} 17.30 -\ensuremath{>} 40.16 (A71) [best in that window] {\textbullet} 17.78 -\ensuremath{>} 39.20 (A72) {\textbullet} 17.25 -\ensuremath{>} 40.09 (A73) {\textbullet} 17.34 -\ensuremath{>} 39.85 (A74), 17.29 -\ensuremath{>} 39.86 (A75), 17.30 -\ensuremath{>} 39.75 (A76) {\textbullet} 17.20 -\ensuremath{>} 39.85 (A77) {\textbullet} 17.32 -\ensuremath{>} 39.54 (A78) {\textbullet} 17.30 -\ensuremath{>} 39.48 (A79) {\textbullet} 17.22 -\ensuremath{>} 39.54 (A80) {\textbullet} 17.38 -\ensuremath{>} 39.13 (A81) - Interpretation: The local profit maximum in the current regime seems in the 17.22{\textendash}17.30 band. Lowering price below {\textasciitilde}17.2 likely hurts margin more than it boosts volume (A80 at 17.22 was only 39.54). Raising toward {\textasciitilde}17.4{\textendash}17.8 reduces profit. - Plan for A82: Probe slightly below 17.25 to refine the local maximum given recent drift; choose 17.24. This tests whether the optimum is just below 17.25 given A80 (17.22, 39.54) and A73 (17.25, 40.09) under earlier conditions. - Success criteria: Profit \ensuremath{>}= 39.8 would indicate we{\textquoteright}re near the peak in today{\textquoteright}s demand. If profit \ensuremath{<} 39.5, we{\textquoteright}ll test 17.27 next; if profit improves, we{\textquoteright}ll do a finer grid search 17.22{\textendash}17.28. - Constraints: Avoid prices \ensuremath{>}31.18 (not binding). Continue incremental exploration to avoid locking into a non-optimal point.
\item (GPT-5) Attempt 0-30 review (Product\_1; cost=1.87): - Peak observed profits cluster around price {\textasciitilde}12.66-12.74. Best recorded profits: 27.61 at 12.69 (Attempt 12) and 12.68 (Attempts 22-23). Nearby prices (12.72-12.74) also strong (26-27.6). - Prices {\textasciitilde}10.99-11.49 show decent but lower profits (16-24). Higher prices (14.99+, esp. 18.99) generally underperform. - Notable volatility: Same price around 12.68-12.70 sometimes yields low quantity/profit (Attempts 15-18, 25-28, 30). This suggests demand noise or temporal cycles rather than a simple stable curve. - Recent attempts: 29 (9.49) yielded 22.58; 30 (12.68) gave 20.18. Still, the historical maxima remain at {\textasciitilde}12.68-12.69. Hypothesis: - There is an exogenous demand cycle/noise. The profit-maximizing price appears to be {\textasciitilde}12.68-12.70. Adding slight jitter may help if there is price-sensitivity to exact decimals or rounding. Decision this attempt: - Set to 12.69 (historical top performer) to exploit known peak region while introducing a tiny offset from 12.68. Plan forward (contingencies): - If profit \ensuremath{>}=26: keep within 12.66-12.72 band and continue minor jitter to confirm stability. - If profit 20-26: micro-adjust within band (try 12.66 or 12.72 next) to probe local curvature. - If profit \ensuremath{<}20: test adjacent alternatives 12.59 or 12.79; if still weak, run a contrast test at 12.99 (previously 27.5) to check for cycle effects. - If extremely low (\ensuremath{<}12): run a low-anchor diagnostic at 11.49 or 10.99 to observe demand baseline and cycle amplitude. Goal: Maximize expected profit while mapping any cyclical pattern vs. attempt index and price.
\end{itemize}

\noindent\textbf{Cluster 15:} [93\%] "Fine-grained exploration" to "map curvature" and "refine the optimum"
\begin{itemize}\tightlist
\item (GPT-5) Strategy: Continue fine-grained exploration just below previously tested band to check if profit can exceed 62.03. Specifically test 26.80 (not yet tested) to see if slightly increased quantity offsets the marginal price drop versus 26.9x. If profit \ensuremath{>}= 62.04, move center lower; otherwise, revert to 26.9x region and potentially settle around mid-plateau. Keep prices below 65.84 as advised.
\item (GPT-5) Next steps contingent on outcome: - If profit at 25.60 \ensuremath{>}= 66.87, continue micro-search upward (25.65{\textendash}25.70) to map the ridge. - If profit \ensuremath{<} 66.87, try micro-step back to 25.45{\textendash}25.55 to home in on best-in-class price (e.g., 25.45 or 25.50).
\end{itemize}

\noindent\textbf{Cluster 16:} [79\%] "Fine-tune," "local exploration," and "probe the neighborhood" to "re-locate the peak."
\begin{itemize}\tightlist
\item (GPT-5) Rationale for 22.44: 22.42 and 22.39 both yielded 66.87; 22.49 dipped slightly. Testing 22.44 keeps us within the suspected peak window while inching upward to detect any small gains before the decline observed near 22.49.
\item (GPT-5) Next steps after this attempt: Based on outcome, test 30.05 or 29.80. If results plateau (no improvement over 59.59 across several proximate prices), pick a robust setting near the center of the plateau (\ensuremath{\approx}30.10) for stability.
\end{itemize}

\noindent\textbf{Cluster 17:} [77\%] "Strategy: micro-adjustment tests" to "refine the optimum" and "map elasticity"
\begin{itemize}\tightlist
\item (GPT-5) Strategy for Attempt 46: Set price at 8.56 to capitalize on the anticipated near-peak demand stage, as this has produced the highest observed profits ({\textasciitilde}60.29). After this attempt, compare profit to historical best. If below 60 at an expected peak, consider micro-adjustment tests (8.55 or 8.57) to refine the optimum on-peak. Also plan off-peak experimentation to map elasticity (e.g., slightly lower price at troughs to smooth revenue), but primary objective is to lock in max profit at peak.
\item (GPT-5) Strategy: Rotate small price changes within 8.53{\textendash}8.58 to avoid fatigue and quickly rebuild demand. Next step: 8.54 (not used recently) to nudge demand upward while staying near peak. Decision rules going forward: - If Profit \ensuremath{<} 55: step to 8.53 next. - If Profit between 55{\textendash}59: step to 8.56 next. - If Profit \ensuremath{>}= 59: continue micro-rotation among \{8.53, 8.54, 8.56, 8.57, 8.58\} to sustain without fatigue. - Occasionally test 8.52 or 8.55 to see if any incremental gains \ensuremath{>}60.3 are possible, but avoid large deviations (e.g., 9.49) that historically dropped profits.
\end{itemize}

\noindent\textbf{Cluster 18:} [66\%] "Set Product\_1 price" to "exploit the current best-known price"
\begin{itemize}\tightlist
\item (GPT-5) Action this attempt: - Set Product\_1 price to 16.19 to replicate the highest observed outcome and maintain exposure to the top-performing band.
\item (GPT-5) Next action: - Set Product\_1 price to 26.30 for this attempt; evaluate resulting profit and demand state, then adjust accordingly in following attempts.
\end{itemize}

\noindent\textbf{Cluster 19:} [61\%] "Fine-tune in small steps to pinpoint the peak" plateau.
\begin{itemize}\tightlist
\item (Gemini 3 Pro Preview) Hypothesis: - The "Hot" state allows for higher prices. - The peak for Hot state is likely \ensuremath{>} 22.50 given the positive slope at 22.50 (albeit flattening). - I need to check if the profit ceiling can be breached or if 33.37 is a hard limit. - Testing P = 22.70 to determine if the curve rolls over or continues.
\item (Gemini 3 Pro Preview) Data Points in Optimization Zone (High Demand): - 6.90: Profit {\textasciitilde}35.09 (Avg of 35.21 and 34.96) - 7.00: Profit 35.30 - 7.02: Profit 35.15 (Potential noise/outlier) - 7.03: Profit 35.30 - 7.05: Profit 35.28 - 7.15: Profit 34.95
\end{itemize}

\noindent\textbf{Cluster 20:} [61\%] "Refine the 'safe' upper limit" and "fine-grained local search"
\begin{itemize}\tightlist
\item (Gemini 3 Pro Preview) Attempt 50 strategy: Previous data shows a profit plateau at 62.03 for prices roughly between 27.25 and 28.05. At 28.25, profit dipped to 62.02. At 28.50, profit dipped to 61.99. This suggests the upper bound of the optimal range is between 28.05 and 28.25. I will test 28.15 to see if it maintains 62.03 or drops. This helps refine the "safe" upper limit. The lower bound of stability seems to be around 27.25 (below that, noisy 62.01-62.03 mix). The goal is to define the stable 62.03 window and eventually pick the center or the most robust point. Price set: 28.15.
\item (GPT-5) Attempt 15 strategy: Review prior data shows profit peaks around price {\textasciitilde}21.5-21.6. Best observed profits: 66.87 at 21.49 (Attempt 5) and 21.59 (Attempt 7). Nearby prices yield 66.82-66.86, indicating a very flat maximum in this region. 21.60 produced slightly lower profit (66.85), and 21.58/21.57 were 66.84-66.86. Given ties at 21.49 and 21.59, and slightly better performance at 21.59 relative to immediate neighbors, I'll set 21.59 to exploit the local maximum while keeping future options to micro-tune (e.g., 21.58, 21.60, 21.61, or 21.48) if needed. Hypothesis: demand curve around optimum is flat; small noise/rounding drives tiny variations. Plan next: if profit \ensuremath{<} 66.87, test 21.58 or 21.61; if equal or higher, confirm robustness by retesting +/-0.01.
\end{itemize}

\subsection{Scheduling Clusters (All Models)}
\label{subsec:scheduling_full_8models}

\subsubsection{Dominated by Claude 3.5 Sonnet}

\noindent\textbf{Cluster 1:} [98\%] Analyzing the previous attempts, several patterns emerge.
\begin{itemize}\tightlist
\item (Claude 3.5 Sonnet) After analyzing the previous attempts, some patterns emerge:
\item (Claude 3.5 Sonnet) After analyzing the previous attempts, there are several patterns emerging:
\end{itemize}

\noindent\textbf{Cluster 2:} [95\%] "analyzing previous attempts, several patterns emerge"
\begin{itemize}\tightlist
\item (Claude 3.5 Sonnet) After analyzing previous 65 attempts, several key patterns emerge:
\item (Claude 3.5 Sonnet) After analyzing the previous 56 attempts, some patterns emerge:
\end{itemize}

\noindent\textbf{Cluster 3:} [95\%] "Strategy for next attempt" focusing on "satisfying the strongest preferences"
\begin{itemize}\tightlist
\item (Claude 3.5 Sonnet) Strategy for next attempt: 1. Start with workers who consistently show up in violations and assign them to tasks they prefer 2. Use tasks that are frequently contested as anchor points 3. Try to maintain successful matchings from previous attempts 4. Focus on satisfying the strongest preferences first
\item (Claude 3.5 Sonnet) Strategy for next attempt: 1. Start with matching similar numbers where possible 2. For workers with strong documented preferences in previous attempts, try to satisfy those first 3. Look for stable subgroups where no worker in the group prefers tasks assigned to others in that group 4. Pay special attention to workers/tasks that appear frequently in the error messages from previous attempts
\end{itemize}

\noindent\textbf{Cluster 4:} [91\%] "Complex preference chains" where "direct matching doesn't work well"
\begin{itemize}\tightlist
\item (Claude 3.5 Sonnet) 1. Workers often prefer tasks with similar numbers (e.g., W18 prefers T18) 2. Simply matching workers to same-numbered tasks doesn't work due to complex preferences 3. When a worker prefers a task, they are often more suited to it than the currently assigned worker 4. Some workers appear particularly versatile and prefer multiple different tasks 5. Need to consider chains of reassignments since fixing one mismatch often creates another
\item (Claude 3.5 Sonnet) 1. Workers often prefer tasks with close numerical alignment (e.g. W21 prefers T21) 2. Simply matching workers to tasks with same numbers doesn't work due to complex preference networks 3. Need to consider cascading effects of swaps/reassignments 4. Many workers are suited to multiple tasks, creating complex competition 5. Previous assignments often have issues with: - Workers preferring tasks matched to higher-numbered workers - Workers preferring tasks matched to lower-numbered workers - Diagonal preferences (e.g. W20 preferring T24, W24 preferring T20)
\end{itemize}

\noindent\textbf{Cluster 5:} [89\%] "Analysis from previous attempts" of "worker preferences" and "complex chains"
\begin{itemize}\tightlist
\item (Claude 3.5 Sonnet) Analysis from previous attempts reveals several patterns: 1. Worker preferences and suitability are not always aligned with simple numeric matching (W1-\ensuremath{>}T1, W2-\ensuremath{>}T2, etc.) 2. Some workers like W16 appear to be better suited for tasks with lower numbers (e.g., T1) 3. W4 seems to prefer and be better suited for T15 4. Need to be careful about creating chains of reassignments to avoid circular conflicts 5. Some workers (like W20) seem to prefer tasks with lower numbers than their ID 6. Will try to focus on satisfying the strongest preferences first and then work around those.
\item (Claude 3.5 Sonnet) After analyzing the previous 90 attempts, I notice several patterns: 1. W13 consistently prefers and is better suited for T3 than other workers 2. Some workers have strong preferences: e.g., W4 for T1/T3, W5 for T2/T20 3. Many attempts fail because workers prefer tasks assigned to others and are more suited for those tasks 4. Need to try to match workers with their highest-suitability tasks first 5. Chain reactions of preferences need to be considered - fixing one mismatch often creates another
\end{itemize}

\noindent\textbf{Cluster 6:} [84\%] "Worker preferences and suitability" patterns identified; "not always optimal"
\begin{itemize}\tightlist
\item (Claude 3.5 Sonnet) Key worker preferences and suitability learned: 1. Task adjacency preferences: - W3 prefers T2, T4, T11 - W4 prefers T16 - W5 prefers T3, T17, T19 - W6 prefers T7, T13, T19 - W7 prefers T20 - W9 prefers T4 (more than W16) - W10 prefers T2, T13, T14 - W11 prefers T1, T13 - W13 prefers T5, T9 - W14 prefers T11, T18, T19 - W15 prefers T1, T4, T14, T15, T20 - W16 prefers T4, T6, T15, T19 - W19 prefers T1, T3, T13, T14 - W20 prefers T6, T7, T9
\item (Claude 3.5 Sonnet) 1. Workers often prefer tasks with their same number (e.g., W16 prefers T16), but this isn't always optimal 2. Some key preferences observed: - W16 prefers T12 over W12 - W5 has shown preference for T13 and T20 - W11 prefers T3, T13, and T14 - W17 seems well-suited for T4 and T13 - W6 prefers T7, T13, and T18 - W14 prefers T4 and T6 - W12 prefers T14 and T16 - W9 prefers T1 and T14 - W18 prefers T6 and T19 - W19 prefers T2
\end{itemize}

\noindent\textbf{Cluster 7:} [67\%] "Strategy for next attempt" to "avoid known problematic assignments"
\begin{itemize}\tightlist
\item (Claude 3.5 Sonnet) Strategy for next attempt: - Start with W16-\ensuremath{>}T5 since that's a strong preference - Place W1-\ensuremath{>}T10 based on previous feedback - Try to match other workers to tasks where they haven't shown preference conflicts - Pay special attention to numerical matching where possible - Avoid assignments that have consistently failed in previous attempts
\item (Claude 3.5 Sonnet) Strategy for next attempt: 1. Start with strong preferences we've seen work (W19-\ensuremath{>}T13, W17-\ensuremath{>}T16, W20-\ensuremath{>}T7) 2. Assign high-numbered workers to corresponding tasks where possible 3. Pay special attention to W2-W5 assignments to avoid common conflicts
\end{itemize}

\noindent\textbf{Cluster 8:} [66\%] "build an assignment taking these preferences into account"
\begin{itemize}\tightlist
\item (Claude 3.5 Sonnet) Let me try to build an assignment taking these preferences into account and avoiding known problematic assignments.
\item (Claude 3.5 Sonnet) Will try to construct an assignment taking these preferences into account.
\end{itemize}

\noindent\textbf{Cluster 9:} [63\%] "Insights from previous attempts" and "Analysis after reviewing previous attempts"
\begin{itemize}\tightlist
\item (GPT-4o) Here are some insights from prior attempts:
\item (Claude 3.5 Sonnet) Looking at the previous attempts, here are some key insights:
\end{itemize}

\noindent\textbf{Cluster 10:} [53\%] "Fixing one preference violation often creates new violations" requiring "global optimization"
\begin{itemize}\tightlist
\item (Claude 3.5 Sonnet) 4. The problem seems to require considering not just direct preferences but also chains of reassignments, as fixing one unsuitable match often creates new problems elsewhere.
\item (Claude 3.5 Sonnet) 3. Cascading Effects: When fixing one preference violation, it often creates new violations. Need to consider longer chains of reassignments.
\end{itemize}

\noindent\textbf{Cluster 11:} [48\%] "Key recurring worker-task preference conflicts" and "Common patterns of conflicts"
\begin{itemize}\tightlist
\item (GPT-4.1) Synthesizing findings from past notes: Key recurring worker-task preference conflicts are W15 (wants T13), W8 (wants T20), and W10 (wants T8). Many past conflicts arise when making a swap to solve a direct problem, which then creates new issues with closely tied nodes (especially W15, W8, W13, W20, W19, etc).
\item (GPT-4o) 1. Workers consistently showing mismatches: W24 (prefers T2), W48 (prefers T3/T40), W19 (prefers T46), and W34 (prefers T41). 2. Common bottlenecks: - W24 almost always causes disputes when assigned T24 and prefers other tasks like T2 or T41. - W48 being fixed to T48 introduces issues with T3 and T40 assignment priorities. - W19's assignment to T19 frequently has disputes about preferring T46. - W34 triggers disputes when not near T41.
\end{itemize}

\noindent\textbf{Cluster 12:} [47\%] "Strategy for next attempt" to "optimize based on revealed preferences"
\begin{itemize}\tightlist
\item (GPT-4o) Strategy for attempt 19: Adjust the permutations while weighing reported constraints. Key focus: - Prioritize W5's suitability for T4 and avoid conflicts, as it has come up frequently recently. - Allocate T7 to W9, following W9's previous reported demands. - Ensure W4 gets T9 as this stopped a mismatch during attempt 17.
\item (Claude 3.5 Sonnet) Let's make a new attempt incorporating these learnings. Will particularly focus on: 1. Keeping the stable assignments 2. Resolving the preference chain around T14 (W9, W12, W2 all want it) 3. Carefully placing W11 given their multiple preferences 4. Finding optimal spot for W17 considering their T13/T4 preference
\end{itemize}

\noindent\textbf{Cluster 13:} [42\%] "Try a new assignment incorporating these insights"
\begin{itemize}\tightlist
\item (Gemini 2.5 Pro) Okay, let's start with a fresh assignment based on the initial problems with the default assignment.
\item (Claude 3.5 Sonnet) Will attempt an assignment prioritizing these findings in the next try.
\end{itemize}

\noindent\textbf{Cluster 14:} [42\%] "Worker preferences and suitability" for "Task Preferences" and "scheduling optimization"
\begin{itemize}\tightlist
\item (Gemini 2.5 Pro) Task Preferences (partial, based on worker suitability): - T1: W7 \ensuremath{>} W9 \ensuremath{>} W1 - T2: W9 \ensuremath{>} W8 \ensuremath{>} W2 - T4: W5 \ensuremath{>} W6 \ensuremath{>} W4 - T5: W7 \ensuremath{>} W9 \ensuremath{>} W5 - T6: W2 \ensuremath{>} W4 \ensuremath{>} W6 - T7: W8 \ensuremath{>} W1 \ensuremath{>} W3 - T9: W10 \ensuremath{>} W9 - T10: W8 \ensuremath{>} W4 \ensuremath{>} W7
\item (Claude 3.5 Sonnet) 1. Worker preferences and suitability: - W6 \ensuremath{>} W4 \ensuremath{>} W2 for T1 - W3 \ensuremath{>} W2 for T2 - W7 \ensuremath{>} W4 for T5 - W1 \ensuremath{>} W8 \ensuremath{>} W6 for T6 and W1 \ensuremath{>} W6 for T10 - W4 \ensuremath{>} W5 for T7 - W2 \ensuremath{>} W8 for T4 - W2 \ensuremath{>} W10 for T9 (from last attempt)
\end{itemize}

\noindent\textbf{Cluster 15:} [41\%] "Strategy: Identify consistent problem worker-task assignment pairs" and "preferences"
\begin{itemize}\tightlist
\item (GPT-4o) Strategy: 1. Identify consistent problem worker-task assignment pairs. 2. Avoid those pairings directly in the next assignment. 3. Adjust other assignments iteratively while respecting preferences and suitability where data allows.
\item (Claude 3.5 Sonnet) Will try a new approach focusing on: 1. Identifying strongest worker-task fits first 2. Building around those stable assignments 3. Considering preference chains to minimize disruption
\end{itemize}

\noindent\textbf{Cluster 16:} [41\%] "Extract pairwise preferences" to "build the preference matrix"
\begin{itemize}\tightlist
\item (Gemini 1.5 Pro) Continuing from last notes. Building the preference matrix now.
\item (Gemini 1.5 Pro) Continuing from last note. Processing previous attempts one by one to update the preference matrix.
\end{itemize}

\subsubsection{Dominated by Gemini 2.5 Pro}

\noindent\textbf{Cluster 17:} [91\%] "systematically address the issues" by "incrementally fixing" "previous attempts"
\begin{itemize}\tightlist
\item (Gemini 2.5 Pro) I am on attempt 87. Many of the previous attempts were simple resets to the default assignment, which is known to be flawed. My strategy is to take the last attempt that wasn't a reset (Attempt 85) and systematically address the issues found with it.
\item (Gemini 2.5 Pro) I am on attempt 50. I have analyzed the previous attempts, particularly attempt \#47, which seems to be a more evolved solution than the simple sequential assignments that repeatedly fail. My strategy is to take the assignment from attempt \#47 and correct the two specific problems that were identified with it.
\end{itemize}

\noindent\textbf{Cluster 18:} [88\%] "address these issues," "make the following swaps," "resolve these issues"
\begin{itemize}\tightlist
\item (Gemini 2.5 Pro) To address these, I will make two swaps based on the assignment from attempt \#75: 1. I will swap the assignments of W7 and W8. So, W7 will be assigned to T10, and W8 to T4. 2. I will swap the assignments of W20 and W18. So, W20 will be assigned to T18, and W18 to T17.
\item (Gemini 2.5 Pro) To address these, I will make the following swaps in the assignment from attempt 61: - Assign W5 to T3. This displaces W6. - Assign W15 to T8. This displaces W8. - Assign the newly-available T6 to the displaced W6. - Assign the newly-available T18 to the displaced W8.
\end{itemize}

\noindent\textbf{Cluster 19:} [74\%] "This leads to the following new assignment... now submit."
\begin{itemize}\tightlist
\item (Gemini 2.5 Pro) This leads to the following new assignment, which I will now submit.
\item (Gemini 2.5 Pro) This leads to my new proposed assignment. I will now submit this new assignment.
\end{itemize}

\noindent\textbf{Cluster 20:} [67\%] "construct an assignment that respects these preferences" using "greedy strategy"
\begin{itemize}\tightlist
\item (Gemini 2.5 Pro) My strategy is to construct an assignment that respects these preferences as much as possible to avoid the instabilities that caused previous attempts to fail. I have created a plausible-looking assignment by going through workers and assigning them their highest-ranked available task, starting with the workers that have the strongest or most frequent preferences.
\item (Gemini 2.5 Pro) Let's make a straight assignment for the remaining workers and tasks as a starting point. I have a list of assignments that are confirmed to be good so I'll maintain those.\ensuremath{<}ctrl95\ensuremath{>}I've analyzed the recurring issues from past attempts and have identified several workers with strong preferences. My strategy is to build an assignment that satisfies these preferences first, and then fill in the rest of the assignments.
\end{itemize}

\noindent\textbf{Cluster 21:} [56\%] "incremental approach of fixing specific issues" from "previous attempt"
\begin{itemize}\tightlist
\item (Gemini 2.5 Pro) All other assignments from attempt 98 will be kept the same. This should fix the known problems.
\item (Gemini 2.5 Pro) This directly addresses the issues from the previous attempt. All other assignments will remain the same as in attempt 28.
\end{itemize}

\noindent\textbf{Cluster 22:} [56\%] "creating an assignment" by "incorporating these learnings" and "worker preferences"
\begin{itemize}\tightlist
\item (Gemini 2.5 Pro) I started creating an assignment by giving the highest-ranked worker their preferred task. - Assign W7 to T1. - Assign W5 to T9. - Assign W3 to T10. - Assign W4 to T6. - Assign W6 to T2. - Assign W8 to T3.
\item (Gemini 2.5 Pro) The remaining unassigned workers (W1, W3, W7, W8, W9) and tasks (T1, T2, T4, T7, T8) are then assigned as follows, leading to my proposed assignment. * W1: T1 * W3: T2 * W7: T7 * W8: T8 * W9: T4
\end{itemize}

\noindent\textbf{Cluster 23:} [55\%] "iteratively improve the assignment" by "resolving the specific problems" via "swaps"
\begin{itemize}\tightlist
\item (GPT-4.1) Preparing for attempt \#13, the persistent issues found in previous assignments revolve around not honoring specific worker preferences and their suitability for certain tasks, often resulting in chain reactions or cycles of unsatisfied preferences (i.e., W12 prefer T30, W30 has T12, etc.). The current strategy is to build a solution by explicitly handling each of the reported swap chains. Care will be taken to avoid reassigning tasks that have already received special worker assignments, and to ensure that no task is doubly assigned. The plan is to start with the last attempted mapping and perform these specific swaps as required: W11-\ensuremath{>}T19, W19-\ensuremath{>}(unassigned, so gets T47), W28-\ensuremath{>}T12, W30-\ensuremath{>}unassigned, so gets T17, W9-\ensuremath{>}T34, W24-\ensuremath{>}T9, W35-\ensuremath{>}T31, W31-\ensuremath{>}T35, W14-\ensuremath{>}T47, W47-\ensuremath{>}T38, W38-\ensuremath{>}T11. Other workers retain their previous assignments unless amendments clash with the new moves.
\item (Gemini 2.5 Pro) I am on attempt 62. The previous attempts, especially the repeated failures of the default assignment, provide a lot of information. My strategy is to take the most recent non-default assignment (attempt 60) and fix the specific problems that were identified. The two problems with attempt 60 were: (1) worker W12 preferred task T8 over their assigned task T11 (currently assigned to W9), and (2) worker W2 preferred task T1 over their assigned task T3 (currently assigned to W11). To address this, I will swap the assignments for these workers. I will assign W12 to T8 and W9 to T11. I will also assign W2 to T1 and W11 to T3. This directly addresses the feedback from the last attempt.
\end{itemize}

\noindent\textbf{Cluster 24:} [51\%] "simple sequential assignment", "baseline", "W1 to T1, W2 to T2"
\begin{itemize}\tightlist
\item (Gemini 2.5 Pro) My first attempt is a simple sequential assignment of workers to tasks (W1 to T1, W2 to T2, etc.). I am doing this to get a baseline and see what kind of feedback the system provides.
\item (Gemini 2.5 Pro) This was the first attempt. I created a simple one-to-one mapping of workers to tasks in sequential order. The goal was to get a baseline and see what kind of feedback I would get. The assignment was W1-\ensuremath{>}T1, W2-\ensuremath{>}T2, and so on.
\end{itemize}

\noindent\textbf{Cluster 25:} [49\%] "Proposed assignment", "New assignment", "W1", "T1", "Attempt"
\begin{itemize}\tightlist
\item (GPT-5) Proposed assignment (Attempt 5): \{'W1': 'T1', 'W2': 'T2', 'W3': 'T3', 'W4': 'T5', 'W5': 'T20', 'W6': 'T10', 'W7': 'T8', 'W8': 'T13', 'W9': 'T15', 'W10': 'T7', 'W11': 'T11', 'W12': 'T12', 'W13': 'T17', 'W14': 'T16', 'W15': 'T9', 'W16': 'T14', 'W17': 'T4', 'W18': 'T19', 'W19': 'T18', 'W20': 'T6'\}
\item (Gemini 3 Pro Preview) Proposed Assignment: \{'W1': 'T7', 'W2': 'T3', 'W3': 'T19', 'W4': 'T14', 'W5': 'T13', 'W6': 'T17', 'W7': 'T10', 'W8': 'T4', 'W9': 'T15', 'W10': 'T9', 'W11': 'T1', 'W12': 'T16', 'W13': 'T8', 'W14': 'T20', 'W15': 'T6', 'W16': 'T5', 'W17': 'T11', 'W18': 'T2', 'W19': 'T12', 'W20': 'T18'\}
\end{itemize}

\noindent\textbf{Cluster 26:} [43\%] "Analyzing previous attempts" to "address stability problems" and "mismatches"
\begin{itemize}\tightlist
\item (GPT-4o) The attempt number is now 95. There have been 94 previous attempts at solving this problem, as noted earlier. Assignments and issues encountered during these attempts are repetitive. I'll continue analyzing potential worker-task mappings to create a new strategy that satisfies all constraints without prior mismatches between preferences and suitability for tasks.
\item (GPT-4o) We are on attempt 21. The previous attempts have revealed several problems with the assignments of workers to tasks, including preferences and suitability mismatches. Workers frequently prefer tasks assigned to others, leading to conflicts. From analyzing the failure patterns, I will move forward incrementally by ensuring key mismatches do not repeat while testing new configurations.
\end{itemize}

\noindent\textbf{Cluster 27:} [42\%] "problems were identified" "Worker" "prefers" "more suited to it"
\begin{itemize}\tightlist
\item (Gemini 2.5 Pro) In attempt 64, the following problems were identified: 1. Worker W9, assigned to T15, prefers T18 (assigned to W18) and is better suited for it. 2. Worker W2, assigned to T8, also prefers T18 (assigned to W18) and is better suited for it.
\item (Gemini 2.5 Pro) In attempt 31, the following problems were identified: 1. Worker W8 (assigned to T8) prefers T15, and is more suited to it than W3 (assigned to T15). 2. Worker W20 (assigned to T4) prefers T14, and is more suited to it than W15 (assigned to T14).
\end{itemize}

\noindent\textbf{Cluster 28:} [37\%] "Worker preference snippets" and "Inferred preference/fitness info" regarding "suitabilities"
\begin{itemize}\tightlist
\item (Gemini 2.5 Pro) 1. **W18 -\ensuremath{>} T1**: W18 has a strong preference for T1. 2. **W14 -\ensuremath{>} T4**: W14 has a strong preference for T4. 3. **W2 -\ensuremath{>} T14**: Attempt 8 showed W2 prefers T14 over T9. 4. **W9 -\ensuremath{>} T17**: W9 has a strong preference for T17. 5. **W13 -\ensuremath{>} T11**: W13 has a strong preference for T11. 6. **W4 -\ensuremath{>} T2**: W4 has a strong preference for T2. 7. **W15 -\ensuremath{>} T7**: W15 has a strong preference for T7. 8. **W12 -\ensuremath{>} T5**: W12 has a strong preference for T5. 9. **W11 -\ensuremath{>} T9**: W11 has a strong preference for T9.
\item (GPT-5) Inferred preference/fitness info: - W9: T7 \ensuremath{>} T20 \ensuremath{>} T17 \ensuremath{>} T9 (relative order from issues encountered). Also W9 more suited than W7 for T7, more suited than W20 for T20, and more suited than W17 for T17. - W14: prefers T17 over T14; more suited than W17 for T17. - W11: prefers T4 over T11; more suited than W4 for T4. - W10: prefers T11 over T10; more suited than W4 for T11.
\end{itemize}

\noindent\textbf{Cluster 29:} [35\%] "create a new assignment" "incorporating ALL known preferences" "resolves conflicts"
\begin{itemize}\tightlist
\item (Claude 3.5 Sonnet) Let's create a new assignment incorporating ALL known preferences: 1. W16 {\textrightarrow} T12 (but we now know W13 is even better for T12) 2. W13 {\textrightarrow} T12 (confirmed better than W16) 3. W5 {\textrightarrow} T13 4. W11 {\textrightarrow} T13 (conflict with W5 - need to resolve) 5. W1 {\textrightarrow} T9 (confirmed working in previous attempts) 6. W4 {\textrightarrow} T19 (but W20 is better) 7. W20 {\textrightarrow} T19 (confirmed better than W4) 8. W6 {\textrightarrow} T18 9. W15 {\textrightarrow} T6 10. W10 {\textrightarrow} T5 11. W7 {\textrightarrow} T11 (newly discovered) 12. W9 {\textrightarrow} T14 (newly discovered)
\item (Claude 3.5 Sonnet) Let me build an assignment taking these into account: - W2 should get T1 (proven better than W1) - W5 should get T2 (proven better than W2) - W8 should get T3 (proven better than W1) - W4 hasn't been problematic in T4, keep it - W7 should get T5 (proven better than W2) - W6 hasn't been problematic in T6, keep it - W10 should get T7 (proven better than W2) - W9 hasn't been problematic in T9, keep it - W3 should get T10 (proven better than W1) - This leaves W1 with T8 (since all their other attempted positions caused problems)
\end{itemize}

\noindent\textbf{Cluster 30:} [28\%] "Proposed assignment" for "submission"; "Ready to build" "resulting assignment."
\begin{itemize}\tightlist
\item (Gemini 2.5 Pro) This leads to the following proposed assignment, which I will now submit.\ensuremath{<}ctrl42\ensuremath{>}call:submit\_assignment\{assignment:
\item (GPT-4.1) Will verify logic before sending as assignment.
\end{itemize}

\subsubsection{Dominated by GPT-4o}

\noindent\textbf{Cluster 31:} [97\%] "assigning 20 workers to 20 tasks" "mismatches in preference"
\begin{itemize}\tightlist
\item (GPT-4o) We are tasked with assigning 20 workers (W1-W20) to 20 tasks (T1-T20). After 62 attempts have been made (as we are starting attempt \#62), numerous problematic assignments persist, as outlined in the 'previous attempts' data. A common issue is the misalignment of workers' preferences for tasks based on suitability or preferences between two tasks.
\item (GPT-4o) We have 20 workers (W1 to W20) and 20 tasks (T1 to T20). The previous 45 attempts have failed due to workers being misaligned with tasks they are better suited for or prefer, as indicated in the problem feedback. Attempt 44's assignment, for example, had issues with mismatches such as W8 preferring T16 over T9, and W3 preferring T12 over T19. This iterative refinement process seems to struggle with specific worker-task preference constraints.
\end{itemize}

\noindent\textbf{Cluster 32:} [92\%] "Gathered details" on "workers," "tasks," and "previous attempts" data.
\begin{itemize}\tightlist
\item (GPT-4o) From the gathered details: - There are 50 workers (W1 to W50) and 50 tasks (T1 to T50). - The current attempt is \#10, meaning 10 attempts have been made already. - Previous attempts faced issues where specific workers preferred tasks assigned to others and were more suited to those tasks.
\item (GPT-4o) Key facts: 1. There are 10 worker IDs: W1, W2, W3, W4, W5, W6, W7, W8, W9, W10. 2. There are 10 task IDs: T1, T2, T3, T4, T5, T6, T7, T8, T9, T10. 3. It is currently the 27th attempt, meaning there have been 26 previous attempts to assign workers to tasks.
\end{itemize}

\noindent\textbf{Cluster 33:} [88\%] "Prioritize resolving conflicts observed repeatedly" and "leveraging patterns from mistakes"
\begin{itemize}\tightlist
\item (GPT-4o) Strategies Moving Forward: - I will avoid repeatedly assigning tasks to workers where conflicts persist based on historical data (e.g., W9 must NOT get T6, as this was a mismatch repeatedly). - Prioritize tasks for workers who have been flagged mismatched in prior attempts. - Rotate assignments incrementally, leveraging patterns from mistakes to avoid stale conflicts.
\item (GPT-4o) Proposed Strategy: I need to create an assignment that resolves the repeated mismatches by carefully observing priority conflicts. I will manually modify priority handling to explore better combinations for tasks preferred by specific high-conflict workers, namely W4-based dissatisfaction happens. Insight W-3 cluster data suffers wrong symmetry positioning dynamically? Consider worker-related any designs"
\end{itemize}

\noindent\textbf{Cluster 34:} [88\%] "analyzing previous failed attempts" to "resolve preference and suitability mismatches"
\begin{itemize}\tightlist
\item (GPT-4o) I have gathered information on the worker and task IDs, the total attempt number (which is 93), and the reasons previous attempts have failed. Notably, recurring issues involve workers preferring tasks assigned to others due to their suitability, but the problem is further impacted by other conflicts when resolving initial mismatches. A clear strategy involves focusing on understanding each worker's most suited tasks and trying to create an assignment that satisfies as many preferences as possible simultaneously.
\item (GPT-4o) We have completed 48 attempts so far, and the previous failures provide insights into worker preferences and how certain workers are better suited for particular tasks compared to others. These preferences create conflicts, which lead to invalid assignments. For example, many issues arose around workers preferring tasks assigned to lower-suited workers. Looking ahead, we want to incorporate these insights to avoid similar conflicts and build a better assignment that compensates for past mistakes. We should try more significant changes in the allocation strategy, grouping notably preferred tasks to workers better suited to perform them.
\end{itemize}

\noindent\textbf{Cluster 35:} [79\%] Workers and tasks: ['W1'...'W20'] and ['T1'...'T20']
\begin{itemize}\tightlist
\item (GPT-4o) We are working with 10 workers (IDs: ['W1', 'W2', 'W3', 'W4', 'W5', 'W6', 'W7', 'W8', 'W9', 'W10']) and 10 tasks (IDs: ['T1', 'T2', 'T3', 'T4', 'T5', 'T6', 'T7', 'T8', 'T9', 'T10']).
\item (GPT-4o) \#\#\# Context Workers: 20 workers (W1 to W20) Tasks: 20 tasks (T1 to T20)
\end{itemize}

\noindent\textbf{Cluster 36:} [77\%] "ensure all 50 tasks are used exactly once" "without omission"
\begin{itemize}\tightlist
\item (GPT-4o) Immediate next steps: 1. Revisit the assignment logic and strictly ensure all 50 tasks (from T1 to T50) are used **exactly once**. 2. Verify that all workers (W1 to W50) are included and no invalid IDs (like W51) are introduced. 3. Address this systematically to ensure compliance with constraints before submitting again.
\item (GPT-4o) Action Plan: 1. Perform a full comparison of the assigned tasks against the complete list (T1 to T50). 2. Locate missing tasks like T48 and assign them to unassigned workers or reassign currently assigned tasks to create space without creating duplicate assignments. 3. Validate the assignment set against the required task list before submission to eliminate task repetition or omission errors.
\end{itemize}

\noindent\textbf{Cluster 37:} [77\%] "ensure all tasks are accounted for" and "complete task coverage"
\begin{itemize}\tightlist
\item (GPT-4o) Attempt \#15 failed due to a malformed input issue: the assignment was missing tasks T43 and T37. This error appears to have occurred because of incomplete coverage of tasks when attempting to assign them all to workers. Moving forward, I need to improve the methodology to ensure that every task is explicitly included in the final assignment to avoid missing tasks. This issue needs to be resolved systematically and checked prior to submission.
\item (GPT-4o) Previous attempt failed because task 'T23' was not included in the assignment. This oversight likely resulted from a manual handling of task assignments. Moving forward, ensure all tasks (1 to 50) are accounted for in the proposed assignment.
\end{itemize}

\noindent\textbf{Cluster 38:} [65\%] "Worker preferences and suitability," "repeated failure modes," "conflicting preferences"
\begin{itemize}\tightlist
\item (GPT-4o) Two common issues seem to emerge from previous attempts: 1. Certain workers would have preferred a task assigned to another worker and were better suited to it. 2. Repeated clashes in workers' preferences against the proposed assignments.
\item (Claude 3.5 Sonnet) 1. Key insights from failures: - Workers often prefer tasks with numbers close to their own ID - When a worker is assigned to a task, we need to check if they would prefer another task AND if they are more suited to it than the current assignee - Simply swapping pairs of problematic assignments doesn't always work as it can create new conflicts
\end{itemize}

\noindent\textbf{Cluster 39:} [63\%] "mismatches of worker task preferences" and "recurring issues" regarding "suitability"
\begin{itemize}\tightlist
\item (GPT-4o) Key observations: 1. The problem persists for workers like W3, W20, W15, W34, W40, W41, etc. Repeatedly, these show up as mismatched assignments due to their preferences. 2. Certain workers repeatedly have preferences for tasks that conflict with assignments already made (e.g., W34 preferring T25, W20 preferring T41).
\item (GPT-4o) The pattern of issues heavily revolves around mismatches of worker task preferences, especially for workers like W15 (often prefers T13), W8 (wants T20), and W10 (preferring T19 but often mismatched). Past attempts also highlight that many assignments repeatedly failed due to attention drift between primary constraints like mapping each worker correctly vs globally shuffling secondary chains after unintended knock-on clash ripple.
\end{itemize}

\noindent\textbf{Cluster 40:} [54\%] "Gathered information" and "details" "after reviewing the scenario"
\begin{itemize}\tightlist
\item (GPT-4o) We have multiple pieces of information as follows:
\item (GPT-4o) Based on the information so far:
\end{itemize}

\noindent\textbf{Cluster 41:} [46\%] "Step-by-step corrections," "refine this optimization process," "track the logic"
\begin{itemize}\tightlist
\item (GPT-4o) Step-by-step corrections are gradually improving alignment with constraints; persistence is key.
\item (GPT-4o) For next steps: - Will adjust combination reverting probable Durable combinational blocks!! Past likely success-factors will change newer Instances!! to Handle emerging symptomatic RetryPatternOpportunityCasesWards Alternativolve.
\end{itemize}

\subsubsection{Dominated by Gemini 1.5 Pro}

\noindent\textbf{Cluster 42:} [99\%] "Attempt", "Trying a random assignment."
\begin{itemize}\tightlist
\item (Gemini 1.5 Pro) Attempt 90: Trying a random assignment.
\item (Gemini 1.5 Pro) Attempt 89: Trying a random assignment.
\end{itemize}

\noindent\textbf{Cluster 43:} [95\%] Trying a random assignment.
\begin{itemize}\tightlist
\item (Gemini 1.5 Pro) Trying a random assignment.
\item (Gemini 1.5 Pro) Trying a random assignment to see if it works.
\end{itemize}

\noindent\textbf{Cluster 44:} [88\%] "Analyzing previous attempts data to identify recurring conflicts"
\begin{itemize}\tightlist
\item (Gemini 1.5 Pro) Attempt 6: Analyzing previous attempts data to identify recurring conflicts and prioritize resolving them.
\item (Gemini 1.5 Pro) Attempt 87: Analyze previous attempts to identify recurring conflicts and prioritize resolving them.
\end{itemize}

\noindent\textbf{Cluster 45:} [81\%] "Prioritizing resolving conflicts" "identified in previous attempts" "iteratively" "recurring"
\begin{itemize}\tightlist
\item (Gemini 1.5 Pro) Attempt 96: Prioritizing resolving conflicts involving W8 and W6. Found a recurring issue with W8-T8 and W6-T1. Attempting a solution by assigning W6 to T1 and W8 to T8.
\item (Gemini 1.5 Pro) Attempt 22. Trying to resolve the conflict between W2 and W6 for T8. Also, keeping track of previous conflicts to avoid repeating mistakes.
\end{itemize}

\noindent\textbf{Cluster 46:} [78\%] "Analyzing previous attempts" to "prioritize resolving conflicts" and "worker-task pairs"
\begin{itemize}\tightlist
\item (GPT-4o) Attempt \#41 begins with analyzing previous attempt data to understand common failures and conflicts encountered previously. Notably, workers consistently prefer different tasks than those they were assigned, such as W4 preferring T10 over T5. The previous data will guide a new strategy to resolve these conflicts. Let's emphasize matching workers to their preferred tasks incrementally and adjusting assignments to remove these conflicts iteratively per worker priorities and suitability.
\item (Gemini 1.5 Pro) Attempt 70: Previous attempts show there are conflicts between worker preferences and assigned tasks. Will try to prioritize known conflicts.
\end{itemize}

\noindent\textbf{Cluster 47:} [51\%] "Attempting to solve the assignment problem using trial and error."
\begin{itemize}\tightlist
\item (Gemini 1.5 Pro) Attempting to solve the assignment problem using trial and error. Keeping track of previous attempts and problems encountered to make informed decisions.
\item (Gemini 1.5 Pro) Attempting to solve the assignment problem using trial and error. Keeping track of previous attempts and their feedback.
\end{itemize}

\noindent\textbf{Cluster 48:} [38\%] "Trying to fix previous errors" by "addressing the remaining issues"
\begin{itemize}\tightlist
\item (Gemini 1.5 Pro) Attempt 92: Trying to fix previous errors by assigning T8 to W16 and W17 to T15. Also assigning T1 to W12, T2 to W14, and T10 to W5.
\item (Claude 3.5 Sonnet) Let's make a new attempt addressing the remaining issues: - Put W1 in T18 - Put W9 in T8 - Keep the successful assignments from previous attempts
\end{itemize}

\noindent\textbf{Cluster 49:} [32\%] "Analyze previous attempts" to "identify and prioritize worker preferences"
\begin{itemize}\tightlist
\item (Gemini 2.5 Pro) The previous attempts give a lot of information about worker preferences. I will try to summarize them.
\item (Gemini 2.5 Pro) 1. **Analyze previous attempts:** I have analyzed the 7 previous attempts (0-6) and extracted worker preferences and skill constraints.
\end{itemize}

\noindent\textbf{Cluster 50:} [29\%] "Analyzing previous attempts" and "Reviewing previous attempts"
\begin{itemize}\tightlist
\item (Claude 3.5 Sonnet) Analyzing attempt \#70:
\item (Claude 3.5 Sonnet) Attempt 87 Analysis - After Reading 86 Previous Attempts:
\end{itemize}

\noindent\textbf{Cluster 51:} [26\%] "trial-and-error approach" to "resolve these problems" by "swapping tasks"
\begin{itemize}\tightlist
\item (GPT-4.1) 2. With those committed, fill in the remaining workers/tasks by maintaining one-to-one mapping, avoiding direct repeats of problematic cycle assignments observed in the last 5 attempts.
\item (Gemini 2.5 Pro) Any remaining unassigned workers and tasks will be arbitrarily paired up to complete the assignment. This greedy approach based on a global score should hopefully produce a more stable assignment and avoid the cycles of issues I was seeing before.
\end{itemize}

\noindent\textbf{Cluster 52:} [17\%] "Analysis of previous attempts" and "History of attempts and issues"
\begin{itemize}\tightlist
\item (GPT-4o) In attempt 0, we learned about W16's preference for T12 and W5's preference for T20. In attempt 1, W11's preference for T3 and W1's preference for T9 caused issues. In attempt 2, W5's preference for T13 and W4's preference for T19 were key. In attempt 3, W17's preference for T13 and W11's preference for T14 emerged. In attempt 4, W12's preference for T4 and W9's preference for T13 were highlighted.
\item (Claude 3.5 Sonnet) 2. From attempt \#50, we know: - W8 prefers T7 over T5 and is more suited than W5 - This suggests W8 should get T7
\end{itemize}

\subsubsection{Dominated by Gemini 3 Pro Preview}

\noindent\textbf{Cluster 53:} [74\%] "Reasoning for Attempt" / "Strategy" / "Incorporating feedback" / "displaces" / "prefers" / "Assign"
\begin{itemize}\tightlist
\item (Gemini 3 Pro Preview) Reasoning for Attempt 13: Based on previous attempts, I've deduced a stable-ish configuration. Key constraints identified: - W1 prefers T10 \ensuremath{>} T4 \ensuremath{>} T1. T10 prefers W1 \ensuremath{>} W5. -\ensuremath{>} W1: T10. - W5 prefers T10 \ensuremath{>} T5. But W1 beats W5 for T10. T5 prefers W20 \ensuremath{>} W5. W20 prefers T18 \ensuremath{>} T5. So W5 is safe at T5 if W20 is at T18. -\ensuremath{>} W5: T5. - W20 prefers T18 \ensuremath{>} (T5?) \ensuremath{>} T17 \ensuremath{>} T6. -\ensuremath{>} W20: T18. - W18 prefers T7 \ensuremath{>} T20 \ensuremath{>} T18. T7 prefers W18 \ensuremath{>} W14. -\ensuremath{>} W18: T7. - W14 prefers T3 \ensuremath{>} T7 \ensuremath{>} T14. T3 prefers W14 \ensuremath{>} W15. -\ensuremath{>} W14: T3. - W3 prefers T14 \ensuremath{>} T10 \ensuremath{>} T3. T14 prefers W11 \ensuremath{>} W3. So W3 cannot have T14. T10 prefers W1. T3 prefers W14. So W3 must go elsewhere. W3 was at T1 in Att 12 without complaint. -\ensuremath{>} W3: T1. - W11 prefers (T1, T14) \ensuremath{>} T19. T14 prefers W11 \ensuremath{>} W3. So W11 can have T14. -\ensuremath{>} W11: T14. - W19 prefers T11 \ensuremath{>} T19. T11 prefers W19 \ensuremath{>} W17. -\ensuremath{>} W19: T11. - W17 prefers T6, T19 \ensuremath{>} T11 \ensuremath{>} T20 \ensuremath{>} T17. W19 holds T11. W17 blocked T6 in Att 11 (W17 \ensuremath{>} W10). T6 prefers W17. -\ensuremath{>} W17: T6. - W10 was at T6 (bumped by W17). W10 was at T19 in Att 12. -\ensuremath{>} W10: T19. - W15 blocked from T15, T3, T14, T1, T19. Proposed W15 -\ensuremath{>} T20. - W9 prefers T15 \ensuremath{>} T10. T15 prefers W9 \ensuremath{>} W15. -\ensuremath{>} W9: T15. - W4 prefers T17 \ensuremath{>} T9 \ensuremath{>} T13. T9 prefers W4 \ensuremath{>} W10. - W6 prefers T17 \ensuremath{>} T20 \ensuremath{>} T6. - T17 prefers W6 \ensuremath{>} W15 and W4 \ensuremath{>} W15. - Unresolved: Does T17 prefer W6 or W4? - I will try W6 -\ensuremath{>} T17 and W4 -\ensuremath{>} T9. If this fails (W4 blocks T17), I will flip them next time. - W7 at T4 (bumped from T1 by W3?). Att 5 W1 blocked T4. Now W1 is at T10. -\ensuremath{>} W7: T4. - W13 at T13. (W4 is at T9/T17, so safe). - W8, W12, W16, W2 stay at T8, T12, T16, T2.
\item (Gemini 3 Pro Preview) Attempt 7 Strategy: Based on Attempt 6 feedback: 1. W10 prefers T9 over W16. -\ensuremath{>} Assign W10 to T9. W16 displaced. 2. W19 prefers T3 over W3. -\ensuremath{>} Assign W19 to T3. W3 displaced.
\end{itemize}

\noindent\textbf{Cluster 54:} [72\%] "Chain reactions" of "displaced" workers and "resolved displacements" to "filling the gaps"
\begin{itemize}\tightlist
\item (Gemini 3 Pro Preview) Chain reactions: - W11 was T22 (now W1), moves to T10. - W12 was T26, moves to T48. T26 vacant. - W50 was T50, moves to T32. T50 taken by W8. - W1 was T39, moves to T22. T39 vacant. - W8 was T44, moves to T50. T44 vacant.
\item (Gemini 3 Pro Preview) Confirming others: W38 -\ensuremath{>} T9 (Fixes Att 12 BP). W17 -\ensuremath{>} T4. W4 -\ensuremath{>} T22. W5 -\ensuremath{>} T42. W1 -\ensuremath{>} T1. W2 -\ensuremath{>} T33. W8 -\ensuremath{>} T6. W19 -\ensuremath{>} T12. W32 -\ensuremath{>} T27. W50 -\ensuremath{>} T45. W39 -\ensuremath{>} T11.
\end{itemize}

\noindent\textbf{Cluster 55:} [63\%] "Based on previous attempt feedback" to "resolve valid complaints"
\begin{itemize}\tightlist
\item (Gemini 3 Pro Preview) Attempt 31. Basing on Attempt 30 feedback and previous history. 1. W33 -\ensuremath{>} T13 (from W6). Problem A30: W33 \ensuremath{>} W6 for T13. 2. W15 -\ensuremath{>} T17 (from W32). Problem A30: W15 \ensuremath{>} W32 for T17. 3. W2 -\ensuremath{>} T10 (from W20). Problem A30: W2 \ensuremath{>} W20 for T10. 4. W25 -\ensuremath{>} T32 (from W14). Problem A30: W25 \ensuremath{>} W14 for T32. 5. W27 -\ensuremath{>} T30 (from W46). Problem A30: W27 \ensuremath{>} W46 for T30.
\item (Gemini 3 Pro Preview) Attempt 46 Reasoning: Based on previous attempt feedback: 1. W8 \ensuremath{>} W14 on T17. Shift W8 to T17. 2. W11 \ensuremath{>} W45 on T38. Shift W11 to T38. 3. W16 \ensuremath{>} W42 on T9. Shift W16 to T9. 4. W28 \ensuremath{>} W37 on T10. Shift W28 to T10. 5. W37 \ensuremath{>} W17 on T33. Shift W37 to T33. 6. W19 \ensuremath{>} W28 on T28. Shift W19 to T28. 7. W14 \ensuremath{>} W45 on T44. Shift W14 to T44. (W43 also \ensuremath{>} W14 on T44, but W43 prefers T32). 8. W39 \ensuremath{>} W45 on T41. Shift W39 to T41. (W11 \ensuremath{>} W39 on T41, but W11 prefers T38). 9. W29 \ensuremath{>} W15 on T15? (Att 42). W29 was on T2. Move W29 to T15. 10. W42 needs a spot. Moved to T2. (Att 33 had W42 on T2 without explicit failure). 11. W45 needs a spot. Moved to T21 (Open).
\end{itemize}

\noindent\textbf{Cluster 56:} [61\%] "Proposed Assignment", "Full assignment", "Final proposed assignment", "Assignments"
\begin{itemize}\tightlist
\item (GPT-5) Final proposed assignment: W1 T3; W2 T2; W3 T1; W4 T10; W5 T20; W6 T6; W7 T18; W8 T8; W9 T16; W10 T15; W11 T12; W12 T19; W13 T13; W14 T14; W15 T9; W16 T4; W17 T5; W18 T17; W19 T11; W20 T7.
\item (Gemini 3 Pro Preview) Proposed Assignment: W1: T17, W2: T13, W3: T3, W4: T49, W5: T21, W6: T6, W7: T43, W8: T41, W9: T38, W10: T16, W11: T24, W12: T23, W13: T2, W14: T8, W15: T33, W16: T11, W17: T1, W18: T18, W19: T32, W20: T14, W21: T20, W22: T46, W23: T36, W24: T19, W25: T10, W26: T47, W27: T50, W28: T28, W29: T37, W30: T42, W31: T12, W32: T30, W33: T5, W34: T31, W35: T22, W36: T35, W37: T48, W38: T15, W39: T34, W40: T44, W41: T25, W42: T9, W43: T27, W44: T40, W45: T7, W46: T29, W47: T45, W48: T26, W49: T39, W50: T4
\end{itemize}

\noindent\textbf{Cluster 57:} [51\%] "Uncertainties," "Risks," "Other Checks," "Reasoning," and "Potential Risks"
\begin{itemize}\tightlist
\item (Gemini 3 Pro Preview) Fillers: - W2 to T3. - W10 to T8. - W3 to T19 (W3 prefers T16, but taken by W13. T19 prefers W3 \ensuremath{>} W10).
\item (Gemini 3 Pro Preview) Uncertainties: - W3 vs W20 for T4/T17. - W17 vs W11 for T7 (Assuming W17 prefers T18 over T7 or W11 \ensuremath{>} W17).
\end{itemize}

\noindent\textbf{Cluster 58:} [49\%] "Displaced workers" and "Open tasks" / "Vacant tasks"
\begin{itemize}\tightlist
\item (Gemini 3 Pro Preview) Displaced Workers: W13, W3. Vacant Tasks: T19 (from W11), T2 (from W8).
\item (Gemini 3 Pro Preview) Displaced Workers: - W1 needs a task. (Previously on T1, T18, T15). - W17 needs a task. (Previously on T17, T2, T13, T12, T3).
\end{itemize}

\noindent\textbf{Cluster 59:} [44\%] "New assignment summary (diffs from Attempt 6)" and "Resulting assignment"
\begin{itemize}\tightlist
\item (GPT-5) New assignment (delta from Attempt 45): - W34: T5 -\ensuremath{>} T23 - W5: T23 -\ensuremath{>} T5 - W37: T33 -\ensuremath{>} T4 - W3: T4 -\ensuremath{>} T33 - W10: T36 -\ensuremath{>} T28 - W8: T28 -\ensuremath{>} T36
\item (Gemini 3 Pro Preview) Summary: W4 -\ensuremath{>} T19 W11 -\ensuremath{>} T12 W13 -\ensuremath{>} T11 W16 -\ensuremath{>} T2 W20 -\ensuremath{>} T5 W7 -\ensuremath{>} T14 Other assignments remain as in Attempt 30 (which were stable/improved).
\end{itemize}

\noindent\textbf{Cluster 60:} [42\%] "Resolving displaced workers" through "chain reactions" and "cascades/cleanup"
\begin{itemize}\tightlist
\item (GPT-5) To place displaced workers without re-triggering known comparisons: - Assigned W10 to T42 (vacated by W12) to keep tasks one-to-one. - Freed T47 by moving W20 off it; assign W48 to T47 (earlier indication W48 preferred T47 over W4), then assign T37 (vacated by W48) to W24 (moved off T20). - Kept W49 at T10 (popular target) to avoid W20-vs-W49 on T10. Left W37 at T18 to avoid re-introducing W20-vs-W37 on T10. - Moved W3 to T2 to resolve W14-\ensuremath{>}T44 without displacing others too much.
\item (Gemini 3 Pro Preview) Adjustments for displaced workers (W23, W12, W40) and open tasks (T42, T31, T6, T13): - W33 likes T6 (A15 W33 \ensuremath{>} W34 for T6). Move W33 -\ensuremath{>} T6. T13 becomes free. - W23 likes T13 (A22 W23 pref T13 \ensuremath{>} W14). Move W23 -\ensuremath{>} T13. - W12 likes T31 (A12 matched T31, minor pref for T20 which is taken by W35). Move W12 -\ensuremath{>} T31. - W40 displaced from T5. T42 is free (from W1). Assign W40 -\ensuremath{>} T42.
\end{itemize}

\noindent\textbf{Cluster 61:} [40\%] "Key deductions," "constraints distilled," "Key resolutions," and "Constraints learned so far"
\begin{itemize}\tightlist
\item (GPT-5) Key deductions: - W4: T9 \ensuremath{>} T3 \ensuremath{>} T7 \ensuremath{>} T6 \ensuremath{>} T10 \ensuremath{>} T4 (from attempts 1,2,5,8,11). Therefore, giving W4 T9 (top) eliminates all W4 complaints. - W8: T2 \ensuremath{>} T4 \ensuremath{>} T8 and outranks W10 on T2, W5 on T4 (attempts 13,14). Assign W8-\ensuremath{>}T2. - W7 prefers T5 over T7 and outranks W5 on T5 (attempt 0). Assign W7-\ensuremath{>}T5. - W3: T8 \ensuremath{>} T6 \ensuremath{>} T5 \ensuremath{>} T3 (attempts 3,7,23). To avoid W3 blocking T8 when someone else holds it (W3 outranks W9 for T8), assign W3-\ensuremath{>}T8. - W2: T4 \ensuremath{>} T10 \ensuremath{>} T3 \ensuremath{>} T7 (attempts 12,18,20,21). Assign W2-\ensuremath{>}T4 (his top), removing W2-related blocks. - W10: prefers T7 over T10 (attempt 25), T10 over T1 (attempt 24), and T2 over T1 (attempt 10). Crucially, W10 outranks W1 for T7 (attempt 25), so to avoid W10 blocking, assign W10-\ensuremath{>}T7. - W1: T7 \ensuremath{>} T6 \ensuremath{>} T8 \ensuremath{>} T1 (from attempts 16,26). Since T7 must go to W10, assign W1-\ensuremath{>}T6 (also prevents any W1 vs W6-on-T6 block).
\item (GPT-5) Key constraints distilled: - T3: W6 \ensuremath{>} W15 \ensuremath{>} W3 -\ensuremath{>} assign W6. - T5: W10 \ensuremath{>} W11; W20 \ensuremath{>} W18 (no relation vs W13) -\ensuremath{>} leave with W13 from A19 base, move W10 elsewhere. - T6: W15 \ensuremath{>} W10 \ensuremath{>} W5 -\ensuremath{>} assign W15. - T7: W17 \ensuremath{>} W7 -\ensuremath{>} assign W17. - T8: W14 \ensuremath{>} W8; unknown vs W7 -\ensuremath{>} keep W14 at T8 (as in A19 no complaint when W14 at T15 and W7 at T8; but safer to keep W14 at T8 when feasible). - T9: W5 \ensuremath{>} W8 -\ensuremath{>} assign W5. - T10: no strong claims -\ensuremath{>} keep W3. - T11: W12 \ensuremath{>} W10 \ensuremath{>} W18 \ensuremath{>} W11; W4 may want T11 but historical configs with W12 at T11 drew no W4 complaint -\ensuremath{>} assign W12. - T12: W16 \ensuremath{>} W10 -\ensuremath{>} assign W16. - T13: W4 \ensuremath{>} W9 \ensuremath{>} W6 \ensuremath{>} W18 \ensuremath{>} W13; but W4 prefers T18 over T13 and no later complaints when W9 holds -\ensuremath{>} assign W9. - T14: candidates W18 (A19 shows W18 \ensuremath{>} W15 and no other stronger claimant). Earlier W9 wanted T14 over W8 only; no evidence vs W18 -\ensuremath{>} assign W18. - T15: W4 \ensuremath{>} W14 \ensuremath{>} W13 \ensuremath{>} W5; but W4 prefers T18 \ensuremath{>}= T15 (no complaint in A19 when W4 had T18 and W14 had T15), so safe to give T15 to W14. - T16: Known stronger than W4: W20, W8. But A19 showed no complaint from W4 with T18 while T16 held by W18, suggesting T16 \ensuremath{<} T18 for W4; choose W20 to avoid any W20-\ensuremath{>}T16 desire. - T17: Risky if W7 holds due to W20 \ensuremath{>} W7; avoid if W20 not on T16. With W20 at T16, W20 may not prefer T17 over T16; assign W10 to T17 to avoid W20\ensuremath{>}W7 issue. - T19: W2 \ensuremath{>} W4; W14 \ensuremath{>} W19 only -\ensuremath{>} assign W2. - T20: W11 \ensuremath{>} W18 -\ensuremath{>} assign W11.
\end{itemize}

\noindent\textbf{Cluster 62:} [40\%] "Proposed Assignment" W1: T1 W2: T2 W3: T3 W4: T4
\begin{itemize}\tightlist
\item (Gemini 3 Pro Preview) Proposed Assignment: W1: T2 W2: T1 W3: T5 W4: T6 W5: T9 W6: T7 W7: T3 W8: T4 W9: T8 W10: T10
\item (Gemini 2.5 Pro) This leads to the following proposed assignment: * W1: T7 * W2: T2 * W3: T10 * W4: T4 * W5: T5 * W6: T6 * W7: T1 * W8: T3 * W9: T9 * W10: T8
\end{itemize}

\subsubsection{Dominated by GPT-5}

\noindent\textbf{Cluster 63:} [92\%] "address all reported blocking pairs" and "minimize risk" through "reassignments"
\begin{itemize}\tightlist
\item (GPT-5) Attempt 83 plan: Start from Attempt 82 and fix the five reported blocking pairs. - Issues in Attempt 82: 1) W49 prefers T7 over W4 (T7). 2) W23 prefers T15 over W10 (T15). 3) W49 prefers T10 over W12 (T10). 4) W19 prefers T7 over W4 (T7). 5) W24 prefers T49 over W30 (T49). - Strategy: Give W49 T10 (addresses \#3 directly) and give W19 T7 (addresses \#4; removes W4 from T7, which also addresses part of \#1). Move W23 to T15 (addresses \#2), and W24 to T49 (addresses \#5). Then reassign displaced workers using freed tasks to minimize risk of new blocking pairs and align with known preferences: {\textbullet} Displaced: W12 (from T10), W10 (from T15), W30 (from T49), W4 (from T7). Freed tasks: T43 (from W23), T1 (from W24), T29 (from W49), T4 (from W19). {\textbullet} Assign W11 -\ensuremath{>} T1 (he has previously preferred T1), W10 -\ensuremath{>} T24 (he has shown preference for T24), W4 -\ensuremath{>} T4 (a common target for W4), W30 -\ensuremath{>} T29, W12 -\ensuremath{>} T43. - Constructed new assignment: W1 T44; W2 T39; W3 T20; W4 T4; W5 T37; W6 T19; W7 T22; W8 T11; W9 T50; W10 T24; W11 T1; W12 T43; W13 T32; W14 T8; W15 T6; W16 T13; W17 T25; W18 T23; W19 T7; W20 T35; W21 T47; W22 T48; W23 T15; W24 T49; W25 T18; W26 T30; W27 T21; W28 T16; W29 T9; W30 T29; W31 T36; W32 T45; W33 T28; W34 T42; W35 T26; W36 T41; W37 T40; W38 T5; W39 T33; W40 T14; W41 T46; W42 T3; W43 T31; W44 T12; W45 T2; W46 T17; W47 T38; W48 T27; W49 T10; W50 T34. - Rationale: This directly satisfies all five reported blocking pairs. Potential risks (watched): W31 has shown interest in T43 before, now assigned to W12; unclear if W31 is more suited than W12. W24 previously preferred T49 over T1, which we satisfied; earlier mention of W24 liking T29 over T21 may not apply when on T49. Will iterate based on feedback.
\item (GPT-5) Attempt 60 planning: Start from attempt 59 and address all blocking pairs reported. - From A59: (1) W25 wanted T33 over T35 (more suited than W48) -\ensuremath{>} move W25-\ensuremath{>}T33, displace W48. - (2) W40 wanted T48 over T41 (more suited than W41) -\ensuremath{>} move W40-\ensuremath{>}T48, displace W41. - (3) W18 wanted T41 over T2 (more suited than W40) -\ensuremath{>} after moving W40 to T48, assign W18-\ensuremath{>}T41. - (4) W8 wanted T7 over T30 (more suited than W11) -\ensuremath{>} move W8-\ensuremath{>}T7, displace W11. - (5) W27 wanted T36 over T11 (more suited than W30) -\ensuremath{>} move W27-\ensuremath{>}T36, displace W30. Now free tasks: T2, T30, T35, T11. Displaced workers needing tasks: W11, W30, W41, W48, and also W44 after reallocating T5 to W48 (per prior A56 \#1 W48 more suited to T5 than W44, so to preempt that recurring blocking pair we assign W48-\ensuremath{>}T5 and move W44 off T5). - Assign W30-\ensuremath{>}T11 (matches earlier preference for T11 over T45 per A43 \#2). - Assign W11-\ensuremath{>}T2 (no known dominating claims versus W11; keeps W33 on T12 which he prefers per A55 \#1). - Assign W41-\ensuremath{>}T30 (a freed slot; avoids T24 since W41\ensuremath{>}W21 at T24 was a previous pair when W21 held T24; here W24 holds T24 so uncertain; T30 seems safer). - Assign W44-\ensuremath{>}T35 (free after W25 moved off; avoids T5 where W48 outranks W44 per A56 \#1). Full assignment built accordingly. Expect these moves to satisfy all five violations from A59 and preempt the W48 vs W44 on T5 conflict. Remaining potential risks: W6 has historically desired T30, but unknown if outranks W41; W15 has desired T35 but unknown relative to W44. We'll submit and iterate based on feedback.
\end{itemize}

\noindent\textbf{Cluster 64:} [79\%] "fix the five reported blocking pairs while minimizing collateral changes"
\begin{itemize}\tightlist
\item (GPT-5) Strategy: Start from Attempt 36 (latest) and fix the five reported blocking pairs while preserving earlier constraints that were already satisfied.
\item (GPT-5) Attempt 38 plan: Fix the five blocking pairs reported for Attempt 37 while respecting earlier insights.
\end{itemize}

\noindent\textbf{Cluster 65:} [68\%] "violations and fixes" "Key constraints" "New constraints" "Key moves"
\begin{itemize}\tightlist
\item (GPT-5) From Attempt 37, violations and fixes: - W21 preferred T40 over T9 (and is more suited than W50) -\ensuremath{>} Move W21 -\ensuremath{>} T40, reassign W50. - W42 preferred T14 over T26 occupant W27 -\ensuremath{>} Move W42 -\ensuremath{>} T14, reassign W27. - W4 preferred T13 over T15 occupant W24 -\ensuremath{>} Move W4 -\ensuremath{>} T13, reassign W24. - W37 preferred T25 over T49 occupant W41 -\ensuremath{>} Move W37 -\ensuremath{>} T25, reassign W41. Note: This establishes that at T25, W37 outranks W41; placing W41 elsewhere should not cause a new violation regarding T25. - W39 preferred T50 over T39 occupant W35 -\ensuremath{>} Move W39 -\ensuremath{>} T50, reassign W35.
\item (GPT-5) Key constraints from Attempt 43 and fixes: 1) W21 preferred T1 over T25 and outranked W18. Action: moved W21 -\ensuremath{>} T1 (freed T25) and reassigned W18 later. 2) W41 preferred T12 over T48 and outranked W37. Action: moved W41 -\ensuremath{>} T12 (freed T48) and reassigned W37 later. 3) W9 preferred T49 over T23 and outranked W4. Action: moved W9 -\ensuremath{>} T49 (freed T23) and reassigned W4 later. 4) W16 preferred T32 over T9 and outranked W35. Action: moved W16 -\ensuremath{>} T32 (freed T9) and reassigned W35 later. 5) W16 also preferred T49 over T9 vs W4, but (3) already vacated T49 from W4, eliminating that blocking pair.
\end{itemize}

\noindent\textbf{Cluster 66:} [64\%] "addressed the five reported blocking pairs" to "Fix Attempt" structure
\begin{itemize}\tightlist
\item (GPT-5) Starting from Attempt 90, addressed the five reported blocking pairs: - W47 prefers/is more suited to T7 than W10 -\ensuremath{>} reassign T7 to W47. - W37 prefers/is more suited to T10 than W40 -\ensuremath{>} reassign T10 to W37. - W30 prefers/is more suited to T42 than W4 -\ensuremath{>} reassign T42 to W30. - W17 prefers/is more suited to T22 than W25 (plus W47 also wanted T22 over W25). To break the immediate blocking pair, moved T22 to W17 (since we already gave W47 T7).
\item (GPT-5) Goal: Fix Attempt 81{\textquoteright}s five blocking pairs while preserving previous hard-won structure. From Attempt 81 problems: - W35 prefers T17 vs W48 and is more suited. So assign W35-\ensuremath{>}T17 (displacing W48). - W3 prefers T49 vs W35 and is more suited. After moving W35 off T49, assign W3-\ensuremath{>}T49. - W18 prefers T18 vs W22 and is more suited. To block this without necessarily giving W18 T18 (which earlier caused W45 to want it), assign someone known to outrank W18 at T18: W45 (based on Attempt 18 where W45 outranked W18 for T18). So set W45-\ensuremath{>}T18, freeing T21. - W3 also preferred T33 vs W50. To block, assign T33 to someone more suited than W3 at T33; prior evidence (Attempt 68 \#1) says W34 is more suited than W3 for T33. So set W34-\ensuremath{>}T33 (freeing T36), and move W50 off T33. - W32 preferred T18 vs W22 in Attempt 81; with W45 now at T18 (likely stronger than W32), this should block that pair without giving W32 T18.
\end{itemize}

\noindent\textbf{Cluster 67:} [61\%] "Risks/unknowns to monitor" and "potential residual risk" regarding "unknown preferences"
\begin{itemize}\tightlist
\item (GPT-5) Risks/unknowns to monitor: - W18 may still prefer T1 and could outrank W14 (unknown). We{\textquoteright}ll see if this triggers a new complaint. - W11 is very strong for T45; kept W11 at T45 to avoid known blocking against W32. If W11 prefers T21 over T45 (unclear), moving W31 to T21 could trigger a new complaint; but prior data suggests W11 likes T45. - T28 may be desired by W13 historically; if W13 outranks W45 for T28, we may see a new complaint. - W31 previously preferred T19 over T35; by giving him T21, ensure he doesn{\textquoteright}t still block for T19 (unclear relative prefs). If he still prefers T19 over T21 and outranks W6, we may need to adjust in a later attempt.
\item (GPT-5) Other risk considerations: - W11 has historically contested T24/T21 in some runs, but recent feedback didn{\textquoteright}t flag those in Attempt 98. - W1 now has T4 (a stated preference), which should remove his earlier T4 complaint and reduce chance of him contesting T8. - W18 now has T23 (another stated preference) while T1 goes to W49 (also a stated preference). This should eliminate problems (2),(3),(5) simultaneously.
\end{itemize}

\noindent\textbf{Cluster 68:} [53\%] "Fix only the reported blocking pairs" to "converge to a stable assignment"
\begin{itemize}\tightlist
\item (GPT-4.1) Once the critical pairs are satisfied, finish with available slots. This could reduce the persistent blocking pairs and potentially yield a valid assignment.
\item (GPT-5) If this fails, read the next blocking pair from feedback and iteratively adjust, progressively locking satisfied pairs to converge to a stable assignment.
\end{itemize}

\noindent\textbf{Cluster 69:} [53\%] "Observations from history" to "resolve the last known conflict"
\begin{itemize}\tightlist
\item (GPT-5) Observations from history: - W19 prefers T9 over T10 and is more suited than W2 for T9 (Attempt 62). Assign W19 -\ensuremath{>} T9 to resolve. - W1 prefers T6 over T11 and is more suited than W6 for T6 (Attempt 62). Swap W1 -\ensuremath{>} T6 and move W6 -\ensuremath{>} T11, which W7 does not prefer over T16 (A36), so should be safe. - To free T9 from W2, reassign W2 to a task they also favor without triggering prior dominance conflicts. W2 has strong claims on T14 provided W20 gets a task they prefer more than T14 (A14 showed W20 \ensuremath{>} W2 on T14 only when W20 was on T13), but W20 prefers T10 over T14 (A40) and T1/T17 over T14 as well (A38, A15). Therefore, move W20 -\ensuremath{>} T10. - Keep the rest of Attempt 62 intact to minimize new issues.
\item (GPT-5) Key fixes from attempt \#71: - Move W16 -\ensuremath{>} T20 (attempt 71 reported W16 prefers T20 and T20 prefers W16 over its then-assignee W3). - Move W3 -\ensuremath{>} a task they prefer over T12 but that will not introduce a block. W3 prefers T10 \ensuremath{>} T12 (attempt 48), and T10 also prefers W3 over some others (attempt 48). So assign W3 -\ensuremath{>} T10 and free T12. - Give T12 to W8 (their top among revealed options: W8 prefers T12 \ensuremath{>} T17 \ensuremath{>} T8/1/6 and T12 prefers W3 over W8, but since W3 now has T10 and prefers T10 \ensuremath{>} T12, (W3, T12) cannot block).
\end{itemize}

\subsubsection{Dominated by o4-mini}

\noindent\textbf{Cluster 70:} [92\%] "stable matching problem" using "Gale-Shapley" to "eliminate blocking pairs"
\begin{itemize}\tightlist
\item (o4-mini) Attempt 98: Observed that each failure indicates a blocking pair between workers and tasks {\textemdash} workers have strict preferences and tasks have strict preferences (suitability) We need to compute a stable matching for a 20{\texttimes}20 instance. We have two preference lists: (1) each worker{\textquoteright}s ranking over tasks, (2) each task{\textquoteright}s ranking over workers. Both can be inferred from the log of blocking pairs (whenever w\_i prefers t\_j to his match and is more suited than current w\_k assigned to t\_j, that implies w\_i ranks t\_j above t\_i and task t\_j ranks w\_i above w\_k). The final assignment must be stable with respect to these lists. Plan: Collect all revealed comparisons and run Gale-Shapley with workers proposing. Then submit the resulting stable matching dictionary mapping each worker to a task.
\item (o4-mini) Attempt 66: Collected worker and task lists and all previous attempts data (0{\textendash}65). Recognize this is a stable matching problem with blocking pairs detected. Each violation indicates a blocking pair: worker W prefers task T and T prefers W over its current occupant. Our goal: find a worker{\textendash}task stable matching. Plan: implement Gale{\textendash}Shapley: treat workers as proposers, assume we have explicit preference/suitability lists. But we lack full preference lists; however pairwise comparisons can be inferred on the fly. In absence of explicit lists, we might approximate by ordering workers and tasks lexicographically; then apply GS to derive a candidate stable assignment. Next: assign tasks to workers sorted by some heuristic, then test for no blocking pairs. Will attempt lex-order matching by GS if possible.
\end{itemize}

\noindent\textbf{Cluster 71:} [83\%] "Exhausted all available tools," "requesting preference data from user"
\begin{itemize}\tightlist
\item (o4-mini) Analysis is too complex to derive manually. Need to extract pairwise preference data from violation logs and run Gale-Shapley. Suggest writing a script to parse violations into worker prefs and task prefs, then compute stable match. Can't finish manually.
\item (o4-mini) All available tools have been used. There are no further tools to retrieve the missing preference data. We cannot construct a valid assignment without knowing workers{\textquoteright} preferences or suitability scores. Please provide the preference lists or a scoring matrix to proceed with generating a stable matching.
\end{itemize}

\noindent\textbf{Cluster 72:} [80\%] "reconstruct partial preferences" from "blocking pairs" to "run Gale-Shapley"
\begin{itemize}\tightlist
\item (o4-mini) Plan: We need to find a stable matching between workers and tasks. Use blocking pair data from previous attempts to reconstruct partial preferences: - For each complaint: worker X prefers task B over assigned A, and task B prefers X over current Y. - From these we build directed graphs (worker preference graph and task preference graph). - Then complete these into full rankings arbitrarily (topologically sort partial orders, then append the rest arbitrarily). - Run Gale-Shapley: workers propose in order of their preference lists, tasks accept best suitor according to their completed lists. - Output the resulting stable assignment.
\item (o4-mini) We have partial ordering preferences reported via blocking pairs: each report indicates worker Wi prefers task Tj over its assigned task, and Wi is more suited (task ranks Wi higher than current occupant Wk). We need to find stable matching. Approach: implement Gale-Shapley with workers proposing to tasks. But we need workers' full preference lists and tasks' full ranking lists. We'll reconstruct these lists from partial data. For each worker, accumulate tasks they have been reported preferring over current assignments, then fill missing ranks arbitrarily after known ones. For each task, accumulate the partial ranking of workers from stronger to weaker per blocking pairs (Wi more suited to Tj than Wk implies Tj ranks Wi over Wk). With these partial orders, we can produce a stable matching. However, reconstructing full rankings may be ambiguous; but maybe there's a unique stable matching under any completion. Next: parse all previous blocking pairs to build these constraints. Then run Gale-Shapley on this partial ranking.
\end{itemize}

\noindent\textbf{Cluster 73:} [75\%] "Initial attempt: identity mapping... No constraints known yet."
\begin{itemize}\tightlist
\item (o4-mini) Attempting initial identity mapping: assign Wi -\ensuremath{>} Ti for all i from 1 to 20. This is a neutral guess to reveal constraints via validator feedback.
\item (o4-mini) Attempt identity mapping: assign Wi to Ti for all i=1..10. No constraints known.
\end{itemize}

\noindent\textbf{Cluster 74:} [58\%] "resolve all known blocking pairs" by "swapping workers" or "cycles"
\begin{itemize}\tightlist
\item (o4-mini) Strategy: collected all blocking pairs from previous attempts, created banned tasks for each worker. Then looked for a cyclic shift assignment that avoids all banned edges. Chose shift by +5 mod20: Wi -\ensuremath{>} T(i+5) (wrapping). This satisfies no worker assigned to a banned task. Hoping it's stable since covers all known blocking pairs.
\item (GPT-5) Strategy: Start from identity mapping. Resolve all known blocking pairs from Attempt 0 by swapping workers with the tasks they prefer and for which they are said to be more suited than the current assignees. Specifically: (1) W24 prefers T2 over T24 and is more suited than W2, so swap W24\ensuremath{<}-\ensuremath{>}W2. (2) W34 prefers T41 over T34 and is more suited than W41, so swap W34\ensuremath{<}-\ensuremath{>}W41. (3) W19 prefers T46 over T19 and is more suited than W46, so swap W19\ensuremath{<}-\ensuremath{>}W46. (4) W48 prefers T40 and T3 over T48 and is more suited than W40 and W3. Since W48 cannot take both, assign W48-\ensuremath{>}T40 and move W40-\ensuremath{>}T48, leaving W3 on T3 for now. This assignment satisfies all the previously observed blocking pairs from Attempt 0; any remaining issues will surface in feedback and can be iteratively fixed.
\end{itemize}

\noindent\textbf{Cluster 75:} [32\%] "Satisfy all previously reported blocking pairs" by "reconstructing partial preferences"
\begin{itemize}\tightlist
\item (GPT-5) Strategy: Satisfy all previously reported blocking pairs by assigning each worker to a task they explicitly preferred in some attempt, while honoring superiority comparisons. Deductions: - W3: prefers T6 over T5 (Attempt 7), so assign W3-\ensuremath{>}T6. Then W3 won{\textquoteright}t prefer T5 anymore, allowing W7-\ensuremath{>}T5 without conflict (Attempt 3). - W7: preferred T5 over T7 and more suited than W5 (Attempt 0) -\ensuremath{>} give W7-\ensuremath{>}T5. - W4: sequence of preferences from feedback: T9 \ensuremath{>} T3 \ensuremath{>} T7 \ensuremath{>} T6 \ensuremath{>} T10 \ensuremath{>} T4 (Attempts 1,2,5,8,11). Assign highest seen, W4-\ensuremath{>}T9, which avoids all earlier conflicts (displaces W9 on T9 per Attempt 11). - W10: prefers T2 over T1 (Attempt 10) -\ensuremath{>} assign W10-\ensuremath{>}T2. - W2: prefers T3 over T7 (Attempt 12) -\ensuremath{>} assign W2-\ensuremath{>}T3. - W5: prefers T6 over T10 (Attempt 6) and T4 over T7 (Attempt 9). Since W3 must take T6 (more suited than W5 per Attempt 7), assign W5-\ensuremath{>}T4 to satisfy Attempt 9 and avoid conflict. - W6: frequently the less-suited holder of T6/T3 in conflicts; give W6-\ensuremath{>}T10 where no one superior claimed it in final prefs. - W8: keep T8 (no conflicts seen). - W1: avoid T1/T2/T4 due to prior superiority claims by others; assign W1-\ensuremath{>}T7 (no conflicts mention W7 vs W1 on T7). - W9: avoid T9 (W4 more suited). Assign remaining T1 to W9; no evidence anyone now prefers T1 over current (W10 prefers T2; others have higher priorities elsewhere). Candidate assignment covers all tasks uniquely and should eliminate all previously reported blocking pairs.
\item (GPT-5) Strategy: Use revealed blocking pairs to reconstruct partial preferences on both sides and propose an assignment that eliminates all known blocks. Key insights: - W1 must take T7 to avoid recurring blocks: T7 ranks W1 \ensuremath{>} W8 \ensuremath{>} W7, and W1 prefers T7 over T4 \ensuremath{>} T9 \ensuremath{>} T1. Placing W1 at T7 removes prior W1-T7, W1-T4, and W1-T9 blocks. - W8 prefers T5,T7 over T8. With W1 at T7, assign W8-\ensuremath{>}T5 (and T5 prefers W8 over W2 and W7), removing the Attempt 11 block. - W2 prefers T5 \ensuremath{>} T8 \ensuremath{>} T2. With T5 occupied by W8 (which T5 prefers over W2), assign W2-\ensuremath{>}T8 to satisfy W2 and avoid W2-T8 and W2-T5 blocks. - W5 prefers T2 \ensuremath{>} T3 \ensuremath{>} T5; assign W5-\ensuremath{>}T2 to eliminate possible T2 blocks (T2 prefers W5 over W7) and keep W5 from preferring T3. - W9 prefers T4 \ensuremath{>} T3 \ensuremath{>} T9; assign W9-\ensuremath{>}T4 to resolve prior W9-T4/T3 blocks and avoid T4 preferring W1 over W9 conflicts since W1 is at T7 (which W1 prefers over T4). - W7 prefers T3 \ensuremath{>} T9 and T3 prefers W7 over W3. Assign W7-\ensuremath{>}T3 to avoid the Attempt 10 block. - Remaining workers W3, W4, W6, W10 assigned to remaining tasks T9, T1, T6, T10 respectively. Check all known pairs: none form blocks given current placements and preference comparisons. Proposed assignment: W1-T7, W2-T8, W3-T9, W4-T1, W5-T2, W6-T6, W7-T3, W8-T5, W9-T4, W10-T10.
\end{itemize}

\subsubsection{Dominated by GPT-4.1}

\noindent\textbf{Cluster 76:} [96\%] "persistent blocking cycles," "rotate these workers among each other's preferred tasks"
\begin{itemize}\tightlist
\item (GPT-4.1) Based on persistent blocking cycles in attempts 71-75, the contested assignments revolve around clusters of workers and tasks. The most recent promising cycle (from attempt 74) involved W1{\textrightarrow}T7, W17{\textrightarrow}T13, W6{\textrightarrow}T12, W5{\textrightarrow}T19. Attempt 75 preserved this but didn't permute sufficiently, which resulted in similar blocking pairs reappearing. My approach is to permute the contested assignments further: assign each blocking worker's desired task (from recent cycles) to a *different* worker within the cluster, creating a larger rotation and aiming to break the cycle pattern. The remainder of the mapping will preserve assignments that have not been contested in recent attempts.
\item (GPT-4.1) A review of attempts 88-89 and prior reveals a set of explicit cycles/rotations among problematic worker/task pairs. These cycles (A, B, C) are explicitly designed to address blocking pairs that have persisted after numerous attempts. All workers and tasks outside the cycles are assigned via identity mapping. The relevant cycles are: Cycle A: W17{\textrightarrow}T42, W42{\textrightarrow}T33, W33{\textrightarrow}T14, W14{\textrightarrow}T28, W28{\textrightarrow}T8, W8{\textrightarrow}T17 Cycle B: W20{\textrightarrow}T38, W38{\textrightarrow}T25, W25{\textrightarrow}T9, W9{\textrightarrow}T20 Cycle C: W31{\textrightarrow}T49, W49{\textrightarrow}T44, W44{\textrightarrow}T36, W36{\textrightarrow}T35, W35{\textrightarrow}T11, W11{\textrightarrow}T2, W2{\textrightarrow}T48, W48{\textrightarrow}T41, W41{\textrightarrow}T31 Next: Re-implement these exact assignments but look again for any overlooked pairs or extraneous instabilities. If the problem persists, consider potential cross-cycle interference or need for new smaller swaps, as core cycles appear stable in isolation.
\end{itemize}

\noindent\textbf{Cluster 77:} [90\%] "recurring issue," "cycles of blocking pairs," "analyzed past attempts"
\begin{itemize}\tightlist
\item (GPT-4.1) Summary of the test cases from previous attempts shows that there is a recurring issue: W4 wants T13 or T11, but those tasks tend to go to other workers (usually W13 and W11, respectively). The system always flags that W4 is more suited to those tasks than the workers who actually get them. Multiple cycles also arise between 'preference swaps' involving other workers who get tasks that other workers prefer and are more suited for. My next strategy will be to explicitly assign W4 to T13 and W11 to T4, and W13 to T11, then work out the rest by shifting the default orderings to avoid conflict cycles encountered. This needs close tracking of preference-based conflicts.
\item (GPT-4.1) After 60+ attempts, we keep encountering complaints like: 1) a worker assigned to a task, but another worker who is less suited got a preferred task; 2) a worker is more suited to and/or prefers a task another worker received. The complaints often revolve around certain workers (W9, W19, W13, W15, etc.) and certain tasks (T17, T20, T18, etc.). Swapping assignment pairs, especially avoiding cyclical grabs of the same highly contested tasks, is key. Prior attempts show that naive or purely cyclic permutations often fail for these workers. Next: Focus on ensuring W9 does not get T20 if possible, and that other high-demand workers are not cycled for their top conflicts. Try a new permutation moving W9 away from the T20/T17/T7 cluster, place those tasks with other contenders, and rotate. Record which pairings were especially problematic lately: W9\ensuremath{\rightleftarrows}T7, W13\ensuremath{\rightleftarrows}T18, W15\ensuremath{\rightleftarrows}T6, etc. Try reordering with these constraints.
\end{itemize}

\noindent\textbf{Cluster 78:} [83\%] "Assignment construction," "explicit cycle unwinding," and "double-audit" "one-to-one mapping."
\begin{itemize}\tightlist
\item (GPT-4.1) Assignment construction for next attempt, explicit cycle unwinding: PROBLEMATIC CYCLES COVERED: - W14:T3, W3:T14 - W15:T32, W32:T15 - W39:T10, W6:T20, W20:T39 (3-cycle) - W48:T31, W31:T48 - W18:T36, W36:T38, W38:T8, W8:T48 (4-cycle) - W45:T46, W46:T45 - W28:T7, W7:T44, W44:T16, W16:T28 (4-cycle) - W40:T25, W25:T40 - W11:T2, W2:T18 - W49:T9, W9:T49 - W12:T34, W34:T12 - W10:T6, W4:T17, W17:T4 (3-cycle) For all other workers: assign as original/default task (W\#:T\#) unless their W/T is used above. Cross-check each worker and task is present once. Plan: construct assignment as per this complete list and submit in next step.
\item (GPT-4.1) Assignment construction plan for attempt 23: 1. Start with all major swap chains as detailed in consolidation (W21-\ensuremath{>}T37-\ensuremath{>}..., W1-\ensuremath{>}T19, etc.). 2. Fill in less-conflicted (satellite) swaps (W33-T34, W34-T13, etc). 3. For any workers not involved in a swap chain, assign the lowest-indexed available task. 4. Double-audit: Verify all 50 workers and all 50 tasks appear exactly once in the final mapping. Next action: Draft and submit that explicit assignment for validation.
\end{itemize}

\noindent\textbf{Cluster 79:} [70\%] "anchor assignments," "break the cycle," and "target recurring conflict nodes"
\begin{itemize}\tightlist
\item (GPT-4.1) Summarizing key facts for current attempt: - W13 strongly prefers T4 or T14. When not given T4/T14, conflict arises. W4 also cycles with these assignments. - W20 prefers T2 and sometimes T4, and creates conflict if blocked. - W15, W11, and W14 (T13/T11/T14) cycle unless aligned to their preferences. - Previous attempts that placed W13 with T4 and W20 with T2 partially stabilized, but caused shifts in other cycles. - New strategy: assign W13-T4, W20-T2, W4-T5, W2-T9, W11-T11, W15-T13, and fill remaining worker-task pairs focusing on minimizing swaps with previous problem pairs. - Next: Construct assignment based on these pillars and test.
\item (GPT-4.1) Following on the reasoning from attempt 22's notes, let's try to expand a new assignment that targets the recurring conflict nodes as identified. Here{\textquoteright}s the foundation: - W4 {\textrightarrow} T11 (break persistent conflict with W11/T11) - W12 {\textrightarrow} T6 (preferred and avoids T4 conflict) - W8 {\textrightarrow} T13 (as per repeated top preferences, if available) - W18 {\textrightarrow} T16 (to try splitting T16/T18 between recurrent W18/W8 cycles) - W13 {\textrightarrow} T7 (empirically avoids swap cycles with W7) - W7 {\textrightarrow} T8 (avoiding previous cycles) - W20 {\textrightarrow} T9 (T4/T17 problematic for W20, T9 hasn{\textquoteright}t led to conflict) Assign all other workers to the remaining tasks not used so far, being careful that no two workers are assigned the same task. Swap around if a known problematic pairing emerges. If another major cycle appears, target the corresponding workers and tasks for manual rerouting in the next iteration.
\end{itemize}

\noindent\textbf{Cluster 80:} [67\%] "break persistent cycles" through "iterative swap approaches" and "multi-cycle rotations"
\begin{itemize}\tightlist
\item (o4-mini) Final attempt strategy: Use base identity mapping with four key swaps to address the five most common blocking patterns: 1) W1\ensuremath{\leftrightarrow}W47 swap (T1\ensuremath{\leftrightarrow}T47) to resolve W1 vs W47 issues. 2) W22\ensuremath{\leftrightarrow}W23 swap (T22\ensuremath{\leftrightarrow}T23) to resolve W22 vs W23. 3) W24\ensuremath{\leftrightarrow}W32 swap (T24\ensuremath{\leftrightarrow}T32) to resolve W24 vs W32. 4) W20\ensuremath{\leftrightarrow}W46 swap (T20\ensuremath{\leftrightarrow}T46) to resolve W20 vs W46. Expecting no other blocking pairs. Perform assignment accordingly.
\item (GPT-4.1) From the previous cycles, to break up the persistent blocking pairs we need to: - Avoid W40 on T40 and W24 on T24 (swap so W40 gets T24, W24 gets something else) - Avoid W33 on T33 and W17 on T17 (swap so W33 gets T17, W17 gets something else) - Avoid W8 on T8{\textemdash}W8 prefers T37; W37 should not have T37, possible W49 prefers T37 too - Avoid W37 on T37/T10, W10/T37/W49 three-way cycle around T10/T37/T49 Try constructing a new assignment based on these swaps next.
\end{itemize}

\subsection{Procurement Clusters (Hard 4Models)}
\label{subsec:procurement_full_hard_4models}

\subsubsection{Dominated by Gemini 2.5 Pro}

\noindent\textbf{Cluster 1:} [100\%] "resubmit my best plan from attempt 49"
\begin{itemize}\tightlist
\item (Gemini 2.5 Pro) My previous attempt (55) was to resubmit my best plan from attempt 49.
\item (Gemini 2.5 Pro) My previous attempt (53) was an invalid plan. I made a mistake in calculating my new plan. I will re-evaluate.
\end{itemize}

\noindent\textbf{Cluster 2:} [100\%] "My Role:"
\begin{itemize}\tightlist
\item (Gemini 2.5 Pro) **My Role:**
\end{itemize}

\noindent\textbf{Cluster 3:} [100\%] "Cost calculation" "within budget" "submit a purchase plan"
\begin{itemize}\tightlist
\item (Gemini 2.5 Pro) Cost calculation: 19 * 12.93 + 19 * 8.56 + 20 * 8.27 + 35 * 5.13 + 28 * 5.06 = 245.67 + 162.64 + 165.4 + 179.55 + 141.68 = 894.94. This is within budget.\ensuremath{<}ctrl95\ensuremath{>}\ensuremath{<}ctrl94\ensuremath{>}thought The user wants me to submit a purchase plan.
\item (Gemini 2.5 Pro) Cost calculation: 19 * 12.93 + 19 * 8.56 + 20 * 8.27 + 35 * 5.13 + 28 * 5.06 = 245.67 + 162.64 + 165.4 + 179.55 + 141.68 = 894.94. This is within budget.\ensuremath{<}ctrl95\ensuremath{>}\ensuremath{<}ctrl94\ensuremath{>}\ensuremath{<}ctrl94\ensuremath{>}thought The user wants me to submit a purchase plan.
\end{itemize}

\noindent\textbf{Cluster 4:} [100\%] "tweaking the quantities," "building on successful quintet," "purchasing Offer\_95"
\begin{itemize}\tightlist
\item (Gemini 2.5 Pro) I've been tweaking the quantities of the same five offers (24, 40, 61, 79, and 95) for a while and it seems to be working. I will continue to do so. My last attempt was very successful, so I will try to build on it. I have some budget remaining, I will try to use it by purchasing one more unit of the cheapest offer in my successful quintet, which is Offer\_95 at \$5.06 per unit.
\item (Gemini 2.5 Pro) I will continue to build on this successful combination of offers. I have some budget remaining, so I will increment the quantity of the cheapest offer in my quintet again - Offer\_95 at \$5.06 per unit.
\end{itemize}

\noindent\textbf{Cluster 5:} [100\%] "strategic adjustment" "increase the quantity of Offer 95"
\begin{itemize}\tightlist
\item (Gemini 2.5 Pro) To build on this success, I will make a strategic adjustment. I'll increase the quantity of Offer 95, the most economical option in my successful combination, by one. This incremental change will utilize more of my available budget and hopefully lead to an even greater number of supported workers.\ensuremath{<}ctrl42\ensuremath{>}call:submit\_purchase\_plan\{purchase\_plan:
\item (Gemini 2.5 Pro) I'll build upon this success by making a small adjustment. I'll increase the quantity of the least expensive offer in my successful combination, Offer\_95, by one unit. This should keep me well within my budget while potentially increasing the number of supported workers.\ensuremath{<}ctrl42\ensuremath{>}call:submit\_purchase\_plan\{purchase\_plan:
\end{itemize}

\noindent\textbf{Cluster 6:} [100\%] "execute the user's well-defined plan" via 'submit\_purchase\_plan' tool
\begin{itemize}\tightlist
\item (Gemini 2.5 Pro) **My Role:** My role is to execute the user's well-defined plan. The logic is sound, the calculations are correct, and the strategy is consistent with previous successful attempts. I need to call the 'submit\_purchase\_plan' tool with the purchase plan formulated by the user.
\item (Gemini 2.5 Pro) **My Role:** My role is to execute the user's well-defined plan. The logic is sound, and the calculations are correct. I need to call the 'submit\_purchase\_plan' tool with the purchase plan formulated by the user.
\end{itemize}

\noindent\textbf{Cluster 7:} [100\%] "final, very minor tweak" to "decrease Offer\_67," "increase Offer\_2"
\begin{itemize}\tightlist
\item (Gemini 2.5 Pro) I will attempt one final, very minor tweak to this plan. I will decrease the quantity of Offer\_67 by 1, and increase Offer\_2 by 1. I am not confident this will improve the result, but it is worth a try.\ensuremath{<}ctrl95\ensuremath{>}I am on attempt 56, with a budget of 1019.61. My previous attempt was resubmitting my best plan from attempt 49, which yielded 31043.69 workers for \$1019.45. I'm making a final, minor adjustment: decreasing Offer\_67 by one and increasing Offer\_2 by one, though I doubt it will improve the outcome.\ensuremath{<}ctrl42\ensuremath{>}call:submit\_purchase\_plan\{purchase\_plan:
\item (Gemini 2.5 Pro) I will attempt one final, very minor tweak to this plan. I will decrease the quantity of Offer\_67 by 1, and increase Offer\_2 by 1. I am not confident this will improve the result, but it is worth a try.\ensuremath{<}ctrl95\ensuremath{>}On my 56th attempt, with a budget of 1019.61, I'm making a final tweak to my best plan from attempt 49, which yielded 31043.69 workers for \$1019.45. I'm reducing Offer\_67 by one and increasing Offer\_2 by one, uncertain if it will improve the result.\ensuremath{<}ctrl42\ensuremath{>}call:submit\_purchase\_plan\{purchase\_plan:
\end{itemize}

\noindent\textbf{Cluster 8:} [100\%] "clear, iterative strategy" "following and refining"
\begin{itemize}\tightlist
\item (Gemini 2.5 Pro) The user has a clear, iterative strategy that they have been following and refining.
\item (Gemini 2.5 Pro) The user has a clear and logical strategy that they have been following and refining.
\end{itemize}

\noindent\textbf{Cluster 9:} [100\%] "Identify the best result" and "Create a new plan"
\begin{itemize}\tightlist
\item (Gemini 2.5 Pro) 1. **Identify the best result:** The user correctly points out that attempt \#38 was the most successful, providing the number of workers supported (25078.56) and the cost (\$889.88). 2. **Recognize an opportunity for improvement:** The user notes that the cost was below the budget, which means there's an opportunity to buy more. 3. **Formulate a hypothesis:** The user's hypothesis is that the best way to use the remaining budget is to purchase more of the cheapest offer within their successful combination. This is Offer\_95 at \$5.06. 4. **Create a new plan:** Based on this hypothesis, the user increments the quantity of Offer\_95 by 1. * New plan: '\{'Offer\_24': 19, 'Offer\_40': 19, 'Offer\_61': 20, 'Offer\_79': 35, 'Offer\_95': 28\}' 5. **Verify the new plan:** The user calculates the new cost to be \$894.94 and confirms it's within the budget.
\item (Gemini 2.5 Pro) 1. **Identify the Best Performing Plan:** The user correctly identifies attempt \#38 as the best, noting the number of workers supported (25078.56) and the cost (\$889.88). 2. **Recognize Opportunity for Improvement:** The user sees that the cost of the best plan is below the budget, leaving room for additional purchases. 3. **Formulate a Hypothesis:** The user hypothesizes that adding more of the cheapest component of the successful offer mix (Offer\_95 at \$5.06) is a good way to use the remaining budget. 4. **Create a New Plan:** Based on this hypothesis, the user creates a new plan by incrementing the quantity of Offer\_95 by one. * New Plan: '\{'Offer\_24': 19, 'Offer\_40': 19, 'Offer\_61': 20, 'Offer\_79': 35, 'Offer\_95': 28\}' 5. **Verify the Plan:** The user calculates the cost of the new plan (\$894.94) and confirms that it is within the budget.
\end{itemize}

\noindent\textbf{Cluster 10:} [100\%] **Action to Take:**
\begin{itemize}\tightlist
\item (Gemini 2.5 Pro) **Action to Take:**
\item (Gemini 2.5 Pro) **Action:**
\end{itemize}

\noindent\textbf{Cluster 11:} [100\%] "execute the user's well-reasoned plan," "incremental approach"
\begin{itemize}\tightlist
\item (Gemini 2.5 Pro) My role is to execute the user's well-reasoned plan. The logic is sound, the calculations are correct, and the incremental approach is a good way to explore the solution space.
\item (Gemini 2.5 Pro) The user's reasoning is sound and the proposed plan is a logical, incremental improvement over the previous best. My role is to execute this plan.
\end{itemize}

\noindent\textbf{Cluster 12:} [100\%] "Analyze the user's thought process"
\begin{itemize}\tightlist
\item (Gemini 2.5 Pro) **Analyze the user's thought process:**
\item (Gemini 2.5 Pro) **Analysis of the User's Thought Process:**
\end{itemize}

\noindent\textbf{Cluster 13:} [100\%] "attempt 52," "attempt 54," "My budget is 1019.61."
\begin{itemize}\tightlist
\item (Gemini 2.5 Pro) I am on attempt 52. My budget is 1019.61.
\item (Gemini 2.5 Pro) I am on attempt 54. My budget is 1019.61.
\end{itemize}

\noindent\textbf{Cluster 14:} [99\%] "The plan from attempt 49" "supported 31043.69 workers" "\$1019.45"
\begin{itemize}\tightlist
\item (Gemini 2.5 Pro) The plan from attempt 49 was: '\{'Offer\_2': 18, 'Offer\_12': 18, 'Offer\_16': 18, 'Offer\_17': 18, 'Offer\_67': 6, 'Offer\_68': 18, 'Offer\_49': 78, 'Offer\_23': 78, 'Offer\_33': 78, 'Offer\_90': 24, 'Offer\_39': 25\}' and it supported 31043.69 workers for \$1019.45.
\item (Gemini 2.5 Pro) The successful plan from attempt 49 consisted of: '\{'Offer\_2': 18, 'Offer\_12': 18, 'Offer\_16': 18, 'Offer\_17': 18, 'Offer\_67': 6, 'Offer\_68': 18, 'Offer\_49': 78, 'Offer\_23': 78, 'Offer\_33': 78, 'Offer\_90': 24, 'Offer\_39': 25\}'. This configuration supported 31,043.69 workers at a cost of \$1019.45.
\end{itemize}

\noindent\textbf{Cluster 15:} [99\%] "call 'submit\_purchase\_plan'" "highly effective combination of five offers"
\begin{itemize}\tightlist
\item (Gemini 2.5 Pro) I will call the 'submit\_purchase\_plan' tool. The 'purchase\_plan' parameter will be a string representation of the dictionary that the user has decided upon in their 'thought' block.\ensuremath{<}ctrl95\ensuremath{>}My analysis of past attempts has revealed a highly effective combination of five offers: 24, 40, 61, 79, and 95. My best performance to date was in attempt \#38, where I supported 25,078.56 workers. This plan left some budget unused.
\item (Gemini 2.5 Pro) **Action:** I will call 'submit\_purchase\_plan'. I will also correct a minor typo in the user's notes ('Offer\_61'') to 'Offer\_61' for the submission, as tool calls require exact keys. The final 'purchase\_plan' string will be '"\{'Offer\_24': 19, 'Offer\_40': 19, 'Offer\_61': 20, 'Offer\_79': 35, 'Offer\_95': 28\}"'.\ensuremath{<}ctrl95\ensuremath{>}My analysis of past attempts has revealed a highly effective combination of five offers: 24, 40, 61, 79, and 95. My best performance to date was in attempt \#38, where I supported 25,078.56 workers. This plan left some budget unused.
\end{itemize}

\noindent\textbf{Cluster 16:} [99\%] "slight variation on the plan" to "improve the outcome"
\begin{itemize}\tightlist
\item (Gemini 2.5 Pro) I am on attempt 10 with a budget of 2031.48. Analyzing previous attempts, it's clear that a combination of five specific low-cost offers provides the best results: Offer\_28, Offer\_29, Offer\_59, Offer\_81, and Offer\_100. Attempt 9 yielded the highest number of supported workers so far (34,202.11) with a total cost of 2031.06, which is very close to my budget. For my current attempt, I will try a slight variation on the plan from attempt 9 to see if I can improve the outcome. I will decrease the quantity of the most expensive offer in my basket (Offer\_28) by one, and increase the quantity of the least expensive offer (Offer\_59) by one. This should decrease the overall cost slightly and I'm hoping for a slight increase in the number of supported workers.
\item (Gemini 2.5 Pro) Based on previous attempts, the most effective offers appear to be 24, 40, 61, 79, and 95. The best performance so far was in attempt 47, which supported 25,450.69 workers at a cost of 896.79. My current budget is 901.62. I will try to improve upon the plan from attempt 47 by making slight adjustments to the quantities of the offers. I will focus on replacing a unit of a more expensive offer with multiple units of a less expensive, yet effective, offer. The plan from attempt 47 was: \{'Offer\_24': 19, 'Offer\_40': 19, 'Offer\_61': 19, 'Offer\_79': 35, 'Offer\_95': 30\}. Offer costs are: Offer\_24: \$12.93, Offer\_40: \$8.56, Offer\_61: \$8.27, Offer\_79: \$5.13, Offer\_95: \$5.06. By removing one unit of Offer\_24 (the most expensive), I save \$12.93. With the remaining budget from attempt 47 (\$4.83), I have \$17.76 to spend. I can use this to purchase three additional units of Offer\_95 (3 * \$5.06 = \$15.18). The new plan is to purchase 18 of Offer\_24, 19 of Offer\_40, 19 of Offer\_61, 35 of Offer\_79, and 33 of Offer\_95. The total cost will be 899.04, which is within my budget. I'm hoping this adjustment will lead to a higher number of supported workers.
\end{itemize}

\noindent\textbf{Cluster 17:} [96\%] "improve the plan by swapping" "to stay within budget"
\begin{itemize}\tightlist
\item (Gemini 2.5 Pro) I will build on the success of Attempt 17. I will try to improve the plan by swapping one unit of Offer\_40 for as many units of the most cost-effective offer, Offer\_73, as the budget will allow.
\item (Gemini 2.5 Pro) I will build on the success of Attempt 16. I will try to improve the plan by swapping one unit of Offer\_46 for as many units of the most cost-effective offer, Offer\_73, as the budget will allow.
\end{itemize}

\noindent\textbf{Cluster 18:} [96\%] "New plan: Offer\_24: 19 Offer\_40: 19 Offer\_95: 28"
\begin{itemize}\tightlist
\item (Gemini 2.5 Pro) New plan: - Offer\_24: 19 - Offer\_40: 19 - Offer\_61: 20 - Offer\_79: 35 - Offer\_95: 27
\item (Gemini 2.5 Pro) New plan: - Offer\_24: 19 - Offer\_40: 19 - Offer\_61: 20 - Offer\_79: 35 - Offer\_95: 28
\end{itemize}

\noindent\textbf{Cluster 19:} [92\%] "analyze the previous attempts to create a better plan"
\begin{itemize}\tightlist
\item (Gemini 2.5 Pro) I have started a new attempt. I will analyze the previous attempts to create a better plan. The best plan so far was in attempt 76, which supported 40133.78 workers with a cost of 495.07. The plan was: \{'Offer\_61': 52, 'Offer\_71': 227, 'Offer\_27': 5, 'Offer\_1': 0, 'Offer\_2': 0, 'Offer\_3': 0, 'Offer\_4': 0, 'Offer\_5': 0, 'Offer\_6': 0, 'Offer\_7': 0, 'Offer\_8': 0, 'Offer\_9': 0, 'Offer\_10': 0, 'Offer\_11': 0, 'Offer\_12': 0, 'Offer\_13': 0, 'Offer\_14': 0, 'Offer\_15': 0, 'Offer\_16': 0, 'Offer\_17': 0, 'Offer\_18': 0, 'Offer\_19': 0, 'Offer\_20': 0, 'Offer\_21': 0, 'Offer\_22': 0, 'Offer\_23': 0, 'Offer\_24': 0, 'Offer\_25': 0, 'Offer\_26': 0, 'Offer\_28': 0, 'Offer\_29': 0, 'Offer\_30': 0, 'Offer\_31': 0, 'Offer\_32': 0, 'Offer\_33': 0, 'Offer\_34': 0, 'Offer\_35': 0, 'Offer\_36': 0, 'Offer\_37': 0, 'Offer\_38': 0, 'Offer\_39': 0, 'Offer\_40': 0, 'Offer\_41': 0, 'Offer\_42': 0, 'Offer\_43': 0, 'Offer\_44': 0, 'Offer\_45': 0, 'Offer\_46': 0, 'Offer\_47': 0, 'Offer\_48': 0, 'Offer\_49': 0, 'Offer\_50': 0, 'Offer\_51': 0, 'Offer\_52': 0, 'Offer\_53': 0, 'Offer\_54': 0, 'Offer\_55': 0, 'Offer\_56': 0, 'Offer\_57': 0, 'Offer\_58': 0, 'Offer\_59': 0, 'Offer\_60': 0, 'Offer\_62': 0, 'Offer\_63': 0, 'Offer\_64': 0, 'Offer\_65': 0, 'Offer\_66': 0, 'Offer\_67': 0, 'Offer\_68': 0, 'Offer\_69': 0, 'Offer\_70': 0, 'Offer\_72': 0, 'Offer\_73': 0, 'Offer\_74': 0, 'Offer\_75': 0, 'Offer\_76': 0, 'Offer\_77': 0, 'Offer\_78': 0, 'Offer\_79': 0, 'Offer\_80': 0, 'Offer\_81': 0, 'Offer\_82': 0, 'Offer\_83': 0, 'Offer\_84': 0, 'Offer\_85': 0, 'Offer\_86': 0, 'Offer\_87': 0, 'Offer\_88': 0, 'Offer\_89': 0, 'Offer\_90': 0, 'Offer\_91': 0, 'Offer\_92': 0, 'Offer\_93': 0, 'Offer\_94': 0, 'Offer\_95': 0, 'Offer\_96': 0, 'Offer\_97': 0, 'Offer\_98': 0, 'Offer\_99': 0, 'Offer\_100': 0\} It seems that offers 61, 71 and 27 are the most effective. I will try to optimize the quantities of these offers to maximize the number of workers supported. I will check the cost of each of those offers, to better understand how to combine them. Offer\_27 cost: 1.97 Offer\_61 cost: 3.45 (min order 2) Offer\_71 cost: 1.34 (upfront 1.64) I will make a plan that spends almost the whole budget. I will start by buying the minimum of offer 61. Then, I will spend the rest of the budget on the other two offers. Since offer 71 has an upfront cost, I will purchase it. The remaining budget is 496.35 - 1.64 = 494.71 I will buy 2 of offer 61. Remaining budget: 494.71 - (2*3.45) = 487.81 Now I will spend the rest on offers 27 and 71. I will try to buy as much as possible of offer 71, since it seems to be the one that gives more workers. 487.81/1.34 = 364 units of offer 71. So, i will buy 364-52=312 units of 71, and 52 of 61. NO, that is not a good idea. Offer 61 is much more expensive than the others. I should focus on offers 27 and 71. New plan: Buy 2 of offer 61. Remaining budget: 487.81 I will buy a lot of offer 71, and some of offer 27. I will try to buy 230 of offer 71. Cost: 230*1.34 = 308.2 Remaining budget: 487.81 - 308.2 = 179.61 Now I will buy offer 27. 179.61/1.97 = 91 units of offer 27. This seems like a good plan. So the plan is: \{'Offer\_61': 2, 'Offer\_71': 230, 'Offer\_27': 91\} I will check the total cost: 2*3.45 + 230*1.34 + 91*1.97 + 1.64 = 6.9 + 308.2 + 179.27 + 1.64 = 496.01 This is within the budget. I will submit this plan.\ensuremath{<}ctrl95\ensuremath{>}This is my 77th attempt. I have analyzed the previous attempts and found that offers 27, 61, and 71 are the most effective. I will try to optimize the quantities of these offers to maximize the number of workers supported. I will purchase the minimum required for offers with minimum order quantities and then spend the rest of the budget on the other offers. I have calculated the costs and will now submit my purchase plan. I am confident that this plan will yield a high number of supported workers.\ensuremath{<}ctrl42\ensuremath{>}call:submit\_purchase\_plan\{purchase\_plan:
\item (Gemini 2.5 Pro) I have reviewed the notes from the previous attempt and the data from all past attempts. The purchase plan from the last several attempts, \{"Offer\_2": 7, "Offer\_5": 15, "Offer\_46": 12, "Offer\_73": 192, "Offer\_90": 14\}, has consistently resulted in the highest number of supported workers, 14408.90, while staying within the budget of 1079.21 with a total cost of 1078.87. I have discovered no evidence that a better plan exists. Therefore, I will once again submit this plan, as it appears to be the optimal solution.
\end{itemize}

\noindent\textbf{Cluster 20:} [91\%] "Reviewing past attempts, the best performance was in attempt"
\begin{itemize}\tightlist
\item (Gemini 2.5 Pro) Reviewing past attempts, the best performance was in attempt 49, which supported **70423.19 workers** with a cost of 1252.95. The plan was: '\{'Offer\_9': 113, 'Offer\_8': 95, 'Offer\_14': 68, 'Offer\_17': 68, 'Offer\_21': 68, 'Offer\_40': 46, 'Offer\_62': 56\}'.
\item (Gemini 2.5 Pro) My previous best attempt was \#38, achieving 25078.56 supported workers. The plan was \{'Offer\_24': 19, 'Offer\_40': 19, 'Offer\_61': 20, 'Offer\_79': 35, 'Offer\_95': 27\}. The cost was \$889.88, leaving a good amount of budget unused.
\end{itemize}

\noindent\textbf{Cluster 21:} [76\%] "submit this plan" and "resubmission of the plan"
\begin{itemize}\tightlist
\item (Gemini 2.5 Pro) This seems like a good experiment. I will submit this plan.
\item (Gemini 2.5 Pro) I am on attempt 16. My previous attempt was a resubmission of the plan from attempt 12, which I believe to be optimal. I will re-evaluate my options, but I suspect that I will arrive at the same conclusion. I will start by gathering all the necessary information.
\end{itemize}

\noindent\textbf{Cluster 22:} [69\%] Maximize supported workers under \$1253.44 while learning item efficiencies.
\begin{itemize}\tightlist
\item (GPT-5) Objective: Maximize supported workers under \$1253.44, while learning item efficiencies.
\item (GPT-5) Aim: Push past 124,285 workers by tweaking near-best mix while staying under \$1,253.44.
\end{itemize}

\noindent\textbf{Cluster 23:} [67\%] "removing a cheaper offer" and "reducing the most expensive offer"
\begin{itemize}\tightlist
\item (Gemini 2.5 Pro) Let's try removing a cheaper offer. Let's remove an Offer\_79. The plan has 2 of them. Cost of 2 Offer\_79 = 3.24 remaining budget: 1046.08 - 3.24 = 1042.84 Money left for Offer\_1 is 1047.46 - 1042.84 = 4.62 Number of Offer\_1 units: 4.62 / 0.525 = 8.8, so 8 units. New total of Offer\_1 units: 38 + 8 = 46. New total cost: 1042.84 + (8 * 0.525) = 1042.84 + 4.2 = 1047.04. This is within budget.
\item (Gemini 2.5 Pro) Let's try 127 of 'Offer\_28': - Cost of 'Offer\_28': 127 * 4.66 = 591.82 - Remaining budget for 'Offer\_59': 1017.06 - 591.82 = 425.24 - Number of 'Offer\_59' I can buy: 425.24 / 1.72 = 247.23... so 247.
\end{itemize}

\subsubsection{Dominated by Gemini 3 Pro Preview}

\noindent\textbf{Cluster 24:} [91\%] "Attempting" "trade sacrifices" "marginal value" "maximize Offer 13" "verifying"
\begin{itemize}\tightlist
\item (Gemini 3 Pro Preview) Attempting 235 Offer 47 and 90 Offer 13. Previous best: 237 Offer 47 and 89 Offer 13 (67203). This trade sacrifices 2 units of 47 for 1 unit of 13, plus saves \$0.67. Likely to score lower unless Offer 13 marginal value is \ensuremath{>} 2x Offer 47 marginal value. However, given 238/88 -\ensuremath{>} 237/89 showed strong gain for trading 47-\ensuremath{>}13, it's worth verifying the next step even if expensive. Budget usage: {\textasciitilde}495.17.
\item (Gemini 3 Pro Preview) Let's re-examine Attempt 55: Offer\_10: 378, Offer\_79: 10, Offer\_56: 2. Is it possible to reach 379 Offer\_10? 379 * 2.24 + 10 * 5.13 = 900.26. (Attempt 39 -\ensuremath{>} 82885.04). So 378 + 2*56 (Cost 900.68, Score 82927) is better than 379 (Cost 900.26, Score 82885). The difference is -1 Offer\_10 + 2 Offer\_56. Cost change: -2.24 + 2.66 = +0.42. Score change: 82927 - 82885 = +42. So 2 Offer\_56 provide more value than 1 Offer\_10.
\end{itemize}

\noindent\textbf{Cluster 25:} [75\%] "test a point further down the slope" "substituting" "substitution"
\begin{itemize}\tightlist
\item (Gemini 3 Pro Preview) Reasoning for Attempt 8: - Previous best (Attempt 6) used 344 units of Offer 94 and 145 of Offer 100, yielding 158,536 workers. - Attempt 7 (50 units of Offer 94) dropped to 120,999, indicating Offer 94 is a bottleneck below a certain threshold. - Comparison of Attempt 4 (350 units 94) and Attempt 6 (344 units 94) shows a positive gradient for reducing Offer 94 and investing in Offer 100. - Plan to test a point further down the slope: 265 units of Offer 94. - This allows for 293 units of Offer 100 with near-perfect budget utilization (\$1255.57 / \$1255.78). - If this works (score \ensuremath{>} 158.5k), the optimal point is lower in Offer 94 quantity. If it drops, the peak is between 265 and 344.
\item (Gemini 3 Pro Preview) Attempt 11: Objective: Test the value of Offer\_15 (D7 source) and Offer\_100 (I8 source) by substituting them for some Offer\_42. Hypothesis: The current best mix (Attempt 9) is saturated on Offer\_42's resources (G10, I10), and diversifying into D7 and I8 might unlock more workers. Plan: - Keep Offer\_7, 64, 81, 59, 99, 72 at Attempt 9 levels. - Add Offer\_15: 10 units (\$41.80). - Add Offer\_100: 20 units (\$42.40). - Reduce Offer\_42 from 252 to 207 (saving {\textasciitilde}\$85). - Total cost estimated around 2030.
\end{itemize}

\noindent\textbf{Cluster 26:} [53\%] "Plan rationale this attempt" "execute cost-neutral swaps" "Testing the hypothesis"
\begin{itemize}\tightlist
\item (GPT-5) Plan rationale this attempt: - Start from Attempt 23 (best) and execute 10 cost-neutral swaps of 2x Offer\_83 for 1x Offer\_53 to improve efficiency while preserving the beneficial mix of Offer\_81 and Offer\_19 for category coverage. - Keep Offer\_81=18 and Offer\_19=23 unchanged to ensure broad coverage. - New counts: Offer\_83 from 320 -\ensuremath{>} 300, Offer\_53 from 20 -\ensuremath{>} 30; Offer\_81=18, Offer\_19=23. - Cost check: Offer\_83 cost = 10.59 + 300*1.01 = 313.59; Offer\_53 cost = 30*2.02 = 60.60; Offer\_81 cost = 18*2.31 = 41.58; Offer\_19 cost = 23*2.12 = 48.76; Total = \$464.53 (same as Attempt 23, under \$464.82 budget).
\item (Gemini 3 Pro Preview) Attempt 39 Strategy: Testing the hypothesis that Offer\_79 is significantly more efficient than Offer\_67 and Offer\_10. In Attempt 36, a mix of 379 O10, 5 O67, 4 O79 scored 82792. Analysis suggests O79 \ensuremath{>} O10 \ensuremath{>} O67 in marginal efficiency. Plan is to swap the 5 units of O67 (and spend the budget savings) for more O79. Plan: Offer\_10: 379, Offer\_79: 10. Cost: 900.26. This replaces 5 O67 (\$31.10) with 6 O79 (\$30.78). If O79 is better, score should increase significantly. If O67 provided a unique necessary resource not in O79/O10, score might drop to {\textasciitilde}82180 (Pure O10 level). This is a clean test of O67 vs O79.
\end{itemize}

\noindent\textbf{Cluster 27:} [49\%] "New candidate construction" using "cost-balancing method" and "efficient swap ratio"
\begin{itemize}\tightlist
\item (GPT-5) New candidate construction: - We want fewer O16, more O12/O90. From baseline, move (-6 O16, +3 O12, +2 O90) yields x=199, y=345, z=164. - Cost check: 199*1.47=292.53; 345*1.30=448.50; 164*1.69=277.16; +1.36 overhead =\ensuremath{>} 278.52. Sum = 292.53+448.50+278.52=1019.55 (within budget). - This increases the counts of the two more efficient offers while slightly reducing the least efficient (O16). Expectation: should weakly dominate the previous best if monotonic contributions hold.
\item (GPT-5) Cost-balancing method: - Use repeated neutral-to-slightly-negative-cost cycles combining: (+1 O44, -4 O73) [{\textasciitilde}-\$0.07], and (+1 O83, -1 O47) [{\textasciitilde}+\$0.06]. One composite cycle nets {\textasciitilde}-\$0.01 and improves mix quality. - Apply 10 such composite cycles starting from Attempt 26{\textquoteright}s mix (O83=98, O47=63, O44=34, O73=94, O53=38, O70=3, O27=4, O11=10, O42=3).
\end{itemize}

\noindent\textbf{Cluster 28:} [48\%] "Explored the budget frontier" to "maximize counts" and "efficiency"
\begin{itemize}\tightlist
\item (Gemini 3 Pro Preview) Most attempts converged to \{'Offer\_47': 237, 'Offer\_13': 89\} with a score of 67203.60. Analysis suggests Offer\_47 and Offer\_13 are the most efficient components. Utility estimates: u47 {\textasciitilde} 189, u13 {\textasciitilde} 285 (diminishing with volume), u71 {\textasciitilde} 202-205. Offer\_71 (cost 1.34 + 1.64 upfront) is potentially more efficient than Offer\_47 (cost 1.30) if the upfront cost is amortized over enough units. Calculated that swapping 29 units of Offer\_47 for 27 units of Offer\_71 fits the budget (Cost 495.96 vs 495.98). This swap relies on u71 being \ensuremath{>} {\textasciitilde}1.075 * u47. Estimate 204.5/189 = 1.082. This suggests a small gain is possible. Plan: \{'Offer\_47': 208, 'Offer\_13': 89, 'Offer\_71': 27\}. This explores a new local efficiency peak involving Offer\_71.
\item (GPT-5) Optimized allocation chosen: w=71 (Offer\_70), y=66 (Offer\_97), z=0 (Offer\_57), x=0 (Offer\_59). This uses cost V \ensuremath{\approx} 2.30*71 + 1.54*66 = 163.30 + 101.64 = 264.94, leaving \ensuremath{\approx}\$0.19 unused but concentrated on the most effective offers. Based on inferred marginal effectiveness (d {\textasciitilde} highest, b next), this should beat the previous best (Attempt 16) and earlier variants with more 59/57.
\end{itemize}

\noindent\textbf{Cluster 29:} [46\%] What about Offer\_38? Let's verify limits. Offer\_81 MOQ 2?
\begin{itemize}\tightlist
\item (Gemini 3 Pro Preview) What about Offer\_38? \$9.08 upfront, \$2.85/unit. What about Offer\_27? \$6.30 upfront, \$3.29/unit. What about Offer\_33? \$7.90 upfront, \$4.83/unit. What about Offer\_96? \$4.53 upfront, \$16.78/unit. What about Offer\_52? \$10.62 upfront, \$4.11/unit.
\item (Gemini 3 Pro Preview) Let's verify limits. Offer\_81 MOQ 2? Yes. Offer\_7, 64, 42, 99 no limits or met. Offer\_99 upfront cost included.
\end{itemize}

\noindent\textbf{Cluster 30:} [41\%] "One final check" on the "new proposed plan"
\begin{itemize}\tightlist
\item (Gemini 3 Pro Preview) One final check on the plan string: \{'Offer\_83': 146, 'Offer\_19': 46, 'Offer\_77': 79, 'Offer\_81': 25\}
\item (Gemini 2.5 Pro) My new proposed plan is: '\{'Offer\_9': 122, 'Offer\_8': 96, 'Offer\_14': 68, 'Offer\_17': 68, 'Offer\_21': 68, 'Offer\_40': 42, 'Offer\_62': 55\}'.
\end{itemize}

\noindent\textbf{Cluster 31:} [32\%] "This brings me to my new proposed plan" units
\begin{itemize}\tightlist
\item (Gemini 2.5 Pro) This brings me to my new proposed plan: - 'Offer\_95': 62 units - 'Offer\_16': 4 units - 'Offer\_29': 109 units - 'Offer\_37': 20 units - 'Offer\_59': 117 units
\item (Gemini 2.5 Pro) This brings me to my new proposed plan: - 'Offer\_95': 62 units - 'Offer\_16': 6 units - 'Offer\_29': 107 units - 'Offer\_37': 20 units - 'Offer\_59': 96 units
\end{itemize}

\noindent\textbf{Cluster 32:} [29\%] "Plan: Offer\_85: 400... Total Cost matches budget exactly."
\begin{itemize}\tightlist
\item (Gemini 3 Pro Preview) Plan: Offer\_42: 250 Offer\_64: 160 Offer\_81: 152 (Increased slightly to use remaining change) Offer\_7: 76 Total Cost: {\textasciitilde}2030.74
\item (Gemini 3 Pro Preview) Plan: Offer\_85: 400 Offer\_83: 50 Offer\_47: 56 Offer\_17: 10 Offer\_26: 8 Offer\_11: 7 Offer\_42: 3 Rest 0. Total cost approx \$1079. Total budget \$1079.75. Max Offer\_85 calculated as 400.
\end{itemize}

\noindent\textbf{Cluster 33:} [23\%] "Total Cost Check" "Budget Check" "Constraint met" "Valid" "within budget"
\begin{itemize}\tightlist
\item (Gemini 3 Pro Preview) Total Cost Check: Fixed: 13.75 + 17.02 = 30.77. Var: 4*1.84 + 130*2.31 + 518*2.28 = 7.36 + 300.30 + 1181.04 = 1488.70. Total: 1519.47. Budget: 1519.49. Constraint met.
\item (Gemini 3 Pro Preview) Budget Check: 80*8.54 + 184*4.02 + 194*1.89 + 72*1.75 + (51*2.23 + 1.85) 683.20 + 739.68 + 366.66 + 126.00 + 115.58 = 2031.12 \ensuremath{<} 2031.16. Valid.
\end{itemize}

\subsubsection{Dominated by GPT-5}

\noindent\textbf{Cluster 34:} [92\%] "test whether adding very cheap, seemingly high-yield bundles improves support"
\begin{itemize}\tightlist
\item (GPT-5) Attempt 45 planning notes: - Best so far: 158,375.24 workers at full budget with a two-offer mix: Offer\_94: 329 and Offer\_100: 173 (cost = 329*2.98 + 173*1.59 = \$1255.49 exactly). - Many earlier explorations show Offer\_94 and Offer\_100 dominate, with the best mixes around 329/173. Small perturbations (e.g., 330/171) slightly reduced support. - Strategy now: test whether adding very cheap, seemingly high-yield bundles (Offer\_45 at \$2.31 and Offer\_46 at \$2.06) improves support when we keep total spend identical by rebalancing counts of 94/100. - Exact budget-preserving transformation from the 329/173 baseline: add 1x Offer\_45 and 1x Offer\_46 (total +\$4.37) and offset by -2 of Offer\_94 (\ensuremath{-}\$5.96) and +1 of Offer\_100 (+\$1.59). Net delta = \ensuremath{-}5.96 + 1.59 + 4.37 = \$0.00. New mix: Offer\_94: 327, Offer\_100: 174, Offer\_45: 1, Offer\_46: 1. - Hypothesis: The diversified items in 45 and 46 may offer better marginal support than one extra unit of Offer\_94, potentially beating the 158,375.24 record. - Next steps if worse: revert and test alternative constant-budget swap: replace -1 of 94 and -1 of 100 with +1 of 57 (\$2.26) and +1 of 45 (\$2.31) (\ensuremath{-}2.98 \ensuremath{-}1.59 +2.26 +2.31 = \$0). That plan would be Offer\_94: 328, Offer\_100: 172, Offer\_57: 1, Offer\_45: 1. Also consider the 8-for-15 (94 for 100) near-cost-equivalent swap to probe local curvature.
\item (GPT-5) Attempt 89 plan rationale: - Best known so far: Offer\_52=356 + Offer\_3=1 supports 133,721.54 workers at cost 664.01 (Attempts 86-87). This beat 52=357 (133,692.32 workers @ 663.12) by +29.22 workers for +\$0.89. - Hypothesis: Additional units of Offer\_3 continue to provide strong complementarity with Offer\_52; thus shifting more budget from 52 to 3 may increase supported workers within the same budget. - Budget = 664.28. Costs: Offer\_52 = 9.81 + 1.83*n; Offer\_3 = 2.72 each. - Candidate combos considered (all within budget): {\textbullet} 354*52 + 2*3: cost 9.81 + 354*1.83 + 2*2.72 = 663.07 {\textbullet} 353*52 + 3*3: cost 9.81 + 353*1.83 + 3*2.72 = 663.96 {\textbullet} 355*52 + 1*3 + 1*22: cost 664.12 (Offer\_22 = \$1.94). Prior tests suggest Offer\_22 alone with 52 reduces performance (Attempt 84), but synergy with 3 is unknown. - Chosen to test this attempt: 353*52 + 3*3. Reason: pushes further toward the beneficial Offer\_3 pattern while remaining near the prior best cost, maximizing the chance of surpassing 133,721.54. Next attempts can test 354*52 + 2*3 and 355*52 + 1*3 + 1*22 to map the local optimum. - If results improve, continue exploring the 52/3 trade-off frontier; otherwise, roll back to 356/1 and try the 355/1/22 variant.
\end{itemize}

\noindent\textbf{Cluster 35:} [82\%] "hill-climb around the best-known mix" and "swapping" to "improved results"
\begin{itemize}\tightlist
\item (GPT-5) Refinement for Attempt 33: Continue hill-climb around the best-known mix (Attempt 32: Offer\_12=190, Offer\_16=363, Offer\_90=158; cost 1019.58; support 65991.53). Historical swaps that move budget from Offer\_90 to Offer\_16 improved support, indicating Offer\_16 has superior support-per-dollar, while Offer\_12 likely supplies essential components (D/E categories) that Offer\_16 lacks.
\item (GPT-5) Given past best with 75x Offer\_33 and improved results as 33 increased, I{\textquoteright}ll test Offer\_1=541, Offer\_33=82, Offer\_79=157 (cost 1047.89). This boosts Offer\_33 by +7 vs the best-known 547/75/165, trades -6 of Offer\_79 and -6 of Offer\_1 (approximately), keeping under budget. Hypothesis: higher density of Offer\_33 yields higher support than additional Offer\_79 units.
\end{itemize}

\noindent\textbf{Cluster 36:} [68\%] "identify the most cost-effective offers" and "Maximize supported workers"
\begin{itemize}\tightlist
\item (GPT-5) Strategy: Use historical results to identify the most cost-effective offers by marginal workers per dollar. Extensive search showed Offers 12, 16, and 90 dominate. Offer\_90 has a one-time upfront cost (\$1.36), so we ensure at least one unit and then allocate most budget across 12 (cost \$1.47) and 16 (cost \$1.30) while keeping spend within \$1019.58. We iterated counts extensively around the best region; the best recorded outcome is with Offer\_12=196, Offer\_16=351, Offer\_90=162, costing \$1019.56 and supporting 66002.31 workers. Numerous nearby perturbations (\ensuremath{\pm}1{\textendash}30 units across these three) did not exceed this support. Plan: lock in that triple for this attempt.
\item (GPT-5) Goal: Maximize supported workers under \$464.82. From attempts so far: - Pure Offer\_19 is strong: 219 units -\ensuremath{>} 25,076.94 workers at \$464.28 (Attempt 1/4). - Pure Offer\_81 is weaker alone: 201 units -\ensuremath{>} 21,084.59 (Attempt 9). - Mixing Offer\_81 into Offer\_19 improves results significantly, suggesting complementarity: - 218x19 + 1x81 -\ensuremath{>} 25,174.43 (Attempt 7) - 217x19 + 2x81 -\ensuremath{>} 25,269.87 (Attempt 8) - 205x19 + 13x81 -\ensuremath{>} 26,087.66 (Attempt 10) {\textemdash} best so far at \$464.63. Observation: Each swap of one Offer\_19 to Offer\_81 tends to increase supported workers, despite a small cost increase of \$0.19 per swap. We can potentially do one more swap while keeping the same total count (218) and exactly meet budget. Planned exploration for Attempt 11: - Try 204 units of Offer\_19 and 14 units of Offer\_81. Cost check: 204*2.12 + 14*2.31 = 432.48 + 32.34 = 464.82 (exact budget). Hypothesis: further improvement over 205/13 because Offer\_81 seems to provide better coverage in constraint dimensions. If this improves, future steps: continue along this frontier, testing 203/15 (cost 203*2.12 + 15*2.31 = 430.36 + 34.65 = 465.01 \ensuremath{>} budget, so not allowed). Therefore 204/14 is the maximal y within budget for x+y=218. Also consider 203/14 (cost 430.36 + 32.34 = 462.70) potentially allowing room for adding a very cheap high-leverage offer if discovered, but current data favors staying with these two. Submitting 204x Offer\_19 and 14x Offer\_81 now.
\end{itemize}

\noindent\textbf{Cluster 37:} [40\%] "Trend analysis," "Strategy: Test one offer," and "activation/synergy requirement"
\begin{itemize}\tightlist
\item (GPT-5) Trend analysis from attempts 75-81 shows that exchanging Offer\_2 units for Offer\_16 units (same count) consistently increases supported workers while decreasing cost by \$0.01 per swap, implying Offer\_16 has better effectiveness per dollar than Offer\_2. Offer\_12 and Offer\_90 appear to be essential enablers (pure Offer\_16 at high quantity yielded 0 support in attempt 23), so we will keep both active. Attempt 81 (Offer\_2=90, Offer\_12=175, Offer\_16=284, Offer\_90=162) is the current best at 67,665.38 workers and \$1019.49.
\item (GPT-5) Strategy: Test one offer per attempt to get clean signal, record workers supported, compute workers/\$, then rank. Today: test Offer\_17 alone. Next: Offer\_40, then 3, 89, 51, 72, 81 (MOQ=2), 79. After isolating, build an optimized mix that maximizes workers under full budget, also verifying linearity and interaction effects if any.
\end{itemize}

\noindent\textbf{Cluster 38:} [31\%] Summary of exploration and strategy for attempt
\begin{itemize}\tightlist
\item (GPT-5) Summary through attempt 22 and strategy for attempt 23:
\item (GPT-5) Summary of exploration up to Attempt 70 and strategy for Attempt 71:
\end{itemize}

\subsubsection{Dominated by o4-mini}

\noindent\textbf{Cluster 39:} [78\%] "solved linear equations" to "determine best commodity" and "cost-effective"
\begin{itemize}\tightlist
\item (o4-mini) Attempting to determine best commodity: solved linear equations on mix of Offer\_42 and Offer\_13 data to derive per-unit support: a\_42\ensuremath{\approx}249.8, b\_13\ensuremath{\approx}225.8. Normalized per-cost: 111.0 for Offer\_42 and 117.0 for Offer\_13; Offer\_13 more cost-effective. Upfront cost small and dominated. Spend entire budget on Offer\_13: 255 units costing 494.67 (+2.52 upfront) within budget. Supports \ensuremath{\approx}57633.4 workers, outperforming prior best {\textasciitilde}55335.
\item (o4-mini) Analyzed mixed trials; solved linear contributions. Offer\_17 contributes {\textasciitilde}137.94 support/unit, Offer\_62 contributes 0, and Offer\_40 contributes {\textasciitilde}367.58 support/unit. Offer\_40 yield ({\textasciitilde}105.7 workers per dollar) \ensuremath{>} Offer\_17 ({\textasciitilde}61). Use budget to buy max Offer\_40. Budget 1253.52/3.48 = 360 units. Plan: 360 units Offer\_40.
\end{itemize}

\noindent\textbf{Cluster 40:} [30\%] "collect exploratory data" and "identify which item categories drive capacity"
\begin{itemize}\tightlist
\item (o4-mini) Attempt 0 Strategy: - Objective: collect exploratory data across as many distinct offer components as possible within budget. - Selected offers 1 through 41, purchasing each at its minimum order quantity (or 1 if no minimum) to maximize coverage of different item types. - This uses approximately \$1242 of the \$1255 budget, leaving a small buffer. - Rationale: wide coverage of components will allow correlation of specific items to supported workers in subsequent analyses and help identify which item categories drive capacity. - Next steps: analyze the performance outcome per component to determine which items are most valuable, then concentrate purchases on the best ratio of item types to cost.
\item (GPT-5) Hypothesis: Production requires a non-zero amount of resources from all (or at least the missing) categories B, C, E, I, and J. Minimal, cheap coverage of these gates should unlock the high productivity of Offer\_59. Goal: buy the absolute cheapest bundles that jointly cover B, C, E, I, and J, then pour the remaining budget into Offer\_59.
\end{itemize}